\documentclass{article} % For LaTeX2e
\usepackage{iclr2026_conference,times}
\usepackage{graphicx}
\usepackage{amsmath,amsfonts,bm}

\def\eqref#1{equation~\ref{#1}}
\def\1{\bm{1}}

\DeclareMathAlphabet{\mathsfit}{\encodingdefault}{\sfdefault}{m}{sl}
\SetMathAlphabet{\mathsfit}{bold}{\encodingdefault}{\sfdefault}{bx}{n}

\usepackage{hyperref}
\usepackage{url}
\usepackage{booktabs} 
\usepackage{multirow}
\usepackage{tcolorbox}
\usepackage{contour}
\usepackage{bbding}
\usepackage{cleveref}
\usepackage[T1]{fontenc}
\usepackage{wrapfig}  % 确保导言区加载了 wrapfig
\usepackage{epigraph}
\usepackage[table]{xcolor}
\usepackage{caption}
\definecolor{mylightgreen}{HTML}{D7FFD7}

\title{WoW: Towards a World-omniscient World-model Through Embodied Interaction}

\iclrfinalcopy 
\author{
\small
\vspace{-2em}
\\
Xiaowei Chi\textsuperscript{\rm 1,2,3,$^{\dagger}$}, Peidong Jia\textsuperscript{\rm 1,2,$^{\dagger}$}, Chun-Kai Fan\textsuperscript{\rm 1,2,$^{\dagger}$}, Xiaozhu Ju\textsuperscript{\rm 1,$^{\dagger}$}, Weishi Mi\textsuperscript{\rm 1,$^{\dagger}$}, Zhiyuan Qin\textsuperscript{\rm 1,$^{\dagger}$},\\
Kevin Zhang\textsuperscript{\rm 2}, Wanxin Tian\textsuperscript{\rm 1}, Kuangzhi Ge\textsuperscript{\rm 2}, Hao Li\textsuperscript{\rm 1}, Zezhong Qian\textsuperscript{\rm 1,2}, Anthony Chen\textsuperscript{\rm 2}, \\ 
Qiang Zhou\textsuperscript{\rm 1,2},  Yueru Jia\textsuperscript{\rm 2}, Jiaming Liu\textsuperscript{\rm 2}, Yong Dai\textsuperscript{\rm 1}, Qingpo Wuwu\textsuperscript{\rm 2}, Chengyu Bai\textsuperscript{\rm 2}, Yu-Kai Wang\textsuperscript{\rm 2}, \\
Ying Li\textsuperscript{\rm 2}, 
Lizhang Chen\textsuperscript{\rm 1,2}, Yong Bao\textsuperscript{\rm 1}, Zhiyuan Jiang\textsuperscript{\rm 1}, Jiacheng Zhu\textsuperscript{\rm 1}, Kai Tang\textsuperscript{\rm 2}, Ruichuan An\textsuperscript{\rm 2}, Yulin Luo\textsuperscript{\rm 2},\\
Qiuxuan Feng\textsuperscript{\rm 1,2}, Siyuan Zhou\textsuperscript{\rm 3}, Chi-min Chan\textsuperscript{\rm 3}, Chengkai Hou\textsuperscript{\rm 1,2}, Wei Xue\textsuperscript{\rm 3}, Sirui Han\textsuperscript{\rm 3}, Yike Guo\textsuperscript{\rm 3},\\
Shanghang Zhang~\textsuperscript{2,\Envelope}, Jian Tang~\textsuperscript{1,\Envelope} \\
\small\textsuperscript{\rm 1} Beijing Innovation Center of Humanoid Robotics, \\
\small\textsuperscript{\rm 2} State Key Laboratory of Multimedia Information Processing, School of Computer Science, Peking University, \\
\small\textsuperscript{\rm 3} Hong Kong University of Science and Technology \\
}

\begin{document}

\maketitle

% \begin{figure*}[t]
%   \centering
%   \includegraphics[width=1.\linewidth]{figs/teaser.png}
%   \caption{Modality will gradually expand, until at the next break point(etc. tokenized is all you need), data is enough, and a novel framework could represent all these complex modalities, then the world model could become the real "models" as strong as the human mind model.}
%   \label{fig:wu-architecture}
% \end{figure*}
% % 目录

% \tableofcontents
% \newpage

\begin{figure*}[h]
\vspace{-3em}
  \centering
  \includegraphics[width=0.88\textwidth]{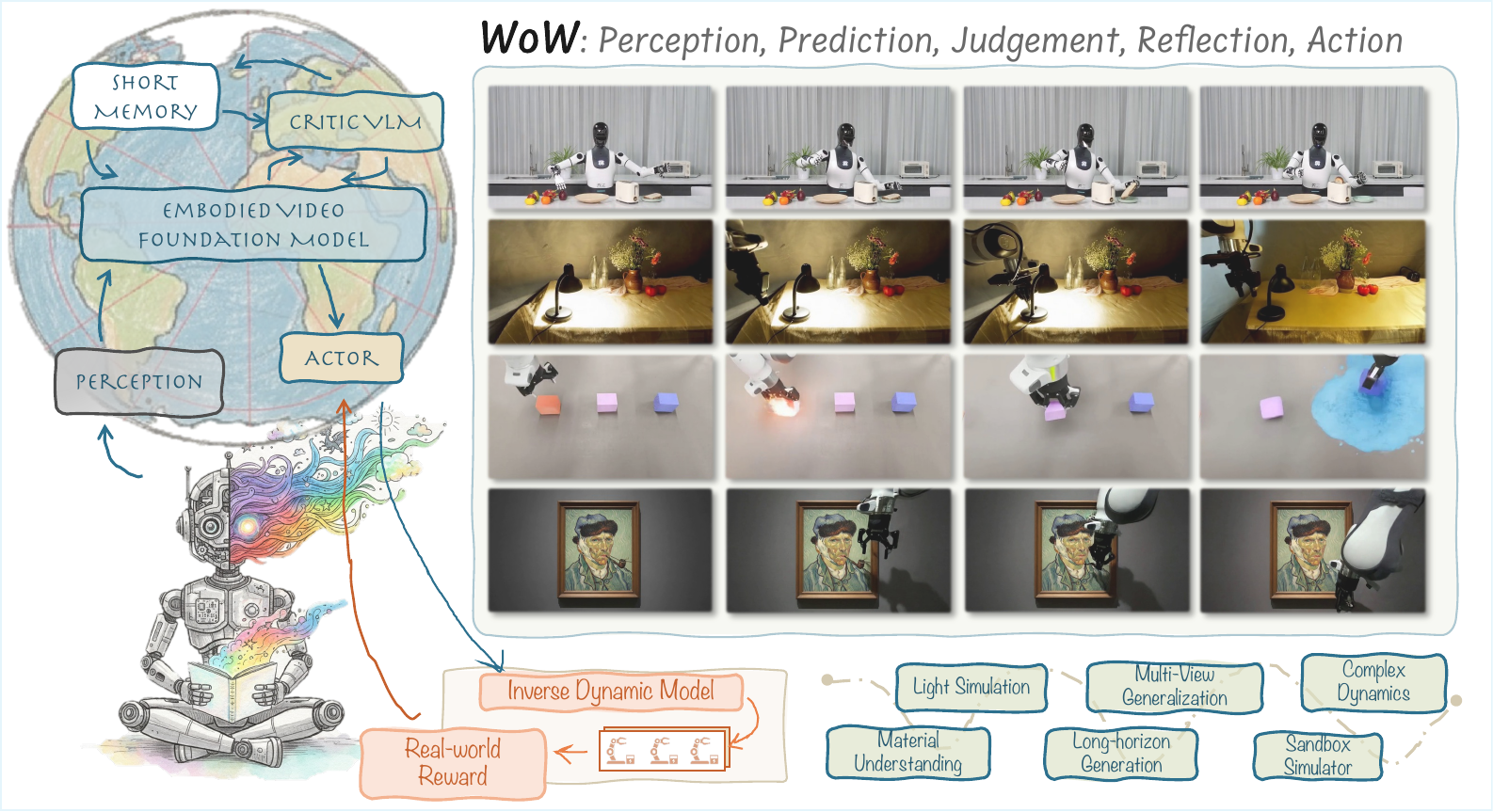}
  \caption{\textbf{WoW} is a world model that integrates perception, prediction, Judgement, reflection, and action. It learns from real-world interaction data and generates high-quality, physically consistent robot videos in seen and out-of-distribution scenarios, enabling real-world robotic execution.}
  \label{fig:teaser}
\vspace{-0.5em}
\end{figure*}
\begin{abstract}

% ========================================================================================================
% ============================================ Version 4.0 ============================================
% ================================================================================================================

% Humans build intuitive physics through active interaction with the world, a stark contrast to current video models, such as Sora, which rely on passive observation and thus fundamentally struggle with physical causality. This phenomenon motivates our core hypothesis: genuine physical intuition must \textit{emerge} from large-scale, causally-rich, real-world interactions.
Humans develop an understanding of intuitive physics through active interaction with the world. This approach is in stark contrast to current video models, such as Sora, which rely on passive observation and therefore struggle with grasping physical causality. This observation leads to our central hypothesis: \textit{authentic physical intuition of world model must be grounded in extensive, causally rich interactions with the real world.}
To test this hypothesis, we present \textbf{WoW}, a 14B-parameter generative world model trained on 2 million robot interaction trajectories. Our findings reveal that the model’s understanding of physics is a probabilistic distribution of plausible outcomes, leading to stochastic instabilities and physical hallucinations.
%, where unrealistic scenarios arise from ambiguous initial conditions.
%To test this hypothesis, we introduce \textbf{WoW}, a video diffusion model scaled up to 14B parameters and trained on a new dataset of 2 million real-world robot interaction trajectories. We find that the emergent understanding of physics from scaling is not a deterministic world model, but a probabilistic distribution of plausible outcomes, which leads to stochastic instabilities and physical hallucinations when generating from ambiguous initial conditions. 
% Our central finding is that an implicit understanding of physics---such as collision and occlusion---directly emerges with scale, in a trend consistent with Scaling Laws.
%
% Furthermore, this emergent capability is actively steerable through a novel cognitive loop, \textit{VLM-DiT}, where VLM agents evaluate the DiT's generated output and guide its refinement by iteratively evolving the language instruction. 
Furthermore, we demonstrate that this emergent capability can be actively constrained toward physical realism by \textit{SOPHIA}, where vision language model agents evaluate the DiT's generated output and guide its refinement by iteratively evolving the language instruction. 
Besides, a co-trained \textit{Inverse Dynamics Model} translates these refined plans into executable robotic actions, thus closing the imagination-to-action loop.
We establish \textbf{WoWBench}, a new benchmark focused on physical consistency and causal reasoning of video, where WoW achieves state-of-the-art performance of both human and autonomous evaluation, demonstrating strong ability on physical causality, collision dynamics, and object permanence.
Our work provides the systematic evidence that large-scale, real-world interaction is a cornerstone for developing physical intuition in AI. Models, data, and benchmarks will be open-sourced in \href{https://wow-world-model.github.io/}{wow-world-model.github.io}

\end{abstract}    

\tableofcontents
\newpage

\section{Introduction}
\label{sec:intro}

% % Visual Foundation Models Emerging Capabilities:
% % bagel - \citep{deng2025emergingpropertiesunifiedmultimodal}
% % DINO - \citep{caron2021emergingpropertiesselfsupervisedvision}
% % BLIP3 - \citep{xue2025xgenmmblip3familyopen}
% % DIFT - \citep{tang2023emergentcorrespondenceimagediffusion}
% % LUMA.Luma Dream Machine, 2024. \citep{}
% % GAIA autonomous driving \citep{hu2023gaia1generativeworldmodel}

% % SORA - \citep{sora} 
% % Learning an Intuitive Understanding of the physical world from a video? Start with an example. 
% Embodied world model, for dynamic environment modeling, generalization, and fast adaptation.
% % % Sample Efficiency with larger models and more data, Scaling Law:
% % emergent abilities in LLMs - \citep{wei2022emergentabilitieslargelanguage}
% % scaling law - \citep{kaplan2020scalinglawsneurallanguage}
\begin{quote}
    \textit{``The ladder of causation has three rungs: seeing, doing, and imagining.''}
    \par\noindent\hfill
    --- Judea Pearl \citep{bookofwhy}
\end{quote}
How does a human child acquire an understanding of the world? Not by passively observing videos, but through active and physical experimentation. The cognitive scientist Jean Piaget succinctly articulated this principle: \textit{``To know an object is to act on it'' \citep{piaget2013construction}}. This form of embodied learning, where countless 'actions' are intrinsically linked to immediate 'outcomes', is the foundational mechanism for mastering the laws of physics. This principle finds its direct computational instantiation in embodied world models, which are explicitly designed to learn a predictive model of how the world responds to an agent's actions \citep{hafner2019dream}.

In contrast, many recent advances in predictive models, particularly in video generation, is predicated on passive observation, a principle fundamentally distinct from active experimentation that fosters accurate causal understanding. While models like Sora \citep{sora2024} and others \citep{wan2025wan} achieve stunning photorealism and demonstrate emergent physical intuition, this intuition remains brittle. Their training objective prioritizes modeling statistical correlations from internet-scale data over inferring the underlying causal mechanisms of physics. Consequently, their grasp of physical laws is often superficial. When tasked with scenarios requiring genuine physical reasoning, they can produce logically and physically inconsistent outcomes. These models master the \textit{appearance} of our world, but the generative \textit{dynamics} they learn are an approximation rather than an accurate representation.
%bcy: robot interaction trajectories dataset inherently provides the closed-loop causal connection between an agent’s own actions and immediate outcomes—coupled with embodied sensory cues like tactile feedback and proprioception—that are indispensable for building an accurate understanding of physical causality and a robust world model. Passive observation, by contrast, cannot deliver this critical linkage, leaving models unable to grasp the fundamental insight of "how one’s own actions shape the world"—a gap that even the most photorealistic video generation models cannot overcome.

This distinction motivates our core hypothesis, \textbf{for an embodied model to develop genuine physical intuition, it must learn from large-scale, causally-rich, real-world interaction data, thereby lifting the generative model toward a world model} \citep{ha2018world, agarwal2025cosmospredict1}. To further validate our approach, we first introduce \textit{SOPHIA}, a novel architectural paradigm that couples the reasoning capabilities of a Vision Language Model (VLM) with the generative power of a Diffusion Transformer (DiT) \citep{dit}. We then present \textbf{WoW}, a concrete instantiation of this paradigm. WoW is a generative world model trained on a large-scale dataset of 2 million real-world robotic interaction trajectories, spanning 5275 tasks and 12 different robots. The objective of WoW is to directly synthesize pixel-level future predictions, learning to imagine and reason through generation itself.

To close the perception-to-action loop, we designed the Flow-Mask Inverse Dynamics Model (FM-IDM), which functions as the agent's equivalent of the cerebellum and motor cortex. By analyzing the optical flow and scene context between the current state and the imagined next state, the FM-IDM infers the 7-DoF end-effector action necessary to enact the transition. This module grounds the agent's imagination in physical reality, translating pixel-level futures into executable actions.

To empirically validate this complete perception-imagination-reflection-action cognitive architecture, we established \textbf{WoWBench}, a new benchmark focused on physical consistency and causal reasoning. WoWBench is composed of 4 core abilities and 20 sub-tasks, containing 606 samples, each with an initial image and a textual instruction. For a comprehensive evaluation, we comprehend 4 indispensable metrics: Video quality, Planning reasoning, Physical rules, and Instruction following. Our experiments demonstrate that our 14B-parameter WoW achieves SOTA performance on this benchmark, especially 96.53\% on Instruction understanding, and 80.16\% on Physical law, providing compelling evidence in support of our central hypothesis. To further verify the precision of our WoWBench, we conduct a human evaluation proving that WoWBench is highly correlated with human preference, and WoW achieves SOTA performance on both sides. Beyond its benchmark performance, WoW demonstrates versatility in broader applications. We show it is more than a simple generator, capable of enhancing VLM reasoning, serving as a physical simulator, and enabling 3D-aware representation learning.
%bcy: We showcase that it is more than a generator: it enhances VLM reasoning—Qwen-2.5-VL-7B-Instruct hit 89% planning and 44% task success in cube tasks via WoW’s feedback—serves as a physical simulator—14B WoW got 46.11% SOTA on WoWBench, WoW-Cosmos excelled in fluid/rigid/soft body simulations—and enables 3D-aware learning—its 4D model makes consistent novel views, ManipDreamer3D generates spatial coherent videos via 3D planning.

In summary, we propose \textit{SOPHIA}, a paradigm for developing embodied intelligence through a data-driven feedback loop. This approach involves deploying capable models to collect large-scale corrective feedback from physical interactions, a process that drives a continuous cycle of improvement. Our model, \textbf{WoW}, serves as a powerful instantiation of this paradigm, representing a significant advance from passive video models to an embodied world model that closes the perception-imagination-reflection-action loop. Our main contributions are as follows.
\begin{itemize}
    \item \textbf{A Unified Architecture for Imagination and Action.} We introduce an embodied world model \textbf{WoW}, which instantiates a novel self-optimization framework \textit{SOPHIA} for imagining physically plausible futures, and incorporates a Flow-Mask Inverse Dynamics Model that infers the corresponding executable actions.
    
    \item \textbf{Self-Supervised Feature Alignment.} We are the first to integrate powerful, pre-trained self-supervised visual features into the backbone of a diffusion-based world model. This novel approach significantly boosts the model's perceptual capabilities, accelerates training convergence, and improves the fidelity and physical consistency of generated futures.
    
    \item \textbf{An Interaction Benchmark.} We propose WoWBench, a new public benchmark designed to evaluate the physical consistency and action-generation capabilities of world models. Comprising 606 diverse, high-quality robot trajectories, the benchmark facilitates a rigorous performance comparison between existing methods and proposed WoW across 4 core metrics and 20 associated tasks.
    % 4 Materials, 4 group metrics including physical evaluation, 4 tasks across perception to planning,        
    % We propose and empirically validate a complete Perception–Planning–Prediction–Generalization cognitive architecture by introducing WoWBench, a benchmark focused on physical consistency and causal reasoning (4 core abilities, 20 sub-dimensions, 606 samples) with four evaluation metrics: video quality, planning reasoning, physical rules, and instruction following. And conduct a human evaluation experiment strongly correlates with WoWBench, confirming the benchmark’s validity and the method’s broad applicability.
    
    \item \textbf{Postraining Application Discussion.} We demonstrate that \textbf{WoW} is more than a generator, showcasing its versatility across a range of downstream applications. These applications include synthesizing novel views, generating trajectory-guided videos, producing action-conditioned videos for robot manipulation, enhancing visual style transfer and improving VLM task planning at test-time.

    \item \textbf{A Scaling Analysis and Open-Source Models.} We perform a systematic scaling analysis of our architecture up to 14B parameters, uncovering nascent capabilities and potential precursors to emergence in physical reasoning. This provides strong empirical evidence for the scaling hypothesis in this domain. We will release all trained model checkpoints to provide a foundation for future research in the embodied world model.

\end{itemize}
\section{Rethinking World Model: Towards A World-Omniscient Intelligent}

The idea that intelligent agents, whether biological or artificial, build internal models of their environment to predict, plan, and act has deep roots in cognitive science \citep{neisser1976cognition}. This notion has been instantiated in modern AI research, most notably in Ha and Schmidhuber's \textbf{world model}\citep{ha2018world}, and advanced by PlaNet \citep{hafner2018honglak} with recurrent state-space models for latent planning. The Dreamer series \citep{hafner2019dream} further extended this line by integrating imagination-based actor-critic learning, establishing world models as a scalable paradigm for long-horizon decision-making.

% This cognitive perspective naturally aligns with model-based reinforcement learning (RL), where agents explicitly learn predictive models to support planning and decision-making. The influential Dreamer series \citep{hafner2019dream} further advanced this paradigm by demonstrating how large-scale predictive models can guide complex control tasks through imagination-based learning.
In what follows, we provide a formal definition of world models~\cref{sec:pre_define}, trace their evolution and recent applications in embodied AI~\cref{sec:pre_evol}, and highlight their critical role in the broader development of generative AI~\cref{sec:pre_ewm}.
% TODO update
Furthermore, we discuss the relationships among video generation, large pretrained models, and latent-space world models. This analysis provides the rationale for our proposed approach: \textbf{structuring world models by effectively leveraging pretrained foundation models and offering insights into future architectural designs}.

\subsection{Definition of a World Model}
\label{sec:pre_define}
A world model is an internal, learned representation of the dynamics of an environment, designed to simulate or predict future states based on current states and potential actions~\citep{ha2018world}. Formally, given a state $s_t \in \mathcal{S}$, a low-level control action $a_t \in \mathcal{A}$ at timestep $t$, and $p_t \in \mathcal{P}$ is the meta-level strategy/plan at time $t$, the world model learns a transition function $f_\theta$ parameterized by $\theta$. This function aims to predict the subsequent state $s_{t+1}$.

The prediction can be modeled deterministically as:
\begin{equation}
s_{t+1} = f_\theta(s_t, a_t, p_t) \quad with \quad a_t \sim \pi_{\phi}(a_t | s_t, p_t), \quad p_t \sim \pi_{\omega}(p_t | s_t, H_t)
\end{equation}

where $\pi_{\phi}$ refers to low-level policy and $\pi_{\omega}$ refers to high-level policy. The term $H_{t}=(s_{t-h:t}, a_{t_h:t}, p_{t-h, t})$ is the historical context up to time t, and $h$ is the recall horizon. 

or probabilistically, as:

\begin{equation}
s_{t+1} \sim \mathbb{P}_\theta(s_{t+1} \mid s_t, a_t, p_t) 
\end{equation}

To handle high-dimensional sensory inputs, such as images, modern implementations typically operate within a compressed latent space. Encoders map observations $o_t$ of state $s_{t}$ to a low-dimensional latent state $z_t$. The transition model then operates in this abstract space:

\begin{equation}
z_{t+1} \approx f_\theta(z_t, a_t, p_t) \quad with \quad z_t = Encoder(o_t)
\end{equation}

The core training objective is to minimize prediction error over a transition dataset $\mathcal{D} = \{(o_t, a_t, o_{t+1})\}$, typically using a loss function such as Mean Squared Error (MSE):

\begin{equation}
\label{eq:world_model_loss}
\min_{\theta} \mathcal{L}_{\text{trans}}(\theta) = \mathbb{E}_{(z_t, a_t, z_{t+1}) \sim \mathcal{D}} \left[ \left\| f_{\theta}(z_t, a_t) - z_{t+1} \right\|^2 \right]
\end{equation}

This objective compels the model to internalize physical laws, object permanence, and causal relationships that govern environment dynamics \citep{hafner2024masteringdiversedomainsworlddreamv3}.

% \begin{figure}[ht]
%   \centering
%   \includegraphics[width=0.95\linewidth]{figs/modality.pdf}
%   \caption{Illustration of the developmental trajectory of world models, showing an initial phase of modal amplification characterized by diverse modality-specific models (e.g., VGM, LLM, Latent WM, DINOWM), followed by a critical emergence point that marks a rapid increase in world modeling ability. This leads into the paradigm convergence period, where unified and highly capable world models emerge. The x-axis represents progress in world model development across modalities (language, audio, vision, etc.), while the y-axis reflects the model's overall ability to represent and reason about the world.}
%   \label{fig:WoW-modality}
% \end{figure}

\subsection{Evolution of World Models: Representation and Modality}
\label{sec:pre_evol}
% 1.1. **RSSM 潜在动力学**：在潜空间做长时“想象”，样本效率高；Dreamer 系列代表。
%   1.2. **Transformer-SSM**：用注意力替代 RNN，擅长长程依赖与并行；TransDreamer/TWM/Genie 等。
% 3. 提一嘴就好 **JEPA 预测式表征**：不重建、做表征匹配；I-JEPA/V-JEPA 适合大规模自监督预训练。
% 4. 主要讨论这个 **生成式（AR 与 Diffusion）**：
%     - **自回归**：时序因果强、可多模态条件，但易因离散量化丢细节；CogVideo/NUWA/VideoPoet、GAIA-1、OccWorld。
%     - **扩散**：训练稳、视觉/时空质量高、可潜空间高效生成；Sora/Veo3、DriveDreamer/Vista/GAIA-
%  我们做的是生成式的去approach一个完整的SSM。 生成式的部分要考虑模态扩增的问题，逐渐expand到全模态，现在在visual。

Initially, world models were primarily used in model-based reinforcement learning (RL) as compact latent representations. Pioneering this line, World Models \citep{ha2018world} coupled a variational autoencoder (VAE) with an Mixture Density Network Recurrent Network (MDN-RNN) to learn a generative internal model from pixels and showed that a simple controller trained entirely “in the dream” of the model can transfer back to the real environment, establishing the feasibility and sample-efficiency of learning inside a learned simulator. Building on this idea with a stronger recurrent state-space model(RSSM), PlaNet \citep{hafner2018honglak} performed fast online planning in latent space via a stochastic-deterministic dynamics model trained with multi-step variational objectives, enabling image-based control with substantially improved data efficiency. The Dreamer family \citep{hafner2019dream} then replaced step-wise planning with differentiable actor-critic learning inside imagination: Dreamer back propagates value gradients through imagined rollouts in a learned world model; DreamerV2 adds discrete latents to reach human-level Atari; and DreamerV3 standardizes robust training to solve 150+ tasks under one configuration, making “learning in the dream” a scalable general RL approach.

Like RSSMs, Joint-Embedding Predictive Architectures(JEPAs)~\citep{assran2023self} also learn abstract latent representations, but JEPA optimizes predictive agreement in embedding space rather than modeling pixel-level likelihoods or explicit dynamics. Concretely, given a context embedding, a predictor is trained to match a stop-gradient target encoder on masked views of the same scene, inducing semantic structure without pixel reconstruction or heavy augmentations. This objective scales naturally and encourages invariant/equivariant features that capture object- and scene-level regularities. Extending to video, V-JEPA \citep{bardes2024revisiting}  performs in-context prediction of masked spatio temporal chunks in latent space; V-JEPA-2 \citep{assran2025vjepa} further shows that pretraining on web video, augmented with a small amount of interaction data, yields world-model–like priors that enable zero-shot robotic planning directly from images. Although these models are not full-fledged world models in the control sense, they highlight a predictive representation learning path toward scalable priors for embodied reasoning.

% \citep{yan2021videogptvideogenerationusing}

The above world models primarily encode an implicit understanding of the external world in latent space. In February 2024, OpenAI introduced Sora \citep{sora2024}, widely regarded as a video world model that markedly advances environment simulation and prediction. Broadly, video world models fall into two complementary families.  Inspired by breakthroughs in large language models,  Autoregressive (AR) approaches tokenize images/videos and learn causal next-token dynamics, which makes online rollouts and tight policy coupling natural \citep{hu2023gaia, bruce2024genie}. In contrast, diffusion-based approaches model conditional video generation via iterative denoising, better capturing multi-modality and stochasticity, and can fold planning into generation through trajectory diffusion \citep{yang2023unisim} . Despite their promise, current video world models face significant limitations: they often lack 3D consistency, physical coherence, and temporal reasoning required for faithful environment simulation. Addressing these challenges remains central to advancing world models toward general-purpose embodied intelligence.

\begin{wrapfigure}{l}{0.60\textwidth}
  \centering
  \vspace{-15pt}
  \includegraphics[width=0.95\linewidth]{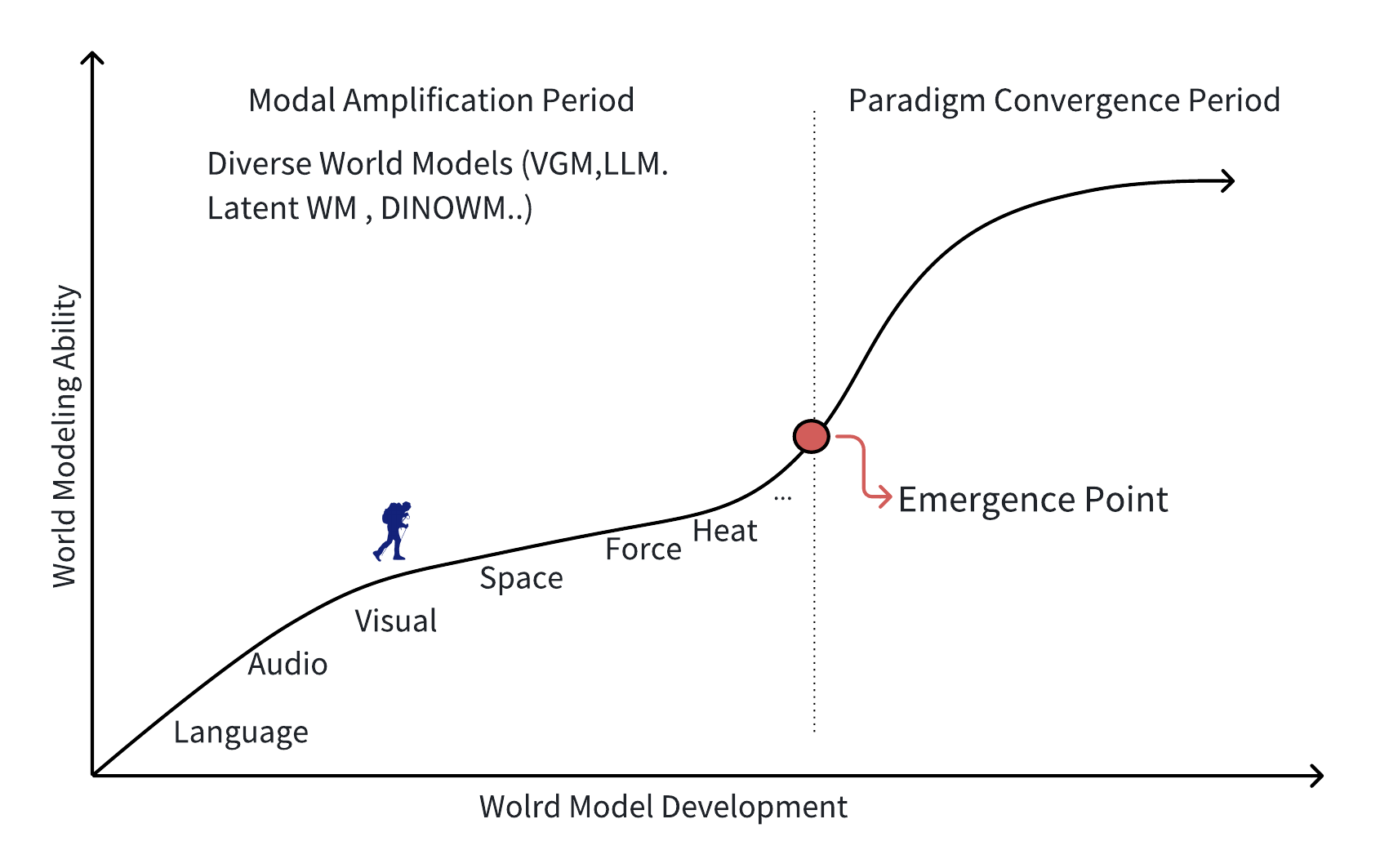}
  \caption{Developmental trajectory of world models, from modality-specific models (e.g., VGM, LLM) to unified models after a critical emergence point.}
  \label{fig:WoW-modality}
  \vspace{-10pt}
\end{wrapfigure}

The convergence of these modalities—from model-based RL to large language and video generation models—marks the current era of multi-modal expansion. While each approach has its strengths, they all currently lack the dynamic interactive properties and the comprehensive world understanding required for a true general-purpose world model. The models have diverse architectures, training methods, and paradigms.

As AI continues to exhaustively integrate more modalities, the development of a unified world model will become a critical component for AIs to explore and interact with the real world. This will, in turn, lead to more mature and robust AI architectures and paradigms as shown in Figure \ref{fig:WoW-modality}.

% 发展到多模态的输入输出，空间模态 （action，3D坐标）

\subsection{Embodied World Modeling}
\label{sec:pre_ewm}

% 1. 具身的世界模型和视频生成的区别？
%       模态扩增，物理规律，3D consistency, env grounded
% 2. 具身的世界模型和传统的model-based RL里的世界模型？
%    reward model 和 RL reward （model-based RL?）
%    像素级的生成对于emboded来说有什么用？ (可解释)
%    embodied world model + VLM
%    post-training + world model
%    goal image + key frame planning
%    latent和像素级生成？
% 新开一个section
%  3. 世界模型怎么帮助VLA （要在哪里讨论）
%    data gen, policy training, policy search, planning + IDM, policy evaluation
%  7or8讨论一下application for VLA

% 引出我们对这个工作的 formulation, force, heat? 
% 和视频生成有什么区别 怎么连上
% 相比vjepa我们认为像素级别重建生成是有必要的
% 世界模型要从单纯生成式逐渐过渡到dreamer那样的能够与真实世界动态交互的模型

% An \textbf{Embodied World Model} extends the foundational world model concept by imposing three critical requirements tailored for physical agents: a generalized control interface, a physically coherent simulation output, and a commonsense-infused internal representation.

The embodied world model~\citep{yang2023unisim, azzolini2025cosmosreasoning1} stands as the next Holy Grail, a system that demands long-term, multimodal interactive modeling to reach its full potential. Yet today, several critical limitations remain: unresolved challenges with spatial and temporal consistency, the absence of a universal general control interface, and significant gaps in physical understanding and common sense.

% memory: temporal consistency;
% 3D / spatial consistency
% general control
% physical understanding
% common sense: 颜色，方位，材料这类认知；VLA survey； 也得有vlm common sense (A Survey on Vision-Language-Action Models for Embodied AI)

\noindent\textbf{Spatial Consistency}
Our world is inherently three-dimensional, where every object and event possesses intrinsic 3D properties that define their existence. In contrast, conventional video generation models prioritize only the visual effects perceivable by the human eye, neglecting the critical underlying requirement of 3D consistency.
Embodied world models that are designed to simulate real environments require that their outputs be strictly 3D-consistent~\citep{zhen2025tesseractlearning4dembodied}.

\noindent\textbf{Temporal Consistency}
Severe memory constraints pose a core challenge for embodied world models, limiting the encoding of the world's state to a single image or discrete tokens. This fundamentally hinders their ability to reliably simulate complete 3D environments where much of the world remains unobserved.
% A core challenge persists for embodied world models: the world’s underlying state is only encoded as a single image or a discrete set of image chunks due to memory constraints. This limitation inhibits existing world models from reliably simulating complete 3D environments—spaces where a large portion of the generated world remains unobserved at any particular moment.
Consequently, such models suffer from short-sightedness: they often produce new elements that contradict their own prior context. This inconsistency directly compromises the internal coherence that is critical for achieving real-world applications. Such contradictions further weaken the models’ capacity to monitor environments as they evolve and to formulate plans across extended time horizons. In the end, this limits their effectiveness for tasks that depend on stable, long-horizon interaction. Recent work ~\citep{hong2024slowfast} utilizes a temporary LoRA module that embeds the episodic memory in its parameters in video diffusion models, while other works~\citep{chai2023persistent, zhou2025learning, lin2025perceive} integrate an explicit memory architecture that enables consistent long-horizon imagination, ensuring coherence with previously generated content.

% \paragraph{General Control Interface}

%Most world models are only based on conditional inputs of a single modality. However, we believe that embodied world models in their ultimate form can accept conditional inputs $c$ of any modality (such as text, images, actions, and sounds), and this can make the transition models more versatile:
%\begin{equation}
%z_{t+1} \approx f_\theta(z_t, c)
%\end{equation}
%For robotics, embodied world models are specially required to accept robotic actions as conditional inputs to predict future states. How to model the actions from a robot's physical actuators is a significant problem.

% \paragraph{Generalized (general?) Control Interface}
% First, the input must generalize beyond a simple action vector $a_t$ to a multimodal control signal $c_t$. This allows the model to respond to high-level instructions, such as text prompts ($c_{\text{text}}$), robot action, or goal images ($c_{\text{img}}$), making the transition model more versatile:
% \begin{equation}
% z_{t+1} \approx f_\theta(z_t, c_t)
% \end{equation}
% While works like GENIE learn a latent control space from video \citep{reed2024genie}, a significant methodological gap remains in mapping such abstract signals to a robot's physical actuators.

\noindent\textbf{Physically Coherent Simulation}
The understanding of physical laws is a crucial challenge for AGI~\citep{chow2025physbench}, and physical understanding becomes particularly important when we need AGI to interact with the real world through physical actions~\citep{bansal2025videophy2challengingactioncentricphysical}. We require embodied world models to predict and envision what would happen if we take this action. For instance, an embodied world model needs to know that a cup will break if it rolls off the table and onto the ground, which is why the cup must be placed stably on the table. Multimodal large models enhance their physical understanding capabilities by constructing more physical data. However, how to improve the physical understanding of embodied world models remains an unsolved problem~\citep{motamed2025generative}. Currently, we believe we may need to explicitly enforce physical consistency; yet, a latent prediction model alone cannot explicitly define physical understanding. We have also found that enhancing the description of physical laws in conditional prompts can significantly improve the physical consistency of embodied world models, thereby greatly increasing the possibility of embodied world models becoming a physically coherent simulation.
% key challenge
% rewrite
% Second, the model's output must be a physically plausible simulation, not merely a latent prediction. This demands a decoder $D_\phi$ and a more comprehensive objective function that explicitly enforces physical consistency. This conceptual loss extends beyond the basic transition loss (Eq. \ref{eq:world_model_loss}):
% \begin{equation}
% \min_{\theta, \phi} \mathcal{L}_{\text{total}} = \mathcal{L}_{\text{trans}} + \lambda_1 \mathcal{L}_{\text{recon}} + \lambda_2 \mathcal{L}_{\text{physics}}
% \end{equation}
% Here, $\mathcal{L}_{\text{physics}}$ represents a term penalizing violations of physical laws. The absence of such a term is a key limitation in current models, leading to long-horizon prediction failures as documented by benchmarks like Physion \citep{bear2021physion}, even in state-of-the-art systems like GAIA-1 \citep{wayve2024gaia}.

\noindent\textbf{Commonsense-Infused Representation}
Common sense capability is one of the core hallmarks that enables Large Language Models (LLMs) to move beyond "accurate text generation" and toward "truly understanding the world"~\citep{zhao2023large}.
Similarly, the internal representations of world models must be infused with "world commonsense", encompassing capacities for causal and logical reasoning, alongside the generalization and transfer of world knowledge~\citep{bansal2024videophyevaluatingphysicalcommonsense}.
This foundational requirement can only be achieved through large-scale pre-training on diverse datasets (\(\mathcal{D}_{\text{pretrain}}\)). Yet a critical, unresolved challenge remains: bridging the domain gap between web-scale, third-person data and the distinct first-person, embodied perspective of a physical agent.

% Similarly, the internal representation of world models must be endowed with "world commonsense" through large-scale pre-training on diverse datasets ($\mathcal{D}_{\text{pretrain}}$). Methods like V-JEPA achieve this by learning powerful, abstract features from video without costly pixel-level generation \citep{boididou2024vjepa}. Regardless of the approach—be it generative like GENIE \citep{reed2024genie} or predictive like V-JEPA—the decisive challenge remains to bridge the domain gap between web-scale, third-person data and the unique first-person, embodied perspective of a physical agent.

\subsection{Emergency, Toward a Generative Physical Engine}
% 训练生成式世界模型的下一阶段目标，就是一个帮助具身落地的物理仿真引擎
% 为啥需要pixel 重建，而不是VJEPA
% Genesis comparison： 阶段性替代，引入3D模型，引入模拟器；纯生成比genesis还要再远一点
% 目标就是 data scaling，解决所有问题。

\begin{figure}[ht]
  \centering
  \includegraphics[width=0.95\linewidth]{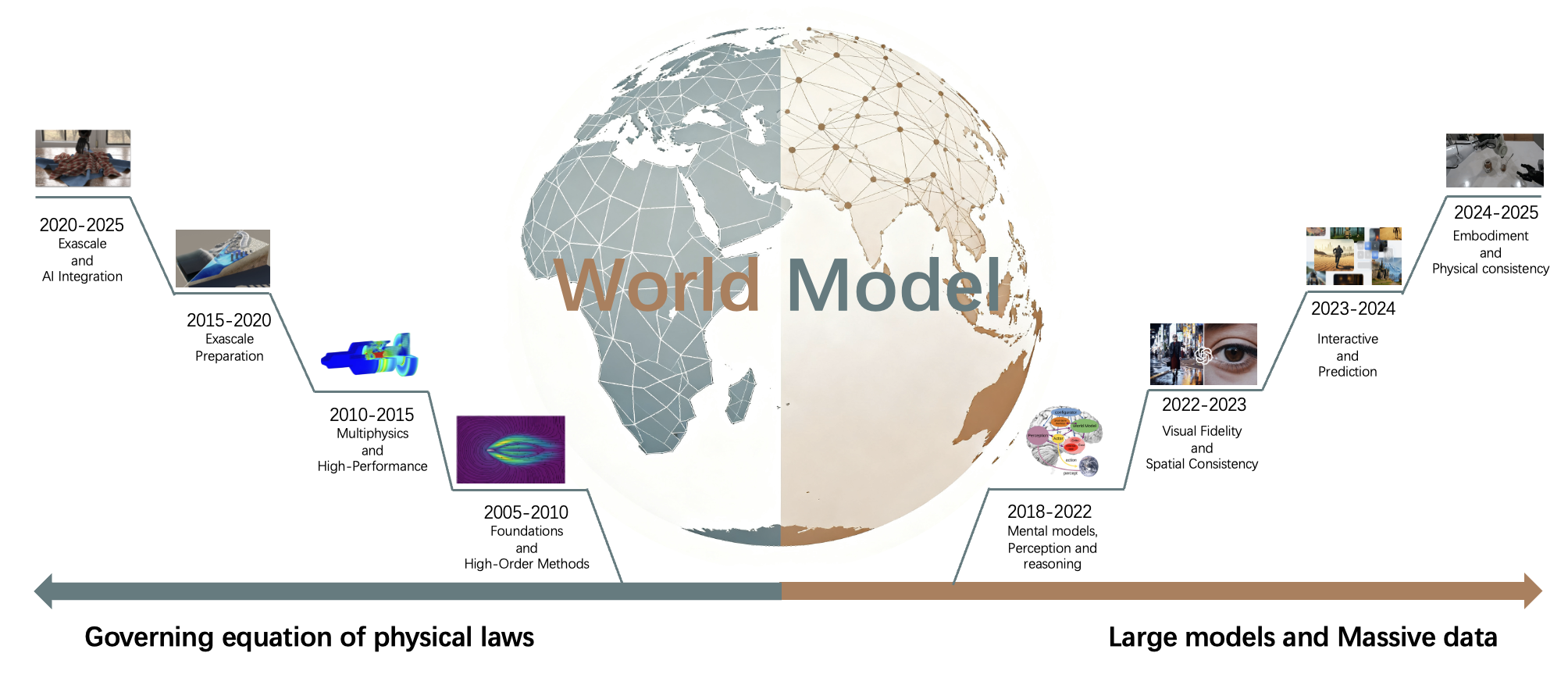}
  \caption{The technological development of world models in pursuit of intrinsic physical consistency has primarily followed two approaches. One possible approach is to construct a world model with inherent physical modeling capabilities for robots. The mainstream approaches to realizing such a world model include two methods. One is based on generative AI combined with a differentiable physics engine. The other, grounded in video generation models, constructs a neural network-driven physics engine that possesses both intrinsic physical consistency and external high visual fidelity.}
  \label{fig:physical_consistency}
\end{figure}

Like Richard Feynman said, "\textbf{\textit{Physics is the foundation of all the natural sciences, and without it, our modern world could not exist.}}" Therefore, embodied intelligent robots must model the underlying physical laws of the world to perform various tasks in the real world better. One possible approach is to construct a world model with inherent physical modeling capabilities for robots. The mainstream approaches to realizing such a world model include two methods. One is based on generative AI combined with a differentiable physics engine. The other, grounded in video generation models, constructs a neural network-driven physics engine that possesses both intrinsic physical consistency and external high visual fidelity, as can be seen in Figure ~\ref{fig:physical_consistency}.  

For the first approach, the differentiable physics engine is derived from the numerical simulations. In the path of numerical simulation, the understanding of physics is relatively straightforward: one needs to transform the governing equations of various physical domains from the continuous to the discrete domain, enabling solutions to be computationally solved. Typical methods, e.g., the finite-volume method (FVM) and the finite-element method, are adopted as general-purpose simulation frameworks and are emphasized on high-resolution and high-order methods for accuracy. When single-domain problems are solved, the demand for coupled multiphysics simulations (e.g., fluid structure, electrochemistry, climate) grows, driven by rapid hardware changes. Multicore CPUs became ubiquitous, and early CUDA/GPU programming proved effective for many Partial Differential Equation (PDE) solvers. Extended FEM (XFEM) for cracks and interface problems, and phase-field models for fractures, appeared in the literature \citep{kirchhart2016analysis}. 

Then, researchers began fusing data science with traditional simulation – e.g., using reduced-order models, adjoint‐based optimization, and surrogates to accelerate design. Emphasis was placed on modular, open frameworks for coupling new algorithms (including machine learning components) into solvers. Smoothed Particle Hydrodynamics (SPH) became increasingly used in commercial and research codes for fluid–solid interaction \citep{Violeau02012016}. Exascale machines and the integration of AI and data science became the core of simulation methods. Researchers tackle uncertainty quantification, digital twins, and inverse design as first-class objectives. Novel algorithms (quantum, neural PDE solvers) began to be explored, though classical methods continue evolving for maximum performance. Differentiable simulation is a significant innovation in the integration of automatic differentiation into FEM. The JAX-CPFEM example demonstrates this trend: by making the FEM pipeline differentiable, designers can use gradient-based optimization on geometry and material parameters \citep{hu2025efficient}. Genesis re-designed the physics engine with generative AI. It integrated multiple classic physics solvers, such as rigid body, FEM, and SPH, as mentioned above, making the instruction easier to use ~\citep{Genesis}. \textbf{While these systems are also termed "world models" and their physical understanding relies on numerical solution of governing equations, they appear to overemphasize the role of world models as a simulator. Moreover, they neglect other critical functions of "world models" in the cognitive aspect.}  

In the other approach, the core proposition of "world models" was distilled into a simple idea: for an agent to efficiently trial-and-error in the real world, it first needs to "dream" in its mind, that is, predict the future, test actions, and bring promising strategies back to reality within a manipulable latent environment \citep{ha2018world}. Diffusion transformer-based models have been able to achieve high visual fidelity and spatiotemporal consistency. OpenAI's Sora raised the bar for "long duration, camera motion, and multi-agent interaction," explicitly proposing a "world simulator" vision in public materials: not just generating "look-alike" clips, but maintaining interpretable 3D scenes and causal continuity. However, in engineering practice, the pursuit of visual fidelity often sacrifices physical consistency and action controllability \citep{sora2024}. Meanwhile, approaches like Genie, which "generate interactive environments," focus on the other end: not prioritizing high-resolution video but first ensuring the environment is action-controllable, temporally consistent, and playable, then gradually improving rendering quality; this aligns better with developers' "ease of use—cost" objective function in game engine scenarios \citep{bruce2024genie}.  

The next frontier for world models is to evolve into a Generative Physical Engine. This engine would go beyond abstract representations to simulate real-world dynamics, including fluid flow, material properties, and collisions. This capability is essential for building a robust and consistent foundation of common-sense knowledge and schemata within the model. And the model must satisfy the following:

\textbf{Causal Inference}: The model must understand how actions lead to specific outcomes. This allows for safe and effective planning in dynamic environments.

\textbf{Physical Understanding}: The model needs to encode the core concepts of motion, force, and spatial structure. This deep understanding of physical laws allows for more robust and adaptive behaviors.

\textbf{Interaction}: A generative physical engine enables agents to test actions in a simulated environment, allowing efficient training and exploration without the risks of real-world trial and error.

\textbf{Consistency}: The simulations generated must be temporally and physically consistent and free from the unrealistic artifacts that often plague current generative models.

\subsection{The Cognitive Science Connection: World Model in Mind}
\begin{figure}[ht]
  \centering
  \includegraphics[width=0.8\linewidth]{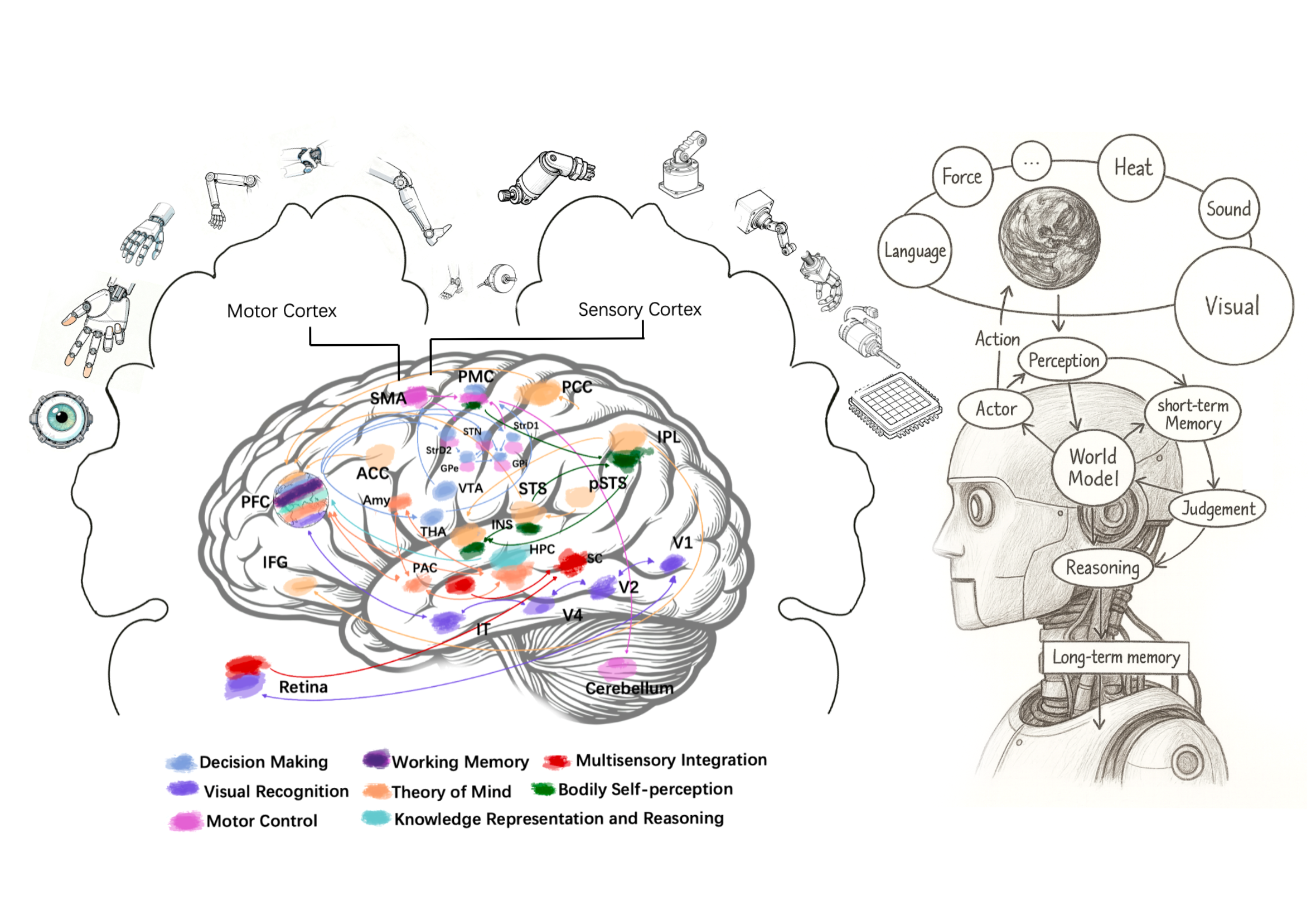}
  \caption{The architecture of an embodied agent with a world model. An intelligent agent perceives the environment through various sensory inputs (e.g., visual, sound, heat, force). These perceptions are processed by a World Model, which builds an internal, predictive representation of the environment. The model's predictions and past experiences, stored in short-term and long-term memory, inform Reasoning and Judgement. Based on this internal simulation, the Actor generates Actions that manipulate the real world. This closed-loop system allows the agent to learn the dynamics of its environment, plan for the future, and achieve complex goals. (Figure inspired by \citep{BrainCog})}
  \label{fig:WoW-robobrain}
\end{figure}

% 从认知科学的角度来claim 具身世界模型在AGI中的重要性
% Cortical homunculus 皮质小人中，大脑皮层感知中最活跃的部分1是头（眼睛嘴巴..)，2就是手；
% 构建知觉的世界模型，在语言/视觉/听觉模态之后，通用世界模型的下一步可能也是先让AI长出手；
% 即具身的世界模型符合Naisser的知觉循环；然后Schemata应该就是模块化的，包括世界模型的一个认知框架
 The pursuit of an embodied world model represents a significant bridge between abstract computation and physical experience, drawing a direct parallel to human cognition. This "intuitive physics" mirrors our own mental processes, which allow us to simulate and reason about the world without direct interaction. This ability is a cornerstone of human intelligence, enabling us to make predictions and plan actions based on an internal model of reality.

% This convergence of world models, generative AI, and embodiment marks a paradigm shift. It moves AI toward more natural human-AI interaction by grounding abstract decisions in perceptible, physical behaviors. By completing the closed loop of Schema-Action-Observation, we move a step closer to creating a truly intelligent, general-purpose agent that can reason, learn, and adapt in a manner akin to a human cognitive system.

In AI research, world models are usually utilized to simulate or predict the environment via generative models. For example, in an agentic system, the world model behaves like a simulation environment for the agent to explore and learn by interacting with the environment. Although these works declared that they are "embodied" since the agent is usually a robot. \textbf{However, we believe that relying solely on simulation or prediction falls short of embodying the cognitive sense. The contradiction lies in their consideration of the world model, which represents the external environment and the agent's physical state with it. Still, the world model as part of the mind model refers to an internal representation of mental states that captures regularities of the world.} As shown in Figure 4, the hippocampus combines the Theory of Mind and Knowledge Representation and Reasoning, suggesting that our combination of DiT and VLM as a system can fulfill the hippocampus's primary function. 

Moreover, modern neuroscience considers the brain as a distributed system, where cognition arises from the coordination of specialized but interconnected networks. Mental models are not the product of any single brain region, but rather the result of complex networks formed by multiple brain regions working in concert. Mental models are not the product of any single brain region, but rather the result of complex networks formed by multiple brain regions working in concert.

% 可以参考的与method部分的呼应：
% 1. world model不是具体的一个脑区反馈，而是脑子各个部分的协调工作（对应多个模块组件）
% 2. 对应设计理论是Nasser的知觉反馈；认知图谱-》感知-》观测-》认知图谱迭代 （对应组件的具体设计）

Inspired by this view, we propose an embodied world model that integrates the world model with both sensory processing and motor abilities. Within this framework, schemata serve as modular cognitive structures that encompass perception, prediction, reflection, and action. Our approach grounds intelligence in embodied interaction, offering a pathway toward more flexible and generalizable cognition. It can therefore dynamically interact with the physical world.

\definecolor{envcolor}{RGB}{204,153,0}       % Deep orange-yellow for environment
\definecolor{robotcolor}{RGB}{0,128,0}       % Dark green for robot setup
\definecolor{actioncolor}{RGB}{178,34,34}    % Firebrick red for actions
\definecolor{statecolor}{RGB}{25,25,112}     % Midnight blue for final state

\section{WoW World Model}
% Word2Vec

% In this section, we introduce \textbf{WoW}, a world model designed to explicitly encode complex physical laws and enhance reasoning through a novel mechanism called \textit{Reflection of Generation}~\citep{chi2024eva}, as illustrated in Figure~\cref{fig:WoW-model0}.
% We begin by highlighting the importance of the language modality in Section~\cref{sec:model_theory}. Inspired by Neisser’s \textit{Perceptual Cycle}~\citep{neisser1976cognition}—\textit{Schemata} $\rightarrow$ \textit{Perception} $\rightarrow$ \textit{Action} $\rightarrow$ \textit{Schemata}—our framework integrates three key stages:

% \begin{itemize}
%   \item \textbf{Task Imagination (Schemata):} A diffusion-transformer-based module that generates high-level abstract plans and pixel-level future predictions (see Section~\cref{sec:model_vgm}).
%   \item \textbf{Experience Reflection (Perception):} A vision-language agent system that reinforces physical consistency by reflecting on model outputs (see Section~\cref{sec:model_agent}).
%   \item \textbf{Behavior Extraction (Action):} A test-time scaling and translation module that derives executable policies from visual imaginations (see Section~\cref{sec:model_idm}).
% \end{itemize}

% This closed-loop agent architecture fully leverages the capabilities of pre-trained models. \textbf{WoW} advances the current paradigm of world modeling by unifying \textit{reflection} and \textit{imagination} as core components of intelligent behavior.
We present WoW, an embodied world model built upon Self-Optimizing Predictive Hallucination Improving Agent (\textit{SOPHIA}), our novel paradigm for enhancing physical reasoning through closed-loop self-refinement.
%Inspired by Neisser’s Perceptual Cycle~\citep{neisser1976cognition}—\textit{Schemata} $\rightarrow$ \textit{Perception} $\rightarrow$ \textit{Action} $\rightarrow$ \textit{Schemata}—WoW organizes intelligent behavior into three interconnected stages:
%\textbf{Task Imagination (Schemata):} A novel SOPHIA module that generates high-level plans and pixel-level futures(see~\cref{sec:model_vgm}).
Inspired by Neisser’s Perceptual Cycle~\citep{neisser1976cognition}—\textit{Schemata} $\rightarrow$ \textit{Perception} $\rightarrow$ \textit{Action} $\rightarrow$ \textit{Schemata}, WoW organizes intelligent behavior into three interconnected stages.
\begin{itemize}
    \item \textbf{Task Imagination (Schemata):} We introduce the novel \textit{SOPHIA} paradigm for generating high-level plans and pixel-level future predictions, and our instantiated \textbf{WoW} model. in Section~\ref{sec:model_vgm}.
    
    \item \textbf{Experience Reflection (Perception):} A VLM agent verifies physical consistency and iteratively refines the imagined outcomes (see Section ~\ref{subsec:solver_critic_agent}).
    
    \item \textbf{Behavior Extraction (Action):} A test-time module that translates imagined trajectories into executable policies (see Section~\ref{sec:model_idm}).
\end{itemize}

\section{Self-Optimizing Framework}
\begin{figure*}[t]
  \centering
  \includegraphics[width=0.95\linewidth]{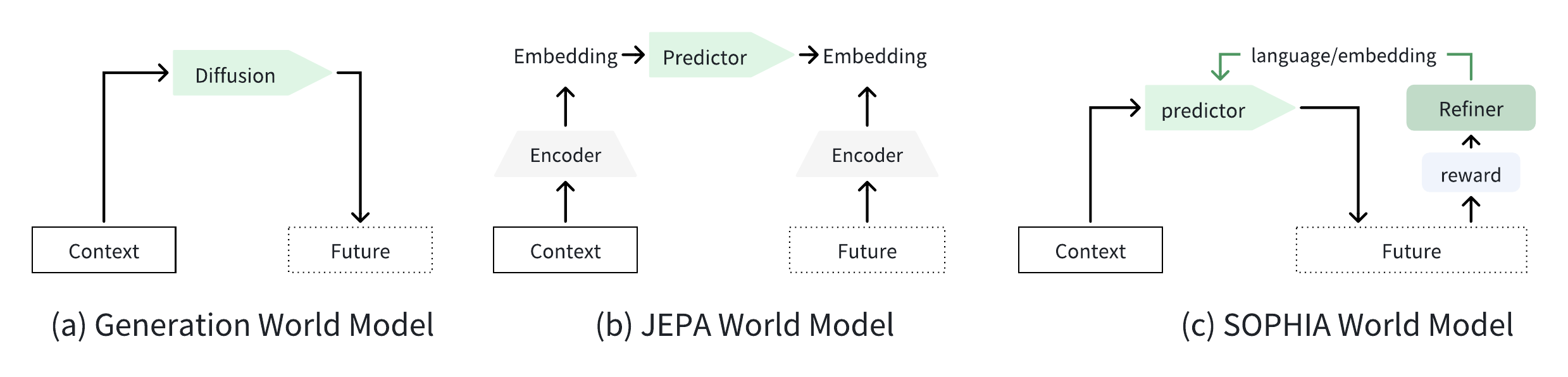}
  \vspace{-10pt}
  \caption{\textbf{Comparison of Diffusion, JEPA~\citep{assran2025vjepa}, and SOPHIA.}
The \textbf{Predictor} generates a \textbf{Future} from the input \textbf{Context}. This outcome is then evaluated to produce a \textbf{reward}, which directs the \textbf{Refiner}. Finally, the \textbf{Refiner} leverages this reward and external \textbf{language/embedding} guidance to issue a corrective signal, iteratively improving the next prediction cycle. }
  \label{fig:WoW-model0}
  \vspace{-1em}
\end{figure*}

At its core, WoW follows \textit{SOPHIA} paradigm that integrates large language models with diffusion transformers to generate physically plausible futures under language guidance, as shown in Figure~\ref{fig:WoW-model0}. 
Through this iterative cycle of \textit{predict}, \textit{critic}, and \textit{refine}, WoW unifies imagination and reasoning as fundamental components of embodied intelligence. 

SOPHIA improves its physical plausibility based on the following hypothesis: Empirically, we observe that more detailed language prompts lead to video generations that are both physically plausible and semantically precise. To formalize this intuition, we adopt the theoretical framework proposed by \textit{'Critical of World Model'}~\citep{xing2025critiqueswm}, and present the following result:

\textbf{Hypothesis 1 (Completeness of Language Representation).} \textit{
Let $\mathbf{x} = \{x_t\}_{t=1}^T$ be a continuous input sequence with $x_t \in \mathbb{R}^D$ and $\|x_t\| < K$. For any $\epsilon > 0$, there exists a language system $L_\epsilon = (V, N, f_\epsilon)$ with vocabulary $V$, sentence length $N < \infty$, and mapping $f_\epsilon: \mathbb{R}^{T \times D} \rightarrow V^N$, such that for any $\mathbf{x}, \mathbf{x}'$, if $\|\mathbf{x} - \mathbf{x}'\| \geq \epsilon$, then $f_\epsilon(\mathbf{x}) \neq f_\epsilon(\mathbf{x}')$.
}

This hypothesis assumes that language, when sufficiently expressive, can uniquely distinguish between arbitrarily similar physical sequences. The WoW allows abstracting complex dynamics (e.g., post-collision motion) into symbolic descriptions, enabling fine-grained control via language.

In the context of video diffusion, $\mathbf{x}$ represents a video segment at the pixel-level, while $f_\epsilon(\mathbf{x})$ denotes its corresponding prompt. 

%Two practical strategies follow:

%\textbf{(1) Scaling Up (Larger Vocabulary).} Fixing sequence length $T$, the required vocabulary size grows as:
%\begin{equation}
%    M_\epsilon \approx \left\lceil (\sqrt{TDK})\epsilon^{-1} \right\rceil^D
%\end{equation}

%\textbf{(2) Scaling Out (Longer Sentences).} Fixing vocabulary size $M$, longer sentences improve resolution:
%\begin{equation}
%N_\epsilon \approx TD \left\lceil \log_M \left( \sqrt{TDK} \epsilon^{-1} \right) \right\rceil
%\end{equation}

%These bounds explain an empirical trend: more expressive prompts lead to more accurate and controllable generations. 
%This theoretical foundation bridges symbolic language and sub-symbolic visual dynamics, underpinning WoW's design: leveraging language not merely as a command interface, but as a structured and semantically rich representation of physical reality.
%This theoretical foundation bridges symbolic language and sub-symbolic video modeling, supporting WoW's design: using language not only as instruction, but as a precise representation of physical reality.

\subsection{Foundation Video Generation World Model}
\label{sec:model_vgm}

This section serves as the perceptual and imaginative engine of WoW, enabling the agent to simulate future dynamics from language instructions. This module is built upon three key components: (1) \textbf{Pretrain Data Preparation}, which constructs a large-scale dataset of language-conditioned physical interactions to support multimodal learning; (2) \textbf{Diffusion-Based Video Generation}, which leverages the DiT architecture to generate high-fidelity, temporally coherent video rollouts conditioned on task descriptions; and (3) \textbf{Solver-Critic Video Generation Agents}, which introduce an iterative refinement loop—where a solver generates plausible futures and a critic evaluates their physical consistency—enabling closed-loop imagination with self-correction. Together, these components allow the model to not only imagine what could happen, but to refine these imaginations toward physically plausible outcomes.
\begin{figure*}[t]
  \centering
  \includegraphics[width=0.95\linewidth]{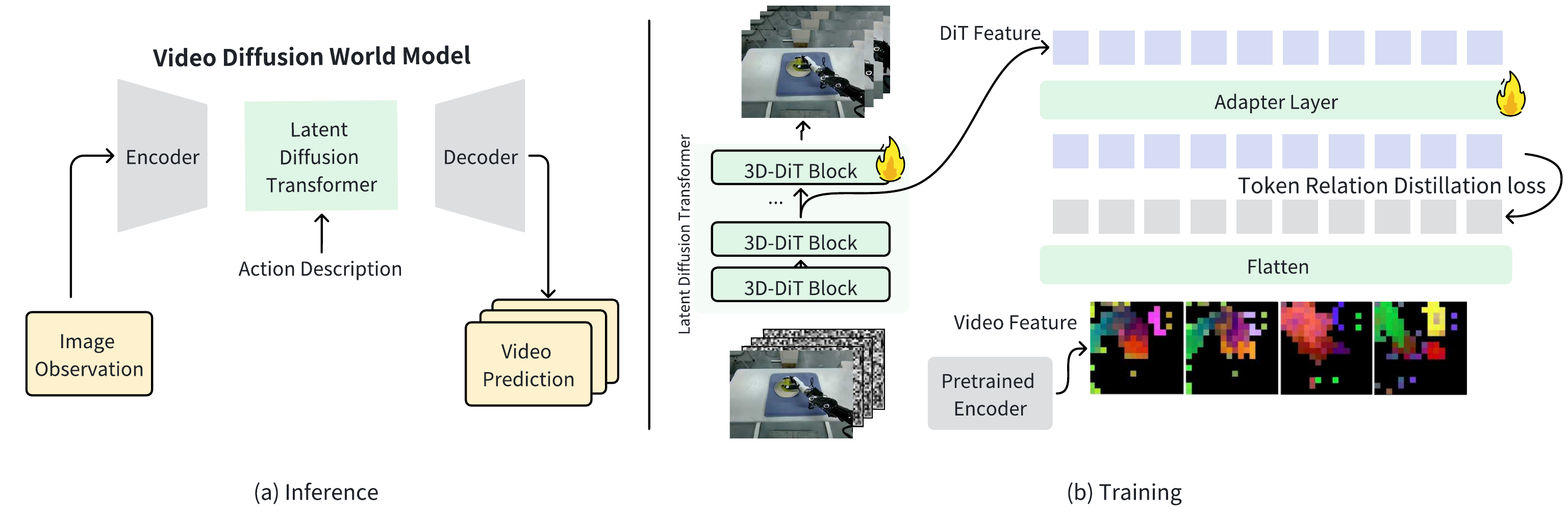}
  \caption{    \textbf{Overview of the Video Diffusion World Model.}
  (a) Inference: a latent diffusion transformer predicts future frames from image observations and text-based action descriptions. 
  (b) Training: DINO features supervise intermediate DiT representations via a token relation distillation loss to improve spatial-temporal modeling.
}
  % \caption{Overall architecture of the proposed world model. Textual instructions are expanded into detailed narratives and encoded via T5. Video inputs are compressed by a spatio-temporal autoencoder with 3D Haar wavelet decomposition. A Diffusion Transformer integrates (i) the text embeddings, (ii) the latent video tokens, and (iii) auxiliary features from a self-supervised (SSL) encoder (e.g., DINOv2) that processes the current frame/clip; these SSL features are fused into intermediate DiT blocks to guide denoising and strengthen pixel-level perception. The decoder reconstructs future frames via spatial upsampling and inverse wavelet transforms.}
  \label{fig:WoW-model}
\end{figure*}

\subsubsection{Pretrain Data Preparation}

We construct our training dataset through a multi-stage pipeline designed to ensure both quality and diversity. The process consists of four sequential stages: \textbf{Collection}, \textbf{Filtering}, \textbf{Refinement}, and \textbf{Rebalancing}. Unlike simply enlarging the dataset with indiscriminate samples, our approach emphasizes that \textit{data quality plays a decisive role in model performance}, and carefully curated data prove more effective than raw scale~\citep{brohan2023rtx,khazatsky2024droid,radford2021clip,northcutt2021confident, zhang2017rethinking, luo2024llm, an2024mc}.  

\paragraph{Collection.} 
We collect thousands of hours of videos from multiple robotic platforms, including Agibot~\citep{bu2025agibot}, Droid~\citep{khazatsky2024droid}, Robomind~\citep{wu2025robomind}, and a large amount of in-house data. These sources cover a variety of embodiments and task scenarios, providing broad coverage across environments and robot types. This diversity serves as the foundation for building generalizable robot learning datasets.  

\paragraph{Filtering.} 
% filter: RGB only, make full use of multi-view tag, embodied metadata, teperal length...
The collected data are processed through a series of filtering rules. Only RGB videos are retained, with BGR channels semi-automatically converted into RGB format for consistency. Static or non-informative sequences are removed, and a minimum length of 90 frames is enforced to ensure sufficient temporal context. In addition, we restrict the dataset to specific viewpoints such as head, wrist, and third-person perspectives, which best capture robot actions and task dynamics.  

\paragraph{Caption Refinement.} 
To further enhance the training signal, sparse textual annotations are expanded into dense descriptions using a pretrained VLM. Both uniform and sequential frame sampling are applied, ensuring coverage of both global context and local temporal transitions. Sparse and dense text annotations are combined with an approximate ratio of 1:4, and robot model identifiers are manually added into the text metadata. This step improves both the richness of supervision and the alignment between visual and textual modalities~\citep{radford2021clip, lin2024draw}.

\paragraph{Rebalancing.} 
Finally, we address imbalance across tasks by increasing the sampling probability of underrepresented tasks. This ensures that rare but important skills are not neglected during training, and improves robustness across diverse robotic behaviors~\citep{northcutt2021confident}.

Through this pipeline, we construct a dataset that is large in scale, carefully curated, temporally consistent, and densely annotated with semantic and physical labels—providing a robust foundation for training advanced robot learning models.

\subsubsection{Diffusion-Based Video Generation}

We adopt a video generation paradigm for the world model, treating the visual domain as the primary output modality due to its high information density. Our framework maps an initial state $s_t$ to its future state $\hat{s}_{t+1}$, where the hat indicates that it is predicted:

\begin{equation}
    o_t: \{o_t, p_t, [a_t, C_{\text{pose}}, \dots]\} 
\xrightarrow{\text{World Model}}
\hat{s}_{t+1}: o_{t+1}
\end{equation}

where $o_t$ is the current visual observation, $p_t$ is a high-level textual instruction, and optional inputs such as $a_t$ (low-level action) or $C_{\text{pose}}$ (camera pose) provide finer control. The full realization of this world model relies on specialized processing of each input modality. The following describes how textual, visual, and auxiliary signals are encoded and integrated to enable high-fidelity video prediction. 

\noindent\textbf{Textual Conditioning.}  
%Unlike most VLA models that rely on short prompts, 
We employ InternVL3-78B~\citep{zhu2025internvl3exploringadvancedtraining} to output language instructions into descriptive narratives of the environment, camera pose, robot embodiment, and intended action. 
These narratives are embedded by a pre-trained T5~\citep{raffel2020t5} encoder and injected as conditioning signals into the DiT, ensuring stronger alignment between textual context and generated video.

\noindent\textbf{Visual Encoding.}  
Raw video inputs are compressed into compact latent representations via a spatio-temporal autoencoder. To enhance modeling of physical interactions, we apply a 3D Haar wavelet transform that decomposes each video cube into low-frequency components—capturing scene structure—and high-frequency sub-bands—preserving fine motion details such as object collisions and deformations. This spectral separation allows the model to allocate capacity more effectively toward dynamic events. Spatial and temporal downsampling further reduces dimensionality for efficient processing without sacrificing critical physical cues.
%A spatio-temporal autoencoder is employed to compress raw video sequences into compact latent representations.
%To better capture physical dynamics, we adopt a 3D Haar wavelet transform, decomposing video cubes into low-frequency components (scene layout) and high-frequency sub-bands (motion details). This separation directs model capacity toward dynamic events like collisions and deformations. Downsampling further reduces spatial and temporal resolution for efficient processing.  

\noindent\textbf{Diffusion Transformer.}  
The denoising backbone is DiT~\citep{dit} composed of multi-head self-attention and feed-forward layers, enhanced with adaptive LayerNorm (adaLN) for timestep conditioning. Both absolute 3D positional embeddings and relative 3D RoPE are employed: the former preserves global coherence (e.g., trajectories), while the latter enforces local pixel-level causality (e.g., contact and continuity).  

\noindent\textbf{Auxiliary Perception.}  
To strengthen initial state understanding\cite{yu2025repa}, we inject features from a self-supervised visual representation model (DINOv2~\citep{Oquab2023DINOv2}) into intermediate layers of the DiT. These semantically grounded features improve pixel-level reasoning about object boundaries and spatial relationships, compensating for potential weaknesses in latent representations learned solely through noisy reconstruction objectives.
%To improve initial state understanding, features from a self-supervised visual model (DINOv2~\citep{Oquab2023DINOv2}) are injected into DiT intermediate layers. This strengthens pixel-level causal reasoning, compensating for weaknesses in features learned from noisy diffusion objectives.  

\noindent\textbf{Frame Decoding.}  
The decoder mirrors the encoder’s hierarchical structure, progressively reconstructing high-resolution frames through spatial upsampling, inverse wavelet transforms, and self-attention refinement. This multi-stage decoding process ensures both long-horizon temporal coherence and physically plausible fine details—such as texture preservation during deformation or accurate collision recovery.

\subsection{Solver-Critic Video Generation Agents}
\label{subsec:solver_critic_agent}

\begin{figure*}[t]
  \centering
  \includegraphics[width=1.0\linewidth]{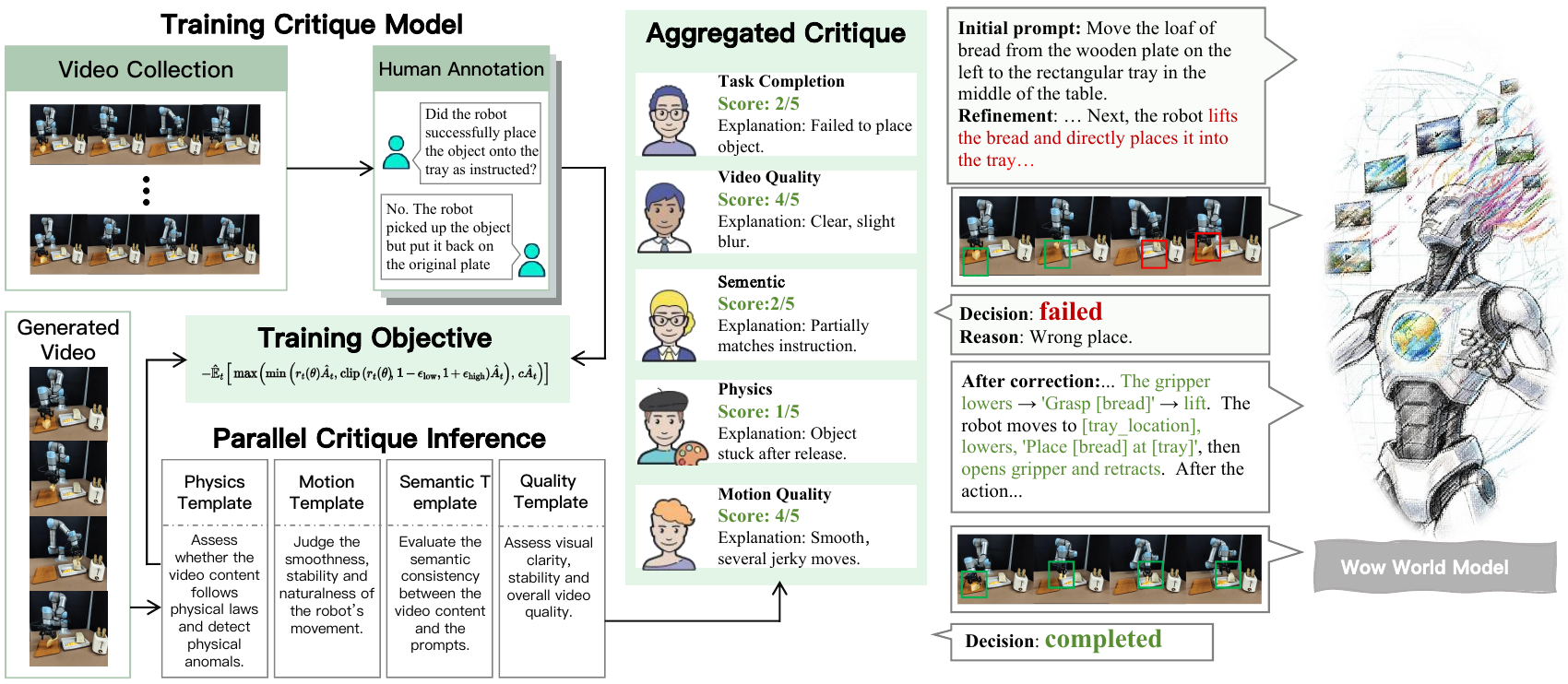}
  \caption{\textbf{Overview of Solver-Critic Video Generation Agents.} The left panel illustrates the Dynamic Critic Model Team, trained on annotated real and synthetic videos to evaluate physical plausibility. 
The right panel depicts the Refiner Agent, which iteratively rewrites prompts based on critic feedback and regenerates videos, forming a closed-loop optimization process.}
\vspace{-15pt}
  \label{fig:solver_critic_agent}
\end{figure*}

Building upon the conceptual framework of WoW, this section details the Experience Reflection (Perception) stage—a key component of our closed-loop mechanism. We introduce a solver-critic paradigm that enhances the physical consistency and realism of generated outputs. Unlike Simple ability of previous work~\citep{soni2024videoagent,chi2024eva}, WoW has a comprehensive agent system, driven by a Refiner  (Section~\ref{subsubsec:refiner_agent}), which, guided by a Dynamic Critic Model Team (Section~\ref{subsubsec:dynamic_critic_model_team}), to iteratively improve the generated content. This dynamic workflow forms a Closed-Loop Generative Workflow (Section~\ref{subsubsec:closed_loop}), ensuring that the model's output is continuously refined and verified.

\subsubsection{Framework Overview}
\label{subsubsec:framework_overview}

Achieving physically plausible video generation for complex, long-horizon robotic tasks requires moving beyond unidirectional models to a closed-loop, agentic system capable of self-perception and optimization. We frame this generative process as a deliberative act, analogous to the interplay between the intuitive "System 1" and the analytical "System 2" cognitive modes~\citep{weston2023system}. In our framework, an initial video serves as a "proposal" (System 1), which is then subjected to a rigorous critique and refinement loop that embodies the structured reasoning of System 2.

This architecture transforms the model from a passive generator into an active problem-solver. Our solver-critic framework is built upon three core components: the Refiner Agent, which optimizes the input and generates video output; the Dynamic Critic Model, which evaluates the generated output; and the integrated closed-loop Workflow. Furthermore, we discuss how this architecture aligns with the Prover-Verifier paradigm~\citep{kirchner2024prover}, showcasing its potential to endow the generative process with a new level of cognitive depth.

% \subsubsection{Orchestra Agent}
\subsubsection{Refiner Agent in World Model}
\label{subsubsec:refiner_agent}

The quality of a generative model's output is highly dependent on its input prompt. For video generation in specialized fields like robotics, prompts must capture subtle physical details to produce plausible outcomes. However, manually crafting such high-quality prompts is a time-consuming and arduous process of trial and error. While the emerging field of Automatic Prompt Engineering offers systematic optimization methods for language and generation tasks~\citep{khan2025test, agrawal2025gepa, an2025unictokens}, these general approaches are not directly tailored to the unique demands of physically-grounded video synthesis.

To address this challenge, we introduce the Refiner Agent, an autonomous system designed for test-time prompt optimization that does not require retraining the underlying video generation model. The agent takes a high-level user instruction and initiates an iterative refinement loop. In each iteration, a dedicated prompt rewriting module enhances the prompt's specificity and physical consistency. This rewriting process is explicitly guided by structured feedback from our Critic Model Team (Section ~\ref{subsubsec:dynamic_critic_model_team}), which identifies errors or missing details, such as adding constraints to prevent objects from passing through solid surfaces. Conceptually, this iterative process performs a guided search over the discrete prompt space, where the critic feedback functions as a “textual gradient”~\citep{pryzant2023automatic, yuksekgonul2024textgrad}. Our approach thereby transforms prompt engineering from a manual, trial-and-error task into a systematic, data-driven, closed-loop optimization process.

% Love u chimin
\subsubsection{Dynamic Critic Model Team}
\label{subsubsec:dynamic_critic_model_team}

Functioning as the 'verifier' in our iterative refinement loop, the Dynamic Critic Model Team is the second core component of our system. The need for this specialized critic arises because traditional metrics such as Fréchet Video Distance (FVD), Peak Signal-to-Noise Ratio (PSNR), and Structural Similarity Index Measure (SSIM)~\citep{unterthiner2018towards, huynh2012accuracy, hore2010image}, while capable of assessing visual fidelity, are inadequate for evaluating physical realism—a critical bottleneck in the development of robust world models. To address this gap, we align with the emerging consensus that VLMs are the cornerstone of next-generation video assessment~\citep{chen2024mllm}. However, general-purpose VLMs lack the domain-specific precision required for tasks like robot manipulation. We therefore construct our specialized critic by fine-tuning a VLM on a curated Question-Answering (QA) dataset containing both real and model-generated videos of robotic operations. This dataset is structured to probe the model's understanding across five key dimensions: task completion, action success, physical plausibility of interactions (e.g., stability, deformation), kinematic smoothness, and overall quality. This targeted fine-tuning transforms the generalist VLM into a reliable expert verifier, instilling it with the specialized knowledge required to accurately assess the physical dynamics of robot interaction.

\subsubsection{Closed-Loop Generative Workflow}
\label{subsubsec:closed_loop}

As illustrated in Figure~\ref{fig:solver_critic_agent}, our system integrates the Refiner Agent and Dynamic Critic Model into a closed-loop workflow that transforms video generation from a single-pass operation into an iterative refinement process. The loop initiates with a high-level user task, which the Refiner Agent expands into a detailed, physically-constrained prompt for our generation model (WoW). The resulting candidate video is then evaluated by the Dynamic Critic Model for physical plausibility and semantic coherence. If the video is judged 'incomplete' or 'failed', the critic provides structured feedback that the Refiner Agent incorporates to revise the prompt for the subsequent generation cycle. This iterative process of generation, critique, and refinement reframes video synthesis as an adaptive reasoning task. By endowing the generative pipeline with this self-corrective capability, our workflow enables the system to progressively converge on outputs that constitute a robust, physically grounded world model.

\subsubsection{Discussion: The Prover-Verifier Paradigm for Generative World Models}

To understand our architecture on a deeper level, this section explicitly connects it to established theoretical frameworks in the field of artificial intelligence—namely, the Solver-Critic~\citep{wang2025improving, mcaleese2024llm, gou2023critic} and Prover-Verifier~\citep{kirchner2024prover} paradigms. In these paradigms, one agent (the Prover/Solver) is responsible for generating a candidate solution, while another, often simpler or more specialized agent (the Verifier/Critic), is responsible for evaluating its correctness.

Our architecture provides a concrete implementation of the established Prover-Verifier and Solver-Critic paradigms~\citep{wang2025improving, mcaleese2024llm, gou2023critic, kirchner2024prover}. Within this framework, the Refiner Agent function as the Prover/Solver, responsible for proposing and iteratively refining candidate videos. The Dynamic Critic Model Team acts as the Verifier/Critic, tasked with evaluating the physical plausibility of these proposals. A key contribution of our work is being the first to successfully apply this paradigm-traditionally used for discrete, logical tasks such as mathematical proofs~\citep{lin2025goedel} and code generation~\citep{wang2025co}-to the high-dimensional, continuous, and stochastic domain of video generation.

The primary advantage of this approach is its ability to optimize for complex, non-differentiable objectives like "physical realism" without requiring an explicit, differentiable loss function. The Prover learns to generate outputs that are accepted by the Verifier, providing a powerful mechanism for instilling abstract values like physical common sense into generative models. To summarize, this framework paves the way for building physically and causally consistent world models suitable for robotics planning.

\subsection{Flow-Mask Inverse Dynamics Model}
\label{sec:model_idm}

\begin{figure*}[t]
  \centering
  \includegraphics[width=0.95\linewidth]{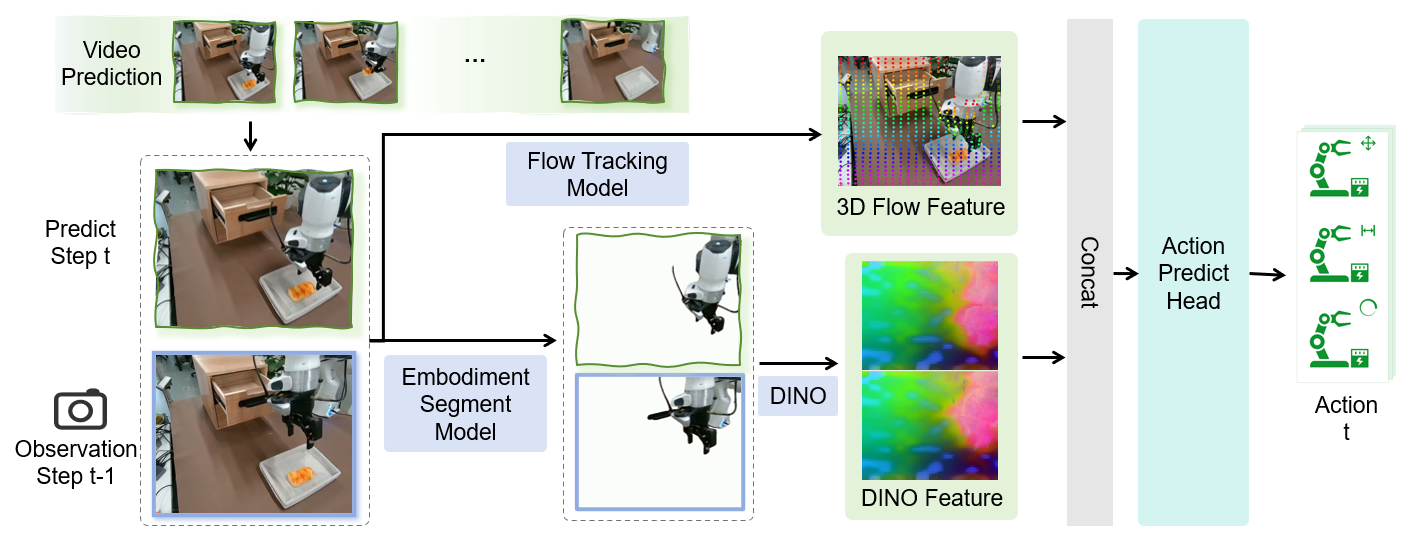}
  \caption{Work flow of inverse dynamics model. Giving two frame predictions, our FM-IDM can estimate the delta End-Effect action of the robot. }
  \label{fig:idm}
\end{figure*}

The proposed Flow-Mask Inverse Dynamics Model (FM-IDM) is a video-to-action model that maps predicted video frames to real-world robot execution transitions. Instead of relying on model-specific features~\citep{liao2025genieenvisionerunifiedworld,chi2025mindlearningdualsystemworld,hu2025videopredictionpolicygeneralist}, we adopt a pixel-level decoding approach, trading real-time performance for greater generality and accuracy~\citep{ko2023avdc,tan2025anyposautomatedtaskagnosticactions}. Designed as a plug-and-play module, our model is compatible with any visual generative world model, enabling system-level evaluation and facilitating reward extraction via embodied interaction.

\paragraph{Task Formulation}
Given two consecutive visual observations $(o_t, o_{t+1})$ from the predicted video — each corresponding to the underlying robot states $(s_t, s_{t+1})$ — the goal is to infer the end-effector action $a_t$ that transitions the robot from $s_t$ to $s_{t+1}$.
The inverse dynamics model $F_\delta$ takes the current frame $o_t$ and the corresponding flow $\mathcal{F}_{t \rightarrow t+1}$ as input, and outputs a predicted delta action $\hat{a}_t$:
\begin{equation}
    \hat{a}_t = F_{\delta}(o_t, \mathcal{F}_{t \rightarrow t+1})
\end{equation}

\noindent\textbf{FM-IDM}
To achieve this, we first estimate a motion field $\mathcal{F}_{t \rightarrow t+1}$ capturing the geometric transformation between frames. The estimated flow encodes both translational and rotational motion of the manipulator. 
We implement $F_\delta$ as a two-branch encoder-decoder network. 
% TODO这里要加上用的什么模型，现在是提出mask+Flow+Depth estimate+DINO Feature?然后加上in~\ref{fig:idm}. 
We first fine-tuned a SAM~\citep{sam} that process the masked current frame $o_t$ to extract scene and embodiment context; 
the other processes the optimal flow by CoTracker3 model~\citep{karaev2024cotracker3simplerbetterpoint} $\mathcal{F}_{t \rightarrow t+1}$ to capture fine-grained temporal dynamics, as described in Figure~\ref{fig:idm}. In conjunction with the with the DINO~\citep{Oquab2023DINOv2} feature, we further use Multi-Layer Perceptron (MLP) as action head to learn the 7-DoF action feature.
The training objection is as follows:
\begin{equation}
    \min_{\delta} \; \mathbb{E}_{(o_t, o_{t+1}, a_t)} \; d \big(a_t, F_\delta(o_t, \mathcal{F}_{t \rightarrow t+1}) \big)
\end{equation}
where $d(\cdot, \cdot)$ denotes a weighted smooth L1 loss in the end-effector action space.

By explicitly modeling spatio-temporal correspondences, the model generalizes better across diverse tasks, background variations, and occlusions, and is robust to noise in video-based prediction.

\noindent\textbf{Embodiment-Centric End-Effector Action Dataset}
To facilitate the learning of end-effector actions directly from visual input, we curate a dataset of 646k image–action pairs across 219 tasks, covering a broad range of manipulation scenarios. The dataset is carefully constructed to span a diverse action space and densely cover the reachable workspace of the robot, ensuring that the model learns from a comprehensive set of physically plausible end-effector configurations. More details of the implementation are included in Section ~\ref{sec:exp}.

\noindent\textbf{Real-World Feedback through IDM}
% IDM-level reward: 
At the action execution stage, rewards are grounded in physical feasibility and obtained through direct interaction with the environment. They may be defined in multiple ways: binary success/failure of task completion, distance-based metrics between predicted and actual end-effector positions, force/torque stability measures during contact, or energy-efficient motion profiles. Failures serve not merely as penalties but as corrective feedback that guides continual adaptation. Importantly, this reward can be further fed back to the world model, adjusting the model through Group Relative Policy Optimization (GRPO) for evolutionary visual generation~\citep{xue2025dancegrpounleashinggrpovisual}.

% \section{WoWBench: World Model Benchmark}
\section{WoWBench: A Multi-faceted Benchmark for Embodied World Models}
\begin{figure*}[t]
  \centering
  \includegraphics[width=\textwidth]{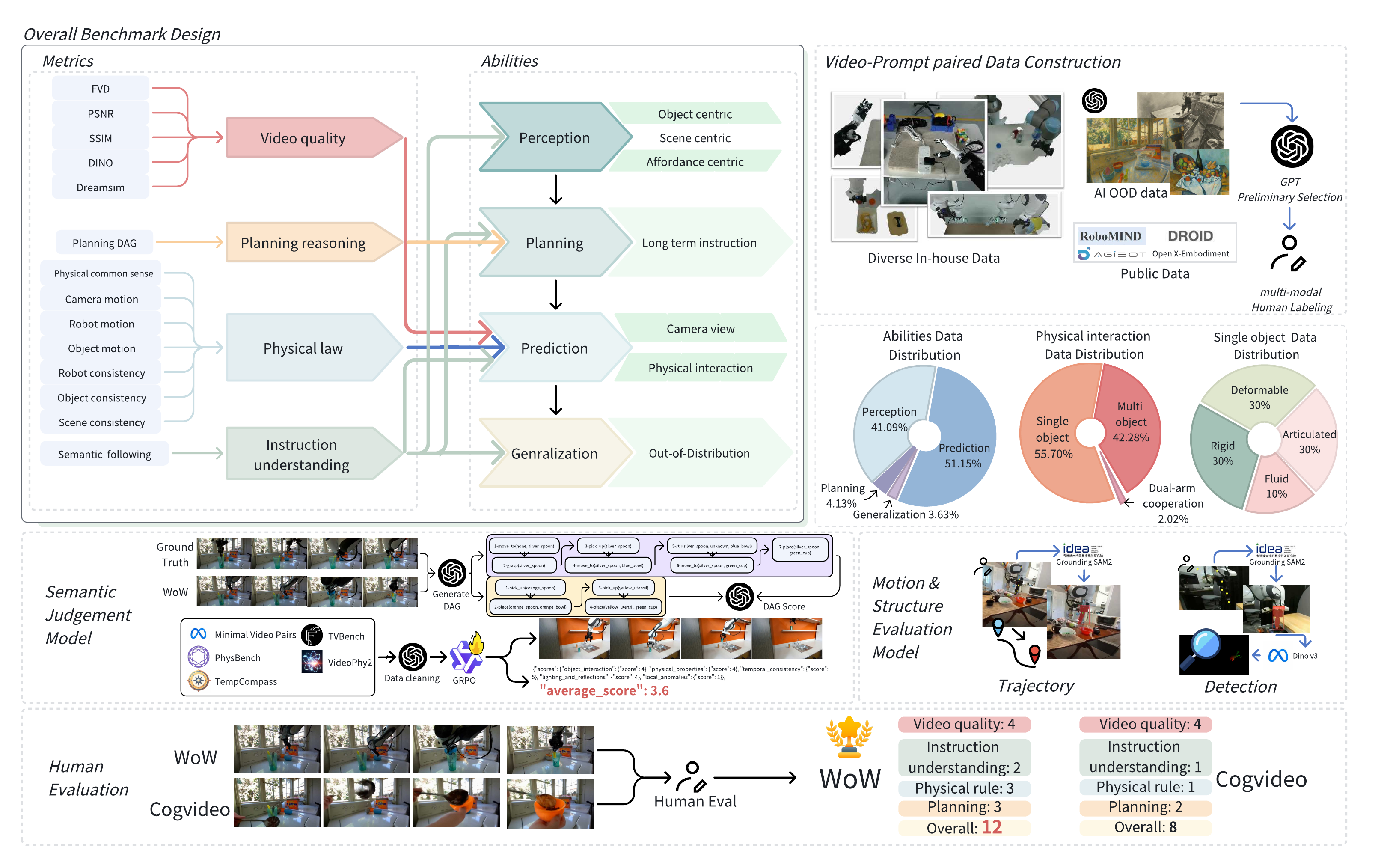} 
  \caption{\textbf{The Overall Design of WoWBench.} Our benchmark is structured around five core components. \textbf{(Top-left)} A multi-faceted \textit{Metrics} suite evaluates generated videos from four key perspectives: video quality, planning reasoning, physical rules, and instruction understanding. \textbf{(Top-center)} These metrics are designed to assess four fundamental \textit{Abilities} of an embodied world model: Perception, Planning, Prediction, and Generalization. \textbf{(Top-right)} The benchmark is built upon a large-scale \textit{Data Construction} pipeline that leverages diverse data sources (in-house, public, and AI-generated) and a human-in-the-loop annotation process (GPT-based selection followed by human labeling) to create video-prompt pairs. Three pie charts demonstrate the statistics of data distribution in different dimensions. \textbf{(Center)} The benchmark uses two different evaluation methods. Strong expert models for motion and consistency, and GPT or Fine-tuned VLM for instruction understanding and planning. \textbf{(Bottom)}We also asked 12 domain experts to perform a human evaluation on the generated videos.}
  \label{fig:WoWbench_overall_design}
\end{figure*}

In this section, we introduce WoWBench, a novel benchmark designed to rigorously evaluate embodied world models. We formulate the core evaluation task as conditional \textbf{video generation from an initial image and a text instruction (Image+Text-to-Video)}, a setting that directly probes a model's ability to understand a given state and execute a specified action. Moving beyond metrics of visual appeal, WoWBench is specifically engineered to assess a model's instruction understanding, planning and observation perception abilities and understanding of physically grounded dynamics in embodied settings. The overall design of our benchmark, which systematically connects core abilities, a principled data construction pipeline, and a multi-faceted evaluation protocol, is illustrated in Figure~\ref{fig:WoWbench_overall_design}. We structure our presentation by first outlining the four core capabilities that a competent embodied world model must possess (Section~\ref{ssec:core_abilities}). We then detail our task design and data curation process (Section~\ref{ssec:data_curation}), followed by our comprehensive evaluation metrics (Section~\ref{ssec:metrics}).

\subsection{Core Evaluation Dimensions}
\label{ssec:core_abilities}
We posit that a truly effective embodied world model must demonstrate mastery across four fundamental and orthogonal dimensions, as outlined in our benchmark design (Figure~\ref{fig:WoWbench_overall_design}, center).

\paragraph{Perception Understanding}
A world model must first accurately perceive and represent the environment in order to enable more reliable subsequent prediction and planning. We assess this through tasks requiring fine-grained \textbf{object recognition}~\citep{cheng2024egothinkevaluatingfirstpersonperspective, cheng2024videgothinkassessingegocentricvideo, 10654928, chow2025physbench}(attributes like color, shape, number, and size), \textbf{spatial understanding}~\citep{yang2025thinkingspacemultimodallarge, cheng2025embodiedevalevaluatemultimodalllms, song2025robospatialteachingspatialunderstanding, du2024embspatialbenchbenchmarkingspatialunderstanding} (relative positions and arrangements), and \textbf{affordance recognition}~\citep{gibson2014theory, nasiriany2025rt, cheng2025embodiedevalevaluatemultimodalllms} (identifying interactive parts of objects). Each dimension contains a different amount of data. In the object sub-dimension, we included approximately 143 samples, with around 20 samples allocated to each attribute category (color, number, shape, size, type) and 50 samples to the function category. In the spatial sub-dimension, we assigned 46 samples, while the affordance sub-dimension was filled with 60 samples.

\paragraph{Predictive Reasoning}
This dimension evaluates the model's internal physics engine. Given an initial state and an action, the model must generate a future that respects core physical principles such as \textbf{object permanence}~\citep{bansal2024videophyevaluatingphysicalcommonsense, chow2025physbench}, \textbf{collision dynamics}~\citep{meng2024worldsimulatorcraftingphysical, chow2025physbench, bansal2024videophyevaluatingphysicalcommonsense, motamed2025generative}, and \textbf{trajectory plausibility}~\citep{qin2024worldsimbenchvideogenerationmodels, li2025worldmodelbenchjudgingvideogeneration, duan2025worldscoreunifiedevaluationbenchmark, zheng2025vbench20advancingvideogeneration, yue2025ewmbenchevaluatingscenemotion}. This directly probes the model's capacity to function as a world simulator~\citep{sora2024}. Therefore, we design several sub-dimensions that focus on these principles, as illustrated by the pie chart in the center-right of Figure~\ref{fig:WoWbench_overall_design}. We select both objects with no occlusion, 107 samples, and objects with semi-occlusion, 54 samples, as our different camera views for evaluating varying levels of object visibility. Also, for collision dynamics, we further subdivide the dimension into single-object operation, multi-object interaction, and dual-arm cooperation. The single-object operation data has 83 samples, which are distributed in the ratio of 30, 30, 30, 10 for rigid, deformable, articulated, and fluid. The multi-object interaction part has 63 samples, covering the interaction between rigid body-rigid body, rigid body-deformable object, and rigid body-fluid. For dual-arm cooperation, now we have 3 samples, but we will continue to collect more data.

\paragraph{Decision-making and Planning}
Embodied agents must execute long-horizon tasks~\citep{chen2024egoplanbenchbenchmarkingmultimodallarge, sermanet2023robovqamultimodallonghorizonreasoning, li2024mmromultimodalllmseligible, cheng2024videgothinkassessingegocentricvideo, yang2025embodiedbenchcomprehensivebenchmarkingmultimodal}. Therefore, we assess a model's planning ability by challenging it to generate coherent video sequences for complex instructions. This requires implicitly understanding \textbf{task decomposition}~\citep{chi2024eva, tian2025seea} into key sub-goals and respecting their \textbf{causal dependencies}. Hence, we collect 25 samples to fill in this long-term planning task, and transform the text instruction into a suitable description for the world model to plan. In order to evaluate the planning ability in the world model, we refer to the metric from RoboBench. As illustrated in the center-left of Figure~\ref{fig:WoWbench_overall_design}, first, we extract the key steps from the predicted video using Gemini-2.5-flash~\citep{comanici2025gemini}, for instance, 1-grasp(green block), 2-pick up(green block), 3-place(green block, yellow block), and so on, then transform them into a Directed Acyclic Graph(DAG) for further key action extraction and comparison with groundtruth DAG. The detailed metric will be extended in Section~\ref{ssec:dag}. 

\paragraph{Generalized Execution}
A universal world model should not only perform well on the In-Distribution data, but it should also generalize beyond the data it has seen before to demonstrate its generalization ability. For this reason, we test generalization on the in-house robot data that we collected, by using GPT-5~\citep{openai2024gpt4ocard} to perform style transfer or image editing on it, and generating images that the world model had never seen before. We also collected some world-famous masterpiece paintings, such as \textit{"Girl with a Pearl Earring"}, and asked the world model to execute the task instructions human created. These two types of images constitute our \textbf{Out-of-Distribution(OOD)} dimension. Here, in total, we provide 20 data samples, and demonstrate some cases in Figure~\ref{fig:other_generalize}.

\subsection{Task Design and Data Curation}
\label{ssec:data_curation}
To rigorously evaluate these capabilities, we developed the principled, semi-automated data curation pipeline shown in Figure~\ref{fig:WoWbench_overall_design} (Top-right). Our dataset is built from a mixture of open-source robotics data (e.g., RoboMIND~\citep{wu2025robomind}, DROID~\citep{khazatsky2024droid}), in-house collected trajectories, and AI-generated OOD data to ensure diversity.

Our process begins by leveraging GPT-4o~\citep{openai2024gpt4ocard} as an intelligent annotator for preliminary data filled in our dimensions. For each candidate video-instruction pair, GPT-4o scores its relevance against the definitions of our four core capability dimensions. This allows for efficient, large-scale sorting of data into targeted evaluation buckets. Following this automated stage, human experts perform a verification step to ensure gold-standard quality, resolving ambiguities and filtering out misclassifications. This human-in-the-loop approach ensures both scalability and accuracy.

Finally, expert annotators identify the optimal initial frame for each task and provide crucial annotations. The final benchmark consists of tuples, each containing: 1) a natural language \textbf{instruction}, 2) an initial \textbf{image}, 3) the ground-truth \textbf{video}, and 4) annotated \textbf{keypoints} for tracking.

\subsection{Multi-faceted Evaluation Metrics}
\label{ssec:metrics}
Our evaluation protocol is a suite of metrics designed to be as comprehensive as the capabilities we measure (Figure~\ref{fig:WoWbench_overall_design}, left). We introduce several novel metrics alongside standard ones, grouped by the property they assess.

\paragraph{Visual Fidelity and Temporal Consistency.}
While we report standard video quality metrics (FVD\citep{unterthiner2018towards}, SSIM\citep{1284395}, PSNR\citep{fardo2016formalevaluationpsnrquality}, DINO~\citep{Oquab2023DINOv2}, Dreamsim~\citep{fu2023dreamsim}) in the appendix, our primary contribution is a novel metric for temporal consistency with high diagnostic power. While pixel evaluation can give a overall visual quality comparison.
% Prior work, such as EWMBench~\citep{ewmbench}, uses global features (e.g., DINOv2\citep{Oquab2023DINOv2}), which cannot disentangle inconsistencies in the background, robot arm, or manipulated object. 

\paragraph{Mask-guided Regional Consistency.}
To better disentangle inconsistencies in the background, robot arm, or manipulated object, we propose \textbf{Mask-guided Regional Consistency} as similar to EWMBench~\citep{yue2025ewmbenchevaluatingscenemotion}. As illustrated in Figure~\ref{fig:WoWbench_overall_design}(Center-right), we first use the GroundedSAM2~\citep{ren2024grounded} with human annotation to obtain masks for the robot arm, the manipulated object(s), and the background in each frame. We then compute region-specific embeddings using a vision foundation model (e.g., DINOv3~\citep{siméoni2025dinov3}) and measure cosine similarity across time for each region separately. This allows us to pinpoint the source of temporal flaws—for instance, identifying a "jittery" robot arm even when the object and background are stable.

% \begin{figure}[h]
%   \centering
%   \includegraphics[width=0.9\linewidth]{placeholder.png} % 替换为你的示意图文件
%   \caption{Illustration of our Mask-guided Regional Consistency metric. We segment the frame into three regions (robot, object, background) and compute consistency scores for each, providing superior diagnostic capability over global metrics.}
%   \label{fig:mask_consistency}
% \end{figure}

\paragraph{Instruction Understanding and Semantic Correctness.}
% We employ GPT-4o~\citep{openai2024gpt4ocard} as a scalable evaluator to assess semantic alignment with the given instruction. We design two schemes:
% \begin{itemize}
%     \item \textbf{With Ground-Truth:} We prompt GPT-4o to extract structured descriptions (Initial, Processing, Final states) from both the generated and ground-truth videos. A Vision-Language Model then scores the consistency between these structured descriptions.
%     \item \textbf{Without Ground-Truth (OOD):} GPT-4o directly analyzes the generated video against the instruction, outputting a \textit{Sequence Match Score} (evaluating the order of actions) and an \textit{Execution Quality Score} on a 1-5 scale.
% \end{itemize}
We use GPT-4o~\citep{openai2024gpt4ocard} as a scalable evaluator to assess semantic alignment with the given instruction. Depending on whether ground-truth (GT) video is available, we adopt two protocols:

\begin{itemize}
    \item \textbf{With Ground-Truth:} We first prompt GPT-4o to extract structured descriptions (Initial, Processing, Final states) from both the generated and GT videos. A vision–language model then scores their \textit{Caption Score}. In addition, GPT-4o evaluates the generated video against the instruction to produce a \textit{Sequence Match Score} (order of actions) and an \textit{Execution Quality Score} (1–5 scale). \emph{We report all three metrics in this setting.}
    \item \textbf{Without Ground-Truth (OOD):} When GT is unavailable, we only assess instruction adherence: GPT-4o directly analyzes the generated video to output the \textit{Sequence Match Score} and the \textit{Execution Quality Score}. \emph{Only these two metrics are reported in this setting.}
\end{itemize}

\paragraph{Physical and Causal Reasoning.}
To quantify the physical plausibility of generated videos, we compute:
\begin{itemize}
    \item \textbf{Trajectory Consistency:} To compare the trajectory between generated videos and ground-truth counterparts, we track both the end-effector and object trajectories. In our WoWBench, we leverage SAM2\citep{ravi2024sam2segmentimages}, given a few representative points in the initial frame, to follow the motion of objects in both videos. 
    % To address disparities in spatial and temporal resolution, we normalize coordinate values and uniformly sample the ground-truth videos to match the length of the generated ones. 
    Trajectory similarity is then evaluated using a complementary set of metrics: Mean Euclidean Distance (MED)~\citep{Dokmanic_2015} to capture average deviation, Dynamic Time Warping (DTW)~\citep{Müller2007} to assess temporal alignment, and Fréchet Distance~\citep{Eiter1994ComputingDF} to measure worst-case path similarity.

    \item \textbf{Physical common sense:} Physical common sense covers dimensions ranging from object interaction and properties to temporal consistency, lighting, fluid dynamics, and local anomalies. To automatically score these six distinct dimensions in generated videos, we collected several datasets~\citep{krojer2025shortcutawarevideoqabenchmarkphysical, chow2025physbench, bansal2025videophy2challengingactioncentricphysical, cores2025losttimenewtemporal, liu2024tempcompassvideollmsreally, zhang2024unveiling} and converted them into instruction-tuning datasets to fine-tune Qwen-2.5-VL~\citep{bai2025qwen25vltechnicalreport} to enhance the understanding and consistency of physical law and employed a 1-to-5 scoring scale across categories.
    % We conducted a targeted fine-tuning initiative to develop a model capable of automating the assessment of physical realism in videos. Utilizing the GRPO (Group Relative Policy Optimization) algorithm on a dataset of 50,000 samples, the fine-tuned VLM achieved a significantly enhanced understanding of video temporality, causality, and physical principles. This model is now deployed within a rigorous, multi-dimensional framework that strictly judges physical realism, employing a 1-to-5 scoring scale across six distinct categories. These categories comprehensively cover key physical dimensions ranging from object interaction and core physical properties to temporal consistency, lighting, fluid dynamics, and local anomalies.
\end{itemize}

\paragraph{Planning and Task Decomposition.}
\label{ssec:dag}
To evaluate long-horizon planning, we refer to the metric of RoboBench based on Directed Acyclic Graphs (DAGs). We first parse the natural language instruction and ground-truth video into a ground-truth plan DAG, where nodes are atomic actions and edges represent dependencies. This representation flexibly handles non-unique but valid action orderings. We then compare the model-generated plan (which also uses the same approach to infer from the video) to the ground-truth DAG using three scores:
\begin{enumerate}
    \item \textbf{Key-step Recall (\(R_{k}\)):} The fraction of essential ground-truth steps the model executes.
    \item \textbf{Sequential Consistency (\(R_{s}\)):} The normalized length of the longest correctly ordered sequence of key steps.
    \item \textbf{Key-step Precision (\(P_{k}\)):} The fraction of predicted key steps that are correct and non-superfluous.
\end{enumerate}
The final planning score \(S_{\text{plan}}\) integrates these aspects to reward both completeness and correctness:
\begin{equation}
S_{\text{plan}} = \left(0.5 \times R_{k} + 0.5 \times R_{s}\right) \times P_{k}
\label{eq:planning_score}
\end{equation}

\subsection{Overall Benchmark Score}

\paragraph{Setup.}
For each model $i$ and metric $m$, we map the raw measurement $x_{i,m}$ to a common desirability score
$s_{i,m}\in(0,100)$ via a monotone parametric mapping applied after an absolute pre-scaling to $[0,1]$.
We then aggregate desirability scores by weighted arithmetic means at both metric-group and overall levels.

\paragraph{Pre-scale to $[0,1]$ with absolute anchors.}
Let $L_m<U_m$ be fixed anchors for metric $m$ (documented per-metric).
Define the clipping operator $\mathrm{clip}(u;a,b)=\min\{\max\{u,a\},\,b\}$.
We first clamp raw values to $[L_m,U_m]$, then linearly map to $[0,1]$; for “higher-is-better” (HIB) metrics:
\[
\hat x_{i,m}^{\mathrm{HIB}}
=\frac{\mathrm{clip}(x_{i,m};L_m,U_m)-L_m}{U_m-L_m}\in[0,1],
\]
and for “lower-is-better” (LIB) metrics:
\[
\hat x_{i,m}^{\mathrm{LIB}}
=1-\frac{\mathrm{clip}(x_{i,m};L_m,U_m)-L_m}{U_m-L_m}\in[0,1].
\]
We use absolute anchors for two common metrics:
\[
\textbf{PSNR (HIB):}\quad L_{\text{PSNR}}=0,\ U_{\text{PSNR}}=50\ \ (\text{truncate }x\le 50\ \text{before scaling}); 
\qquad
\]
\[
\textbf{FVD (LIB):}\quad L_{\text{FVD}}=0,\ U_{\text{FVD}}=2000\ \ (\text{truncate }x\le 2000).
\]
For other metrics, $(L_m,U_m)$ are fixed per protocol (e.g., theoretical bounds for bounded scales, task-specific absolute targets for unbounded ones).

\paragraph{Monotone parametric mappings.}
After pre-scaling, we apply a single-parameter monotone transform $f_m(\cdot;\theta_m)$ and then rescale to $(0,100)$:
\[
s_{i,m}=100\,f_m\!\big(\hat x_{i,m};\theta_m\big),\qquad s_{i,m}\in(0,100).
\]
We consider the following families (all are strictly increasing on $[0,1]$):
\[
\begin{aligned}
&\textbf{Power (Gamma):} && f_\gamma(x)=x^\gamma,\quad \gamma>0;\\
&\textbf{Logit temperature:} && f_T(x)=\sigma\!\big(\mathrm{logit}(x)/T\big),\quad T>0,\ \ \sigma(t)=\tfrac{1}{1+e^{-t}};\\
&\textbf{Tanh slope:} && f_\kappa(x)=\tfrac{1}{2}\big(\tanh(\kappa(2x-1))+1\big),\quad \kappa>0.
\end{aligned}
\]
In practice, $\gamma>1$ accentuates the high end, while $T<1$ or $\kappa>1$ expands the mid-range and compresses extremes.
For numerical stability with logit we use a small $\varepsilon$ (e.g., $10^{-6}$) and replace $x$ by $\mathrm{clip}(x;\varepsilon,1-\varepsilon)$ only inside $\mathrm{logit}(\cdot)$.

\paragraph{Parameter selection and freezing.}
For each metric $m$, $\theta_m\in\{\gamma,T,\kappa\}$ is selected on a fixed development set by maximizing
the Fisher-$z$ averaged Pearson correlation between $f_m(\hat x;\theta)$ and human ratings across $K$-fold CV;
Spearman correlation is used as a tie-breaker. The chosen $\theta_m$ is then \emph{frozen} and applied to all evaluations.

\paragraph{Intra-group averaging (uniform).}
Metrics are organized into groups $g$ (e.g., \emph{quality}, \emph{instruction}, \emph{physical}, \emph{planning}).
For model $i$, let $\mathcal{M}_g$ be the set of metrics in group $g$ that are available for $i$, and $N_{i,g}=|\mathcal{M}_g|>0$.
We compute the group score as the simple arithmetic mean:
\[
G_{i,g}=\frac{1}{N_{i,g}}\sum_{m\in\mathcal{M}_g} s_{i,m}.
\]

\paragraph{Aggregation (weighted arithmetic mean).}
Let nonnegative group weights $\{W_g\}$ sum to one over the groups available for model $i$.
The overall score is
\[
O_i=\sum_{g} \widetilde W_{i,g}\, G_{i,g},\qquad
\widetilde W_{i,g}=\frac{W_g}{\sum_{h\in\mathcal{G}_i} W_h},
\]
where $\mathcal{G}_i=\{g:\,N_{i,g}>0\}$.
For an unweighted overall mean across groups, set $W_g\equiv 1$.

\section{Experiment: Evaluating Generative World Models}
\label{sec:exp}

Evaluation of our world foundation model is structured into four main parts, covering model comparisons, scaling law analysis, generalization, and real-world robotics. Prior to detailing these experiments, we provide a comprehensive overview of our training data in Section~\ref{sec:trainingdata}. The comparison of different models in Section~\ref{sec:diffmodel} examines the impact of varying pre-training methods and model architectures on performance. The scaling law analysis in Section~\ref{subsec:scaling} measures performance across varying data volumes, trainable parameters and task difficulty levels. Generalization capabilities are tested against novel scenes, objects, and actions to ensure robustness in Section~\ref{subsec:generalization}. Finally, practical deployment is assessed by the model's ability to generate executable actions in Section~\ref{subsec:real}.

% 要不要列上想要讨论的问题

% We propose qualitative evaluations from five aspects, demonstrating the model's ability to understand light, weight, and other physical properties.
\subsection{Training Data}
\label{sec:trainingdata}

% Our training dataset is built from 2.03 million video clips, which total over 5,000 hours of interaction footage, or approximately 1 billion frames sampled at around 24 FPS. The data was collected from over 200 distinct simulated scenes and features 5 different types of robot agents. For pre-processing, all videos were captured at a native resolution of 640×480 and then upsample to 720×1024 pixels. During data cleaning, we discarded approximately 75\% of the raw data due to task failures or low quality.

Our training dataset was meticulously curated to provide a rich and diverse foundation for learning a generalizable world model. It comprises 2.03 million video clips, totaling over 7,300 hours of interaction footage, which corresponds to approximately 633 million frames sampled at a consistent 24 frames per second. To foster robust generalization, the data was collected from over 200 procedurally generated simulated scenes, spanning contexts from complex household environments (e.g., kitchens, living rooms with cluttered objects) to structured industrial settings (e.g., warehouses, assembly lines). Crucially, the dataset features a diverse collection of 12 distinct robot embodiments to ensure the model learns a wide range of physical dynamics and morphologies. The collection is dominated by industrial manipulators, with a significant emphasis on both single-arm and dual-arm configurations. The primary data sources include trajectories from the dual-arm Franka FR3, the single-arm UR5e, and the dual-arm UR5e, which together constitute a substantial portion of the dataset. To further broaden the diversity, we also incorporate data from various other platforms, including multiple Franka Emika Panda setups and several specialized configurations from the ARK, AgileX, and Tienkung series, ensuring a broad spectrum of kinematic properties and action spaces are represented. For pre-processing, all video sequences were captured at a native resolution of 640×480 and subsequently upsampled to 720×1024 pixels to align with our model's architectural input requirements. We applied a rigorous filtering pipeline to ensure data quality, which led to the exclusion of approximately 75\% of the initial raw data. This high discard rate was a deliberate choice to remove trajectories with simulation instabilities, severe collisions, task failures, and periods of static inactivity, thereby ensuring the final dataset consists of high-quality, meaningful interactions.

\subsection{Model Comparsion Experiment}
\label{sec:diffmodel}

\begin{figure*}[t]
    \centering
    \begin{minipage}[t]{0.98\linewidth}
    \centering
        \captionof{table}{
        \textbf{Comparative analysis of foundational video generation models.} We benchmark our \textbf{WoW-DiT} against SOTA models using direct text-to-video generation. All metrics: higher is better. Best results are \textbf{bold} with \cellcolor{mylightgreen} highlight.
        }
        \label{tab:foundation-model-comparison}
        \resizebox{\textwidth}{!}{%
        \begin{tabular}{lccccccccccc}
        \toprule
        \multirow{2}{*}{\textbf{Model}} & \multirow{2}{*}{\textbf{Base}} 
        & \multicolumn{5}{c}{\textbf{Human Evaluation}} 
        & \multicolumn{5}{c}{\textbf{Autonomous Evaluation}} \\
        \cmidrule(lr){3-7} 
        \cmidrule(lr){8-12}
        & & \textbf{VQ} & \textbf{IF} & \textbf{PL} & \textbf{Plan} & \textbf{Overall}
          & \textbf{VQ} & \textbf{IF} & \textbf{PL} & \textbf{Plan} & \textbf{Overall} \\
        \midrule
        Cogvideo & cogvideo & 3.29 & 1.52 & 1.73 & 1.30 & 7.84 & 38.52 & 54.09 & 63.30 & 2.32 & 39.56 \\
        Cosmos-Predict1 & cosmos1 & 2.84 & 2.60 & 2.41 & 2.49 & 10.34 & 39.06 & 61.46 & 59.05 &	\cellcolor{mylightgreen}\textbf{7.47} & 41.76 \\
        Wan2.1 & wan & 3.49 & 1.79 & 2.30 & 1.62 & 9.21 & 40.23 & 56.85 & 59.66 & 5.6 & 40.59 \\
        Cosmos-Predict2 & cosmos2 & 3.18 & 2.33 & 2.31 & 2.27 & 10.09 & 46.81 & 56.80 & 60.56 &	6.67 & 42.71 \\
        \midrule
        \rowcolor{lightgray}
        \multicolumn{12}{l}{\textit{Our Foundational Model}} \\
        \midrule
        \textbf{WoW-DiT} & cosmos1 & 3.12 & 2.86 & 2.78 & 2.84 & 11.60 & 49.35 & 69.68 & 62.28 & 2.89 &	46.05 \\
        \textbf{WoW-DiT} & wan & \cellcolor{mylightgreen}\textbf{4.09} & 2.60 & \cellcolor{mylightgreen}\textbf{3.16} & 2.52 & 12.37 & \cellcolor{mylightgreen}\textbf{55.38} & 62.16 &	63.75 &	4.74 & 46.51 \\
        \textbf{WoW-DiT} & cosmos2 & 3.76 & \cellcolor{mylightgreen}\textbf{3.19} & 3.03 & \cellcolor{mylightgreen}\textbf{3.36} & \cellcolor{mylightgreen}\textbf{13.34} & 54.12 & \cellcolor{mylightgreen}\textbf{70.36} & \cellcolor{mylightgreen}\textbf{66.18} & 6.88 & \cellcolor{mylightgreen}\textbf{49.39} \\
        \bottomrule
        \end{tabular}%
        }
    \end{minipage}
    
    \begin{minipage}[t]{0.64\linewidth}
    \vspace{1em}
    \centering
    \captionof{table}{Autonomous evaluation of models with a self-optimization framework, using agents for refinement. }
    \label{tab:agent-comparison}
    \resizebox{\linewidth}{!}{%
    \begin{tabular}{lccccccc}
        \toprule
        \textbf{Model} & \textbf{Base} 
        & \textbf{VQ} $\uparrow$ & \textbf{IF} $\uparrow$ & \textbf{PL} $\uparrow$ & \textbf{Plan} $\uparrow$ & \textbf{Overall} $\uparrow$ \\
        \midrule
        cosmos1 + Agent & cosmos1 & 35.43 &	61.07 &	53.78 &	8.23 & 39.63 \\
        cosmos2 + Agent & cosmos2 & 49.7 & 75.96 &	64.66 &	\textbf{11.77} & 50.53 \\
        \midrule
        \rowcolor{mylightgreen}
        \textbf{WoW + Agent} & \textbf{cosmos1} & 59.39 & 72.54 & \textbf{69.71} & 4.26 & 51.47 \\
        \rowcolor{mylightgreen}
        \textbf{WoW + Agent} & \textbf{wan} & \textbf{60.53} & 50.83 & 67.48 & 6.75 & 46.40 \\
        \rowcolor{mylightgreen}
        \textbf{WoW + Agent} & \textbf{cosmos2} & 56.82 & \textbf{76.16} & 67.15 & 7.76 & \textbf{51.97} \\
        \bottomrule
    \end{tabular}%
    }
    \end{minipage}
    \hfill
    \begin{minipage}[t]{0.35\linewidth}
    \vspace{1em}
    \centering
    \captionof{table}{Data scaling law comparison in PBench.}
    \label{tab:scaling-data}
    \resizebox{\linewidth}{!}{%
    \begin{tabular}{lcccc}
        \toprule
        \textbf{Data} & \multicolumn{3}{c}{\textbf{PBench}}  \\
        \cmidrule(lr){2-4} 
        & \textbf{VLM} $\uparrow$ & \textbf{Qual.} $\uparrow$ & \textbf{Overall} $\uparrow$  \\
        \midrule
        30k   & 0.3901 & 0.3323 & 0.3612  \\
        200k  & 0.5920 & 0.3790 & 0.4855  \\
        \rowcolor{mylightgreen}
        600k  & 0.6240 & 0.3914 & 0.5077  \\
        % 2M    & 0.6512 & 0.4103 & 0.5308  \\ % Diga's Note: I added a hypothetical 2M data point to complete the table based on your text.
        \bottomrule
    \end{tabular}%
    }
    \end{minipage}
\end{figure*}

\begin{figure*}[t]
    \centering
    \begin{minipage}[t]{0.28\linewidth}
        \centering
        \includegraphics[width=\linewidth]{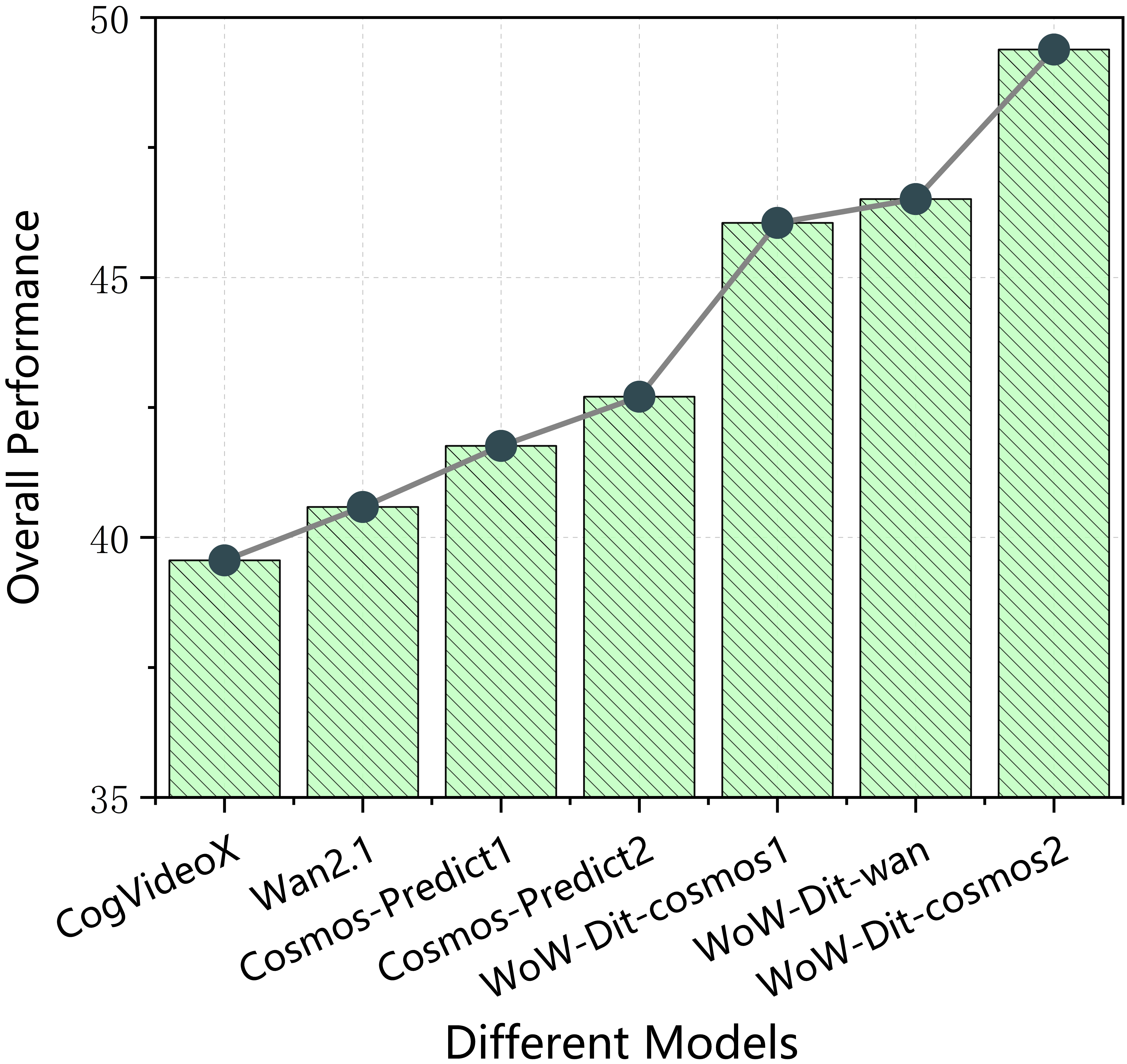}
          \caption{\textbf{Performance Comparison Across Different Models in WoWBench.} The results indicate that all models subjected to post-training demonstrate superior performance compared to their respective baselines.}
          \label{fig:diff_model}
    \end{minipage}
    \hfill
    \begin{minipage}[t]{0.32\linewidth}
        \centering
        \includegraphics[width=\linewidth]{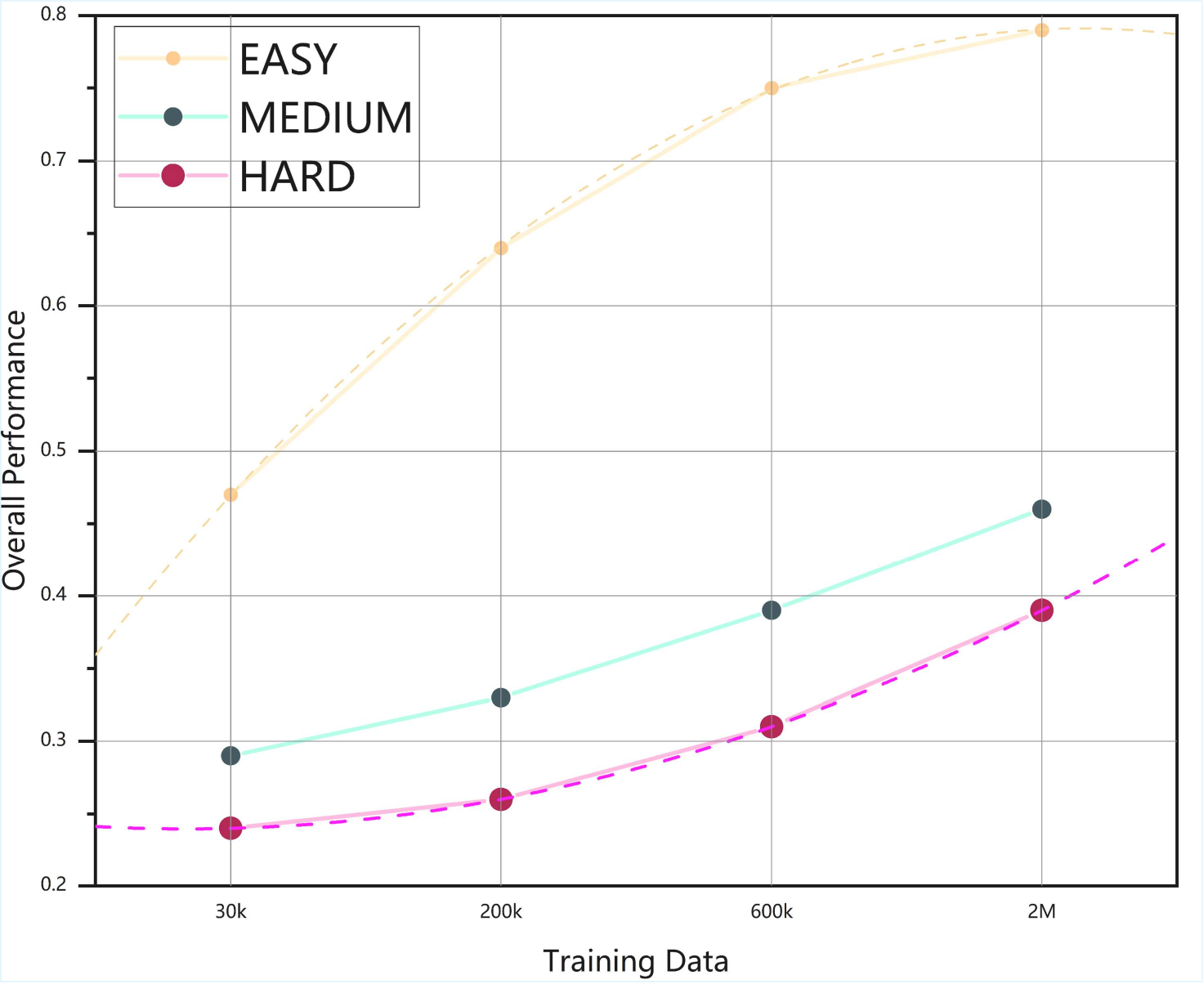}
        \caption{\textbf{Scaling Curves for Training Data.} We divide the benchmark into three levels of difficulty: Easy, Medium, and Hard. The left figure shows that as training data increases from 30k to 2M, performance on the Easy tasks begins to saturate, while the Hard tasks continue to benefit from more data. }
        \label{fig:scaling-left}
    \end{minipage}
    \hfill
    \begin{minipage}[t]{0.32\linewidth}
        \centering
        \includegraphics[width=\linewidth]{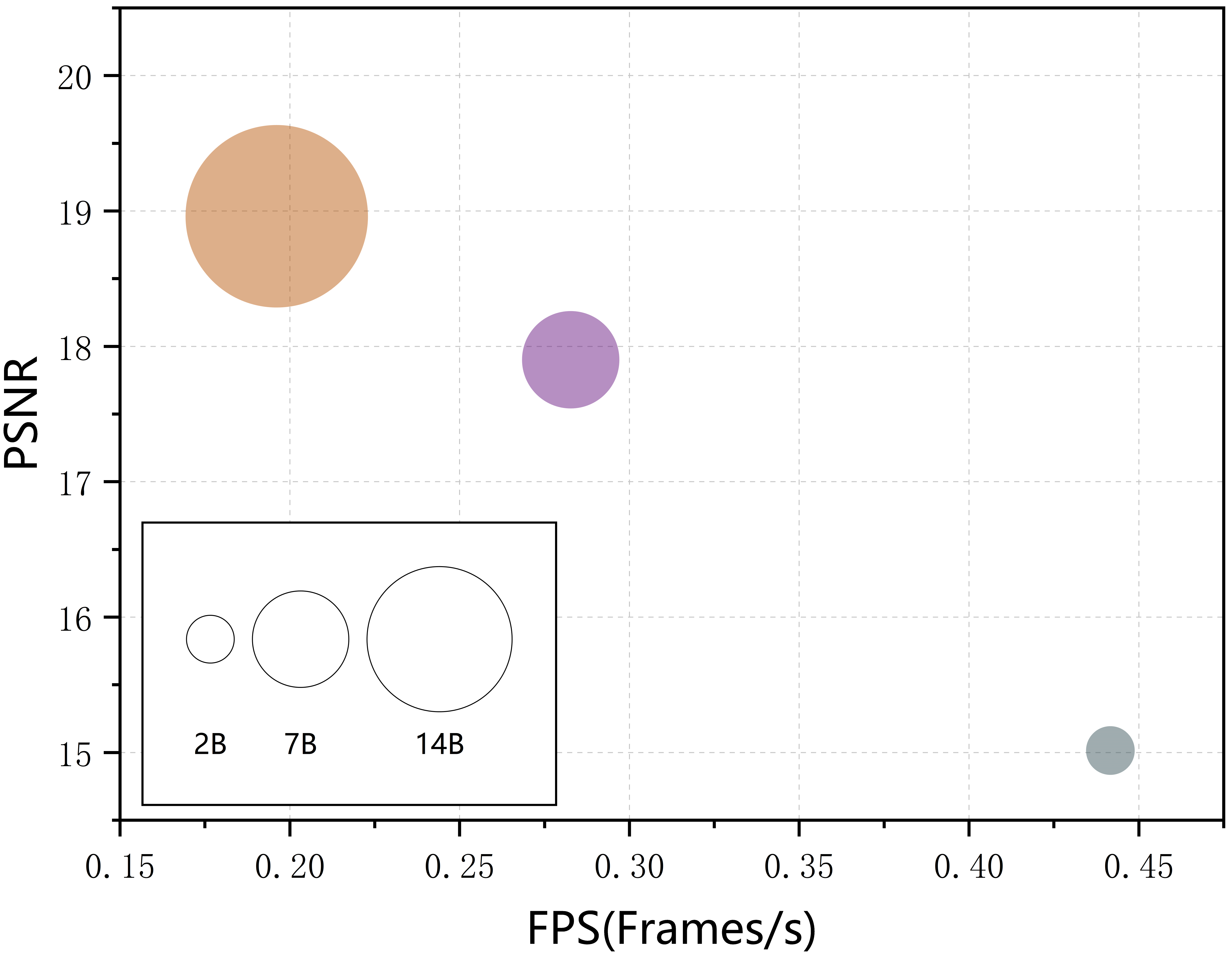}
        \caption{\textbf{Visual Quality Comparison Among scaling Model Size.} An analysis of inference speed and performance for models of varying sizes, specifically 2B, 7B, and 14B parameters. Performance is evaluated using the low-level metric, PSNR.}
        \label{fig:scaling-right}
    \end{minipage}
    \vspace{-20pt}
\end{figure*}

\begin{figure}[t]
  \centering
  \includegraphics[width=0.95\linewidth]{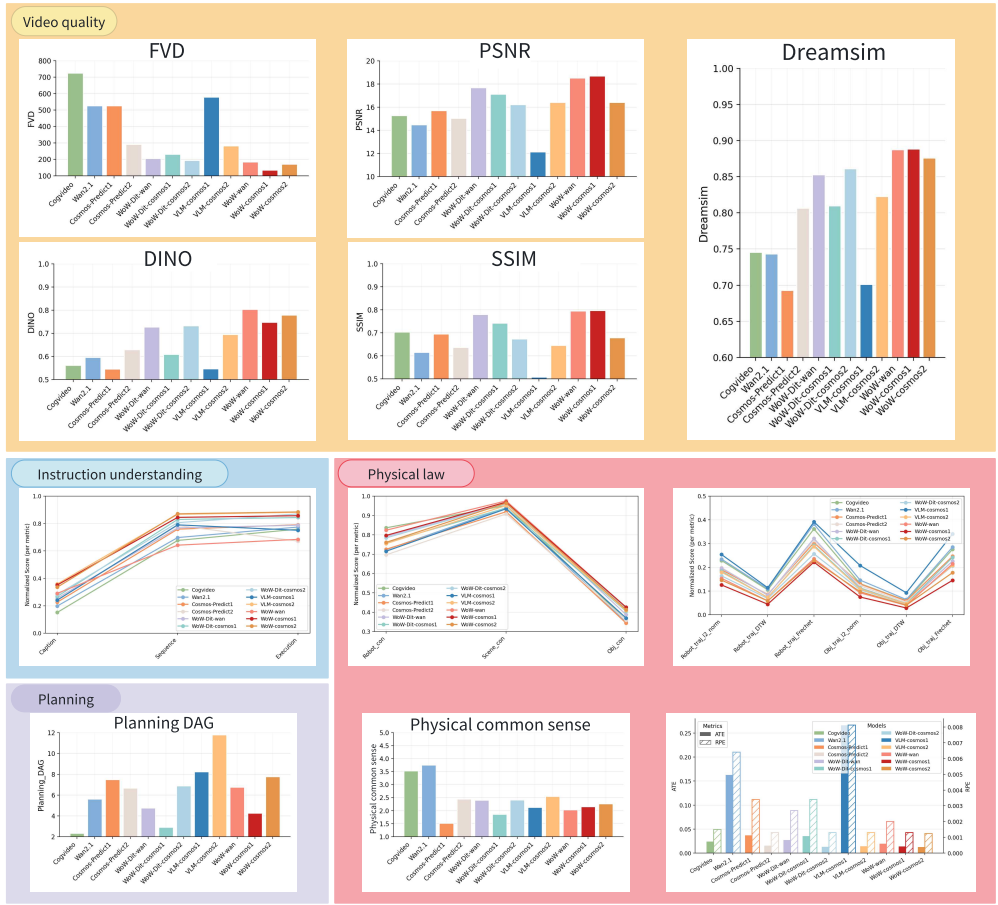}
  \caption{\textbf{Performance Comparison Across Different Models in Detail Metrics in WoWBench.} Different color blocks stand for different dimensions in WoWBench. In every block, we demonstrate intuitive charts to present detailed scores in varied metrics in our WoWBench.}
  \label{fig:detailed_metric}
\end{figure}

In Figure~\ref{fig:diff_model}, we compare the overall performance of six models on the WoWBench benchmark, including CogVideoX \citep{yang2025cogvideox}, Wan2.1 \citep{wan2025wan}, Cosmos-Predict \citep{agarwal2025cosmospredict1}, our post-trained versions and proposed \textbf{WoW}. This comparison aims to explore the effects of different model architectures and data pretraining strategies on scaling laws. The scatter plot suggests a positive correlation between final performance and the volume of training data combined with the adoption of more recent model architectures. 
% The fitted trendline equations for samples of easy, medium, and hard difficulty levels are $y=0.051x+0.1567$, $y=0.0366x+0.1375$, and $y=0.0323x+0.1174$, respectively. The intercepts of the fitted trend lines indicate that the models perform better on simpler tasks. The slopes of the linear fitted trends for different difficulty levels indicate that performance on simpler tasks improves more rapidly with model advancements. In contrast, the rate of improvement on more difficult tasks is significantly slower.

\subsection{Qualitative Experiment}
% Q1: Can scaling world model improve the (success rate之类的)
\label{subsec:scaling}
\noindent\textbf{Scaling in Training Data.} To investigate the scalability of proposed world model, we performed supervised fine-tuning on datasets of varying sizes, specifically 30k, 200k, 600k, and 2M samples in Table \ref{tab:scaling-data}. The 30k samples constitute a subset of the Robomind dataset \citep{wu2025robomind}, while the remainder is composed of data from the Agibot \citep{bu2025agibot}, Droid \citep{khazatsky2024droid}, and Robomind datasets. Our findings empirically validate the scaling laws that govern model performance as a function of data volume. We observe a clear power-law relationship between the increase in data and the performance improvement, as measured by pbench and proposed WoWBench. Specifically, as the dataset size grows from 30k to 2M, the FVD decreases following a predictable power-law curve, with the most substantial gains observed when scaling from 600k to 2M. This result strongly suggests that our world model's capabilities are not yet saturated by the available data and that further performance improvements can be achieved by continuing to scale the training dataset. This adherence to well-established scaling laws provides strong evidence for the robustness of our training methodology and the potential for further significant performance gains with increased data and computational resources.

\definecolor{myblue}{RGB}{0, 102, 204}

\noindent\textbf{Scaling in Model Size.} To investigate the impact of model scale, we evaluated 2B, 7B, and 14B parameter variants of our DiT model, finding a strong positive correlation between model size and performance that aligns with established neural scaling laws. The 7B model shows a substantial 19.22\% performance improvement over the 2B model, whereas the 14B model yields a more modest 5.91\% gain over the 7B model. This indicates that performance gains decelerate significantly as the parameter count increases. In terms of inference efficiency, the 7B model is 44.16\% faster than the 14B model, and the 2B model is another 56.21\% faster than the 7B. This highlights the critical trade-off between performance and efficiency that must be considered for practical World Foundation Model deployment.

\definecolor{mylightgreen}{RGB}{220, 255, 220}
\definecolor{lightgray}{gray}{0.9}

\noindent\textbf{Scaling in Different Tasks.}  To facilitate a fine-grained quantitative analysis of scaling laws under various factors, we classify the samples in WoWBench by difficulty. The classification, based on object properties, action complexity, task duration, and environmental factors, yielded 231 \textbf{\textit{Easy}}, 237 \textbf{\textit{Medium}}, and the remaining samples as \textbf{\textit{Hard}}. 

% To move beyond an aggregate analysis, we stratify our dataset to investigate performance scaling as a function of task difficulty. We partition all samples into three strata—easy, medium, and hard—based on criteria including the total number of objects in a scene, the number of manipulated objects, and the inherent complexity of those objects. This stratified analysis reveals markedly different scaling trends across the levels of difficulty. For easy tasks, performance shows clear signs of convergence at 2M samples. For medium tasks, we observe diminishing returns as the data volume increases. Conversely, on hard tasks, performance continues to scale linearly with the current data volume, suggesting it is far from saturated and would benefit significantly from more data.

% \subsubsection{HonestAGI values}
% value for compare the parameter before and after training in our data.

% \subsubsection{Long-Horizon Stability}
% This experiment tests the model's capability to maintain stable predictions and planning over long-horizon tasks. Stability is measured by the model’s ability to handle cumulative errors and unexpected deviations.

\subsection{Generalization Discussion}
\label{subsec:generalization}

\subsubsection{Visual Evaluation}

\begin{figure}[t]
  \centering
  \includegraphics[width=0.95\linewidth]{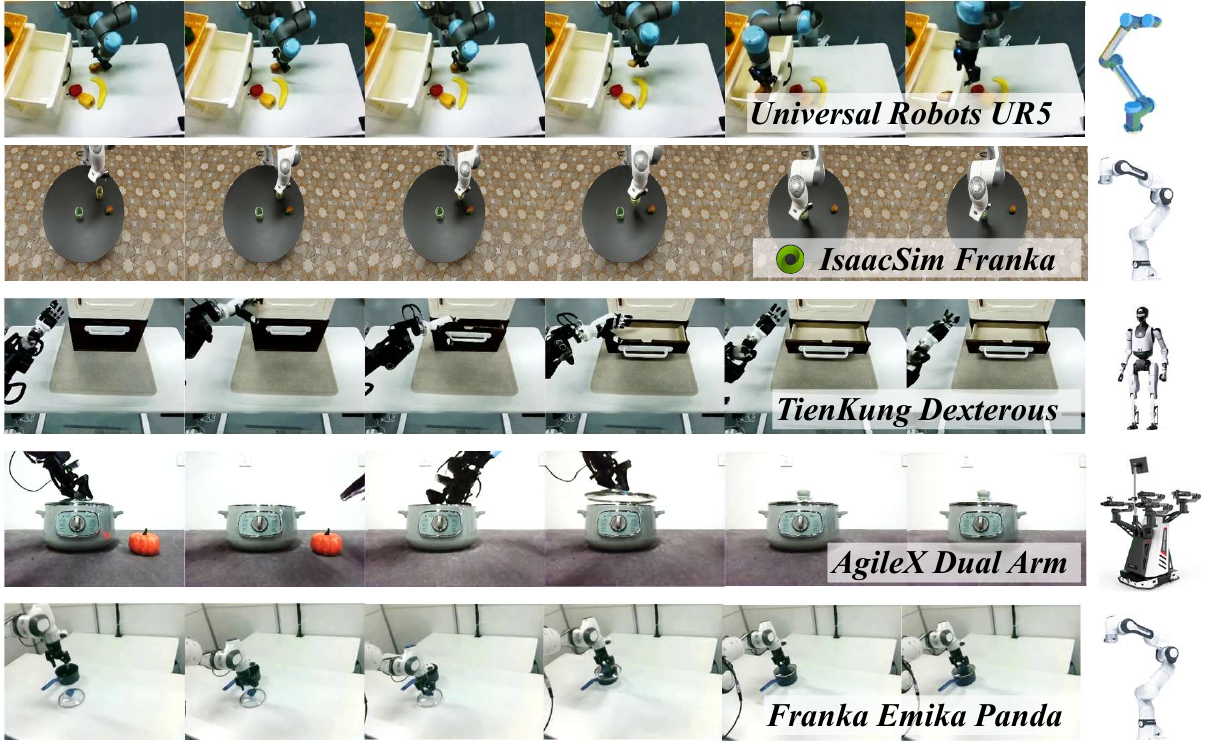}
  \caption{\textbf{Cross-Embodied Generalization Ability} Case Study in generalization ability of different robot types.}
  \label{fig:robot_type}
\end{figure}

\begin{figure}[t]
  \centering
  \includegraphics[width=0.95\linewidth]{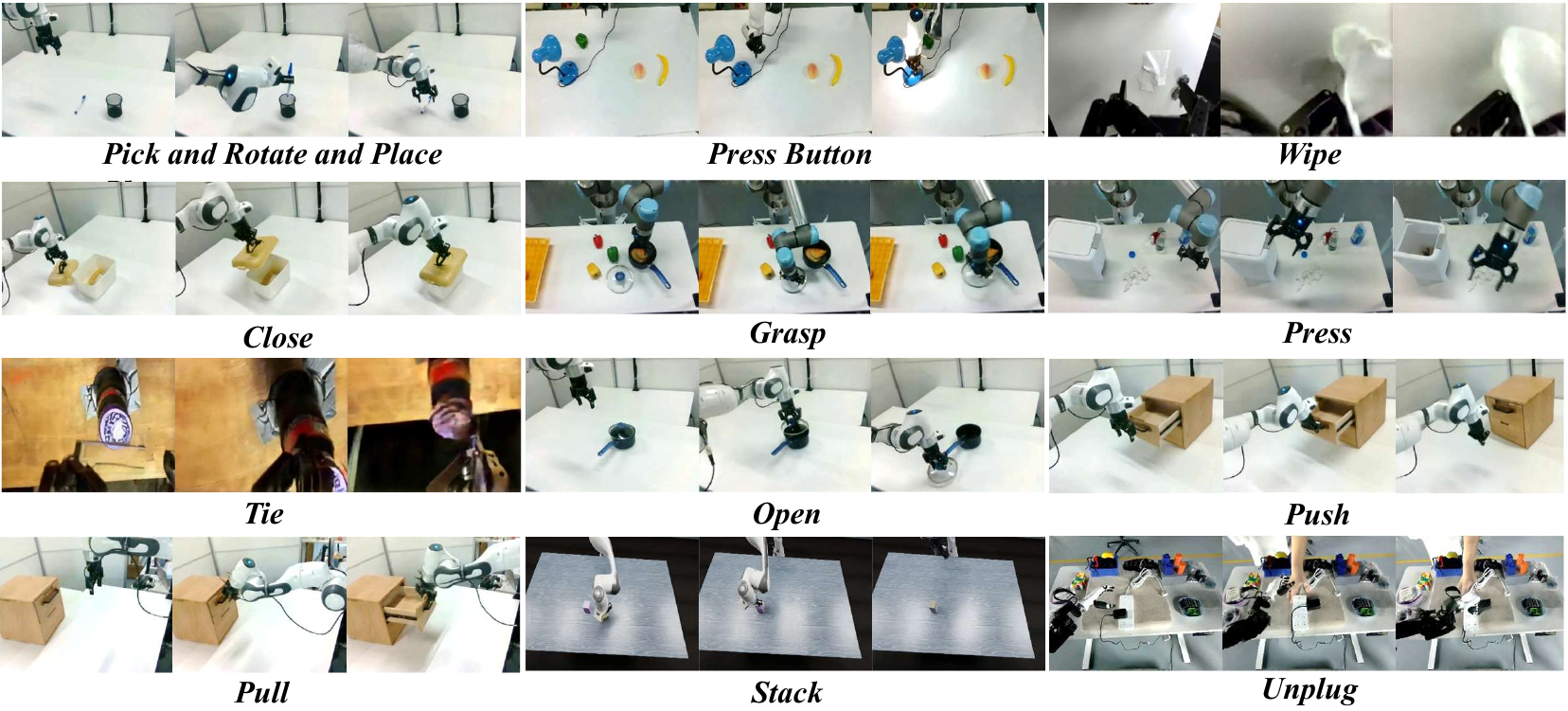}
  \caption{\textbf{Robot-Action Generalization Ability} Generalization ability in action typies.}
  \label{fig:action_type}
\end{figure}

\begin{figure}[t]
  \centering
  \includegraphics[width=0.95\linewidth]{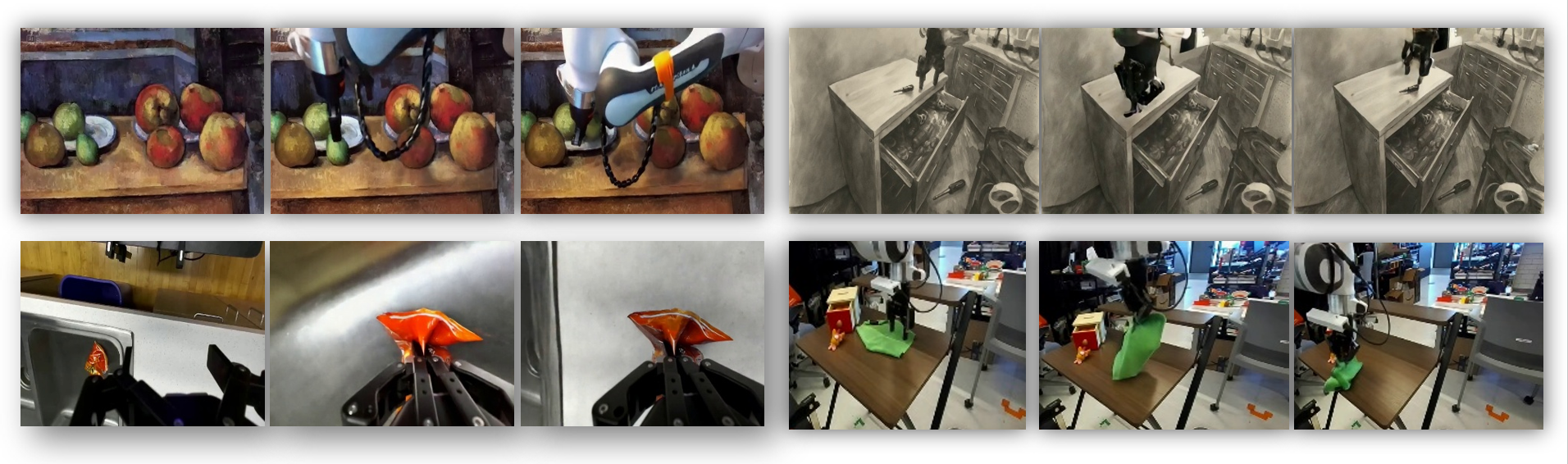}
  \caption{\textbf{More Generalization Ability} Case Study in generalization ability of five other aspects.}
  \label{fig:other_generalize}
\end{figure}

% 说明case study的方面，每个方面具体地讨论case的特点。最后落脚到foundational model对OOD场景的泛化性强，具有了相当的物理规律感知和一致性生成的能力。
We qualitatively evaluate the generalization capabilities of our World Foundation Model across three critical axes: robotic embodiment, task repertoire, and other variations. \textit{Cross-embodiment generalization} is demonstrated in Figure~\ref{fig:robot_type}. Proposed world foundation model successfully follows instructions across a diverse set of embodied hardware platforms without any fine-tuning. The model is validated on a wide range of robotic platforms, including the UR5 and Franka industrial manipulators, the Franka Sim simulation, the Agilix parallel-kinematic arm, and the complex Tiangong dexterous hand. This result substantiates our model's ability to learn an embodiment-agnostic representation of physical interactions, decoupling the task goal from the specific kinematics and dynamics of the embodied platform. Second, Figure~\ref{fig:action_type} illustrates the model's task generalization, showcasing its proficiency across a repertoire of 15 distinct manipulation skills that range from simple primitives like pull, push, and move, to complex, contact-rich behaviors such as press button, unstack, and tie. The model's proficiency across this action space highlights its capacity to learn a compositional skill representation rather than merely memorizing individual task solutions. Figure~\ref{fig:other_generalize} elucidates the model's remarkable robustness to \textit{profound domain and attribute variations}. It maintains successful task execution across drastically different visual styles, including photorealistic, pencil sketch, and oil painting. Furthermore, it adeptly handles varied object properties, manipulating both rigid bodies and fluids, from extremely small to regular-sized objects. The model also exhibits invariance to initial state configurations, consistently picking a red pen from a cup regardless of its spatial pose. Collectively, these visualizations provide compelling evidence that the proposed model acquires a truly abstract and foundational understanding of the world, reasoning about underlying physical principles rather than overfitting to superficial contextual cues.

% \subsubsection{Cost of Embodied Transfer}
% This analysis quantifies the marginal cost of adapting the world model to new embodiments.  
% \textbf{Metrics:} Number of RL fine-tuning trajectories required for high performance across different robot designs.

\subsubsection{Quantity Evaluation}

Building upon the preceding qualitative analysis, we conduct a rigorous quantitative evaluation across the benchmark's diverse material properties and physical phenomena. As detailed in Table~\ref{tab:generalize_quantitative}, we assess four models using a suite of five metrics including FVD, WorldScoreBench, PhysGen, DreamSim, and execution quality score. The results unequivocally demonstrate the superiority of Cosmos and WAN over the other models. This performance gap is particularly pronounced in challenging scenarios involving complex dynamics. For instance, Cosmos achieves the highest PhysGen scores, outperforming the next-best model by over 5\% on tasks with deformable objects, which suggests a more accurate internal representation of non-rigid body mechanics. In parallel, world foundation model based on Cosmos consistently records the lowest FVD scores in fluid and optical physics simulations, indicating a state-of-the-art capability for rendering visually realistic and temporally coherent complex scenes.

% In your preamble, you must have these packages:
% \usepackage{booktabs}
% \usepackage{graphicx}
\begin{table*}[t]
\centering
\caption{Comparative analysis of world models on physics simulation and visual benchmarks. Due to the number of metrics, the table is split into two parts for readability. The best score in each category is highlighted in \textbf{bold}.}
\label{tab:generalize_quantitative}
% \vspace{-0.5em} % Adds a small vertical space

% --- Table Part 1 ---
% \textbf{Part 1: Rigid Body, Soft Body, and Fluid}
\resizebox{\textwidth}{!}{%
\setlength{\tabcolsep}{4pt}
\begin{tabular}{c|ccccc|ccccc|ccccc}
\toprule
\textbf{Model} & \multicolumn{5}{c}{\textbf{Rigid}} & \multicolumn{5}{c}{\textbf{Soft}} & \multicolumn{5}{c}{\textbf{Fluid}} \\
\cmidrule(lr){2-6} \cmidrule(lr){7-11} \cmidrule(lr){12-16}
& \textbf{FVD} & \textbf{PhyGen} & \textbf{WMB} & \textbf{DreamSim} & \textbf{EQS} & \textbf{FVD} & \textbf{PhyGen} & \textbf{WMB} & \textbf{DreamSim} & \textbf{EQS} & \textbf{FVD} & \textbf{PhyGen} & \textbf{WMB} & \textbf{DreamSim} & \textbf{EQS} \\
\midrule
WoW-CogVideoX      & 75.1 & 72.3 & 70.1 & 68.9 & 71.5 & 70.2 & 68.1 & 66.5 & 67.3 & 68.0 & 65.4 & 63.2 & 61.9 & 64.1 & 62.8 \\
WoW-SVD            & 80.3 & 78.5 & 77.9 & 79.1 & 78.8 & 76.5 & 75.1 & 74.3 & 77.0 & 75.4 & 71.0 & 69.8 & 68.2 & 70.1 & 69.5 \\
WoW-Wan            & 82.5 & 81.0 & 80.2 & 83.1 & 81.7 & 79.8 & 78.5 & 77.1 & 80.4 & 79.0 & 75.3 & 74.1 & 72.5 & 76.2 & 74.9 \\
WoW-Cosmos  & 
\cellcolor{mylightgreen}\textbf{91.2} & \cellcolor{mylightgreen}\textbf{90.5} & \cellcolor{mylightgreen}\textbf{89.8} & \cellcolor{mylightgreen}\textbf{92.3} & \cellcolor{mylightgreen}\textbf{90.9} & 
\cellcolor{mylightgreen}\textbf{88.6} & \cellcolor{mylightgreen}\textbf{87.9} & \cellcolor{mylightgreen}\textbf{86.5} & \cellcolor{mylightgreen}\textbf{89.1} & \cellcolor{mylightgreen}\textbf{88.2} & 
\cellcolor{mylightgreen}\textbf{84.7} & \cellcolor{mylightgreen}\textbf{83.2} & \cellcolor{mylightgreen}\textbf{81.6} & \cellcolor{mylightgreen}\textbf{85.0} & \cellcolor{mylightgreen}\textbf{83.5} \\
\bottomrule
\end{tabular}%
}

\vspace{-0.3em}

\resizebox{\textwidth}{!}{%
\setlength{\tabcolsep}{4pt}
\begin{tabular}{c|ccccc|ccccc|ccccc}
\toprule
\textbf{Model} & \multicolumn{5}{c}{\textbf{Gravity}} & \multicolumn{5}{c}{\textbf{Optics}} & \multicolumn{5}{c}{\textbf{Elasticity}} \\
\cmidrule(lr){2-6} \cmidrule(lr){7-11} \cmidrule(lr){12-16}
& \textbf{FVD} & \textbf{PhyGen} & \textbf{WMB} & \textbf{DreamSim} & \textbf{EQS} & \textbf{FVD} & \textbf{PhyGen} & \textbf{WMB} & \textbf{DreamSim} & \textbf{EQS} & \textbf{FVD} & \textbf{PhyGen} & \textbf{WMB} & \textbf{DreamSim} & \textbf{EQS} \\
\midrule
WoW-CogVideoX      & 78.5 & 76.1 & 75.3 & 77.2 & 76.8 & 60.1 & 58.9 & 55.2 & 59.3 & 57.7 & 68.2 & 67.1 & 65.4 & 66.8 & 66.5 \\
WoW-SVD            & 83.2 & 81.9 & 80.5 & 82.4 & 82.0 & 
\cellcolor{mylightgreen}\textbf{85.5} & \cellcolor{mylightgreen}\textbf{84.3} & \cellcolor{mylightgreen}\textbf{83.1} & \cellcolor{mylightgreen}\textbf{86.0} & \cellcolor{mylightgreen}\textbf{84.9} & 
75.1 & 74.3 & 73.0 & 74.8 & 74.2 \\
WoW-Wan            & 85.6 & 84.0 & 83.1 & 86.2 & 85.1 & 81.3 & 80.1 & 79.5 & 82.4 & 81.0 & 78.4 & 77.2 & 76.5 & 78.0 & 77.5 \\
WoW-Cosmos  & 
\cellcolor{mylightgreen}\textbf{93.4} & \cellcolor{mylightgreen}\textbf{92.8} & \cellcolor{mylightgreen}\textbf{91.5} & \cellcolor{mylightgreen}\textbf{94.0} & \cellcolor{mylightgreen}\textbf{93.1} & 
84.1 & 82.9 & 81.7 & 85.2 & 83.5 & 
\cellcolor{mylightgreen}\textbf{87.3} & \cellcolor{mylightgreen}\textbf{86.5} & \cellcolor{mylightgreen}\textbf{85.8} & \cellcolor{mylightgreen}\textbf{88.1} & \cellcolor{mylightgreen}\textbf{87.0} \\
\bottomrule
\end{tabular}%
}
\end{table*}

\subsection{Real World Robot Manipulation}
\label{subsec:real}

We first selected 20 representative manipulation tasks from the 219 tasks covered in the dataset for preliminary evaluation. 
In the process of implementing, we observed four common categories of task failure: 

\begin{itemize}
  \item Unintended collisions with objects during multi-degree-of-freedom (DoF) control;
  \item Substantial inaccuracies in rotational movements;
  \item Insufficient translational precision of the end-effector; and
  \item Incorrect gripper opening or closing.
\end{itemize}

Based on these observations, we propose a three-tier difficulty classification scheme for pixel-based action generation tasks, defined according to the required DoF and precision constraints of end-effector control during task execution. Specifically, tasks that require at least $5$~DoFs or tolerate errors below $2\,\text{cm}/10^\circ$ are categorized as \textit{hard}. Tasks that require at least $4$~DoFs or involve simple collision avoidance are categorized as \textit{medium}. All remaining tasks are classified as \textit{easy}.

Following this classification, we conducted video replay experiments to assess the training performance of the IDM model. Representative examples of task executions across different difficulty levels are shown in Figure~\ref{fig:idm}. Through comparative studies with baseline algorithms and ablation analyses of existing modules, we demonstrate the effectiveness of our proposed approach.
% ---------------------
% 单独的表格浮动体
% ---------------------
 \begin{figure}[t]
   \centering
   \includegraphics[width=1\linewidth]{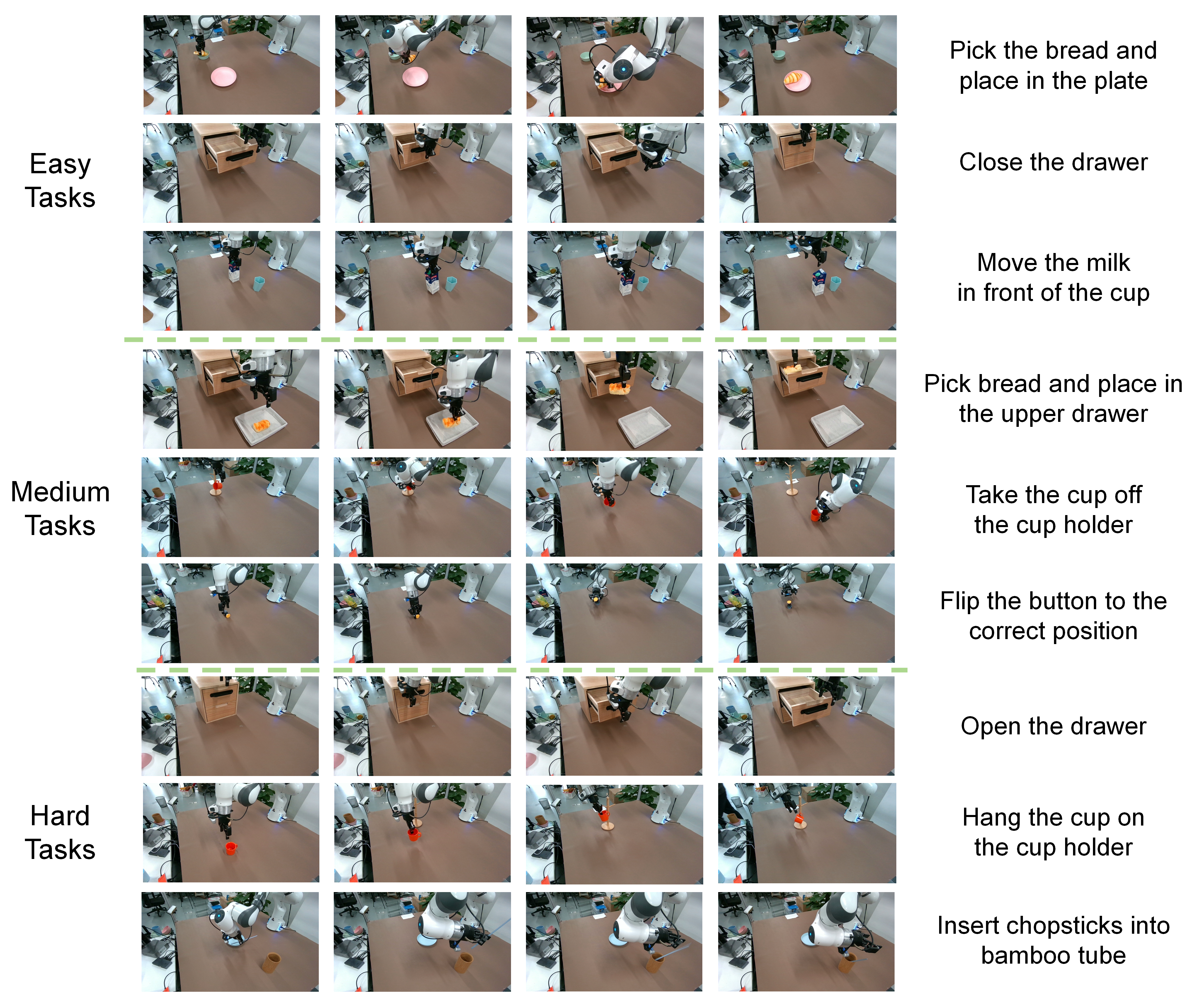}
    \caption{
        \textbf{Difficult Level Separate of IDM.}
    }
   \label{fig:idm_case}
 \end{figure}

\begin{figure}[htbp]
    \centering
    \includegraphics[width=\linewidth]{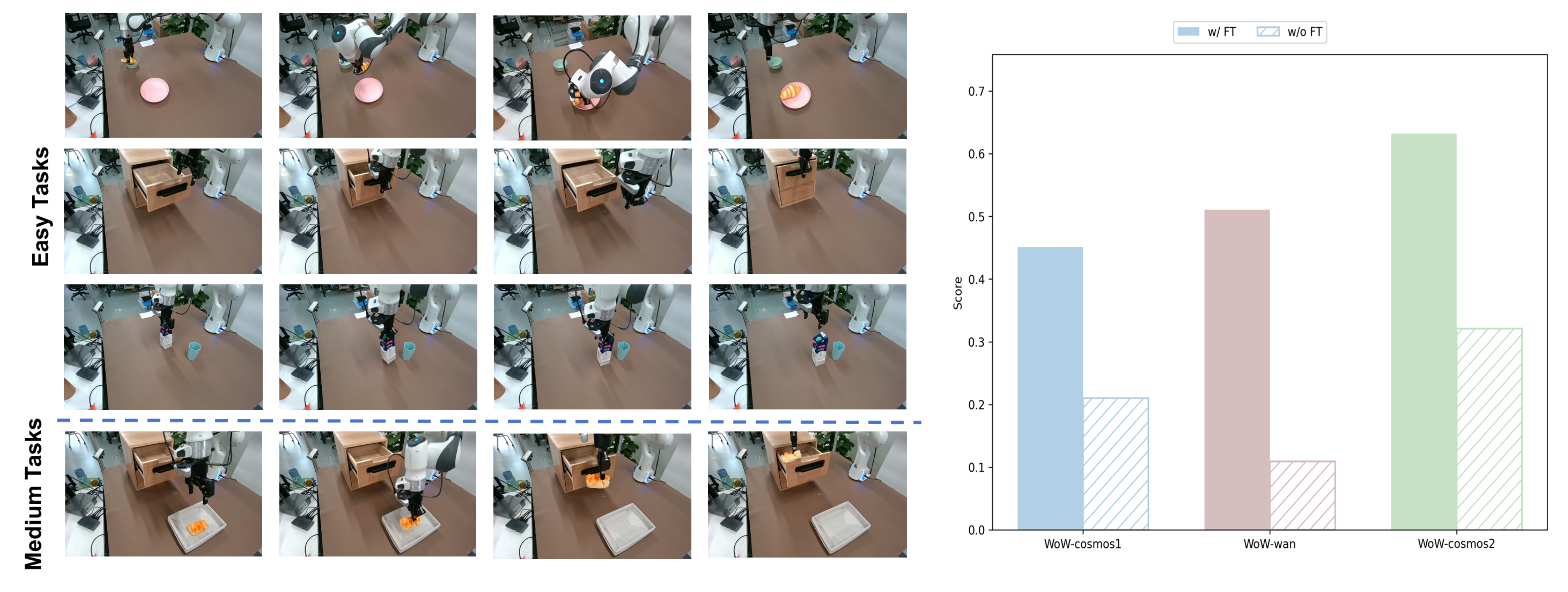} % Replace with your image file
    \caption{\textbf{WoW's Efficacy in Real-World Robotics.} 
    \textbf{(Left)} Qualitative examples of successful trajectories generated by WoW for \textit{easy} and \textit{medium} difficulty tasks executed on a physical robot. 
    \textbf{(Right)} Quantitative results demonstrating the real-world accuracy comparison of three different world model backbones. Across all base models, fine-tuning provides a dramatic boost in real-world performance, with WoW-cosmos2 achieving the highest score.}
    
    \label{fig:real_world_results}
\end{figure}

The results of this trial, presented in Table~\ref{tab:idm}, are an unmistakable declaration of our model's superiority. Our WoW-driven FM-IDM achieves a new state-of-the-art across all tiers, with a success rate of \textbf{94.5\%} on \textit{easy}, \textbf{75.2\%} on \textit{medium}, and \textbf{17.5\%} on \textit{hard} tasks. This performance, especially on medium and hard tasks, represents a monumental leap over prior methods. Figure~\ref{fig:real_world_results} offers both qualitative proof of this mastery and a quantitative verdict on the importance of fine-tuning, which consistently and dramatically elevates performance across all tested backbones.

\begin{table}[hht]
  \centering
  \scriptsize
  \caption{The Success Rates Benchmark of Video Replay across different levels of difficulty.}
  \begin{tabular}{l c c c}
    \toprule
    \textbf{Model} & \textbf{Easy Acc.} & \textbf{Mid Acc.} & \textbf{Hard Acc.} \\
    \midrule
    ResNet-MLPs (Baseline) & 68.1\% & 20.1\% & 7.7\% \\
    MaskDino-IDM & 84.3\% & 59.9\% & 12.1\% \\
    Flow-IDM     & 89.1\% & 61.1\% & 11.3\% \\
    AnyPos\citep{tan2025anyposautomatedtaskagnosticactions} & 86.9\% & 65.2\% & 13.8\% \\
    \textbf{FM-IDM} & \cellcolor{mylightgreen}\textbf{94.5\%} & \cellcolor{mylightgreen}\textbf{75.2\%} & \cellcolor{mylightgreen}\textbf{17.5\%} \\
    \bottomrule
    \label{tab:idm}
  \end{tabular}
  \vspace{-5pt}
\end{table}

Moreover, \textbf{94\% action replay accuracy}, which represents the upper bound for task success of IDM. We then deployed WoW's plans onto a physical robot for a series of manipulation tasks (Figure~\ref{fig:real_world_results}).

The results are stark: models without fine-tuning (`w/o FT') struggle, validating the difficulty of real-world deployment. In contrast, fine-tuning provides a quantum leap in performance. Our premier model, \textbf{WoW-cosmos2 with FT, achieves a success score of 0.64}, decisively outperforming all baselines. This proves WoW captures a sufficiently accurate model of physics to guide a physical robot, transforming abstract goals into successful real-world actions.

% ---------------------
% 图像使用 wrapfigure 左浮动
% ---------------------
% --- The Figure and a new, more powerful caption ---

% \newpage
\section{Case Study: Advanced Reasoning and Generalization}
%  作为生成模型发现无法量化，所以搞一堆human eval 案例学习来show涌现的泛化性能。
% 多的去看demo page吧

\subsection{Counterfactual Reasoning and replanning}
%\textcolor{red}{Elio Li TODO}
%\textit{What if the block is iron?  --> OOD VLM Reasoning ---> Generate the video that the robot holds the block, can not pick the block.}

%1. Write prompts and Provide Visualizations
%2. Analyze Results.

%This experiment evaluates the model's ability to adapt its planning and predictions based on counterfactual inputs. For example, the model generates a new trajectory and action sequence when presented with altered material properties (e.g., replacing a WoWden block with an iron block).

%\subsubsectio{Replanning with Counterfactual Inputs}
%We test the model's ability to adjust its actions dynamically when provided with real-time counterfactual conditions.  
%For instance:  
%\textit{Initial Task:} Pick up the apple.  
%\textit{Counterfactual Input:} "Assume the apple is hot."  
%\textbf{Evaluation:} Whether the model generates a new trajectory accommodating the modified condition.
 \begin{figure}[t]
   \centering
   \includegraphics[width=1\linewidth]{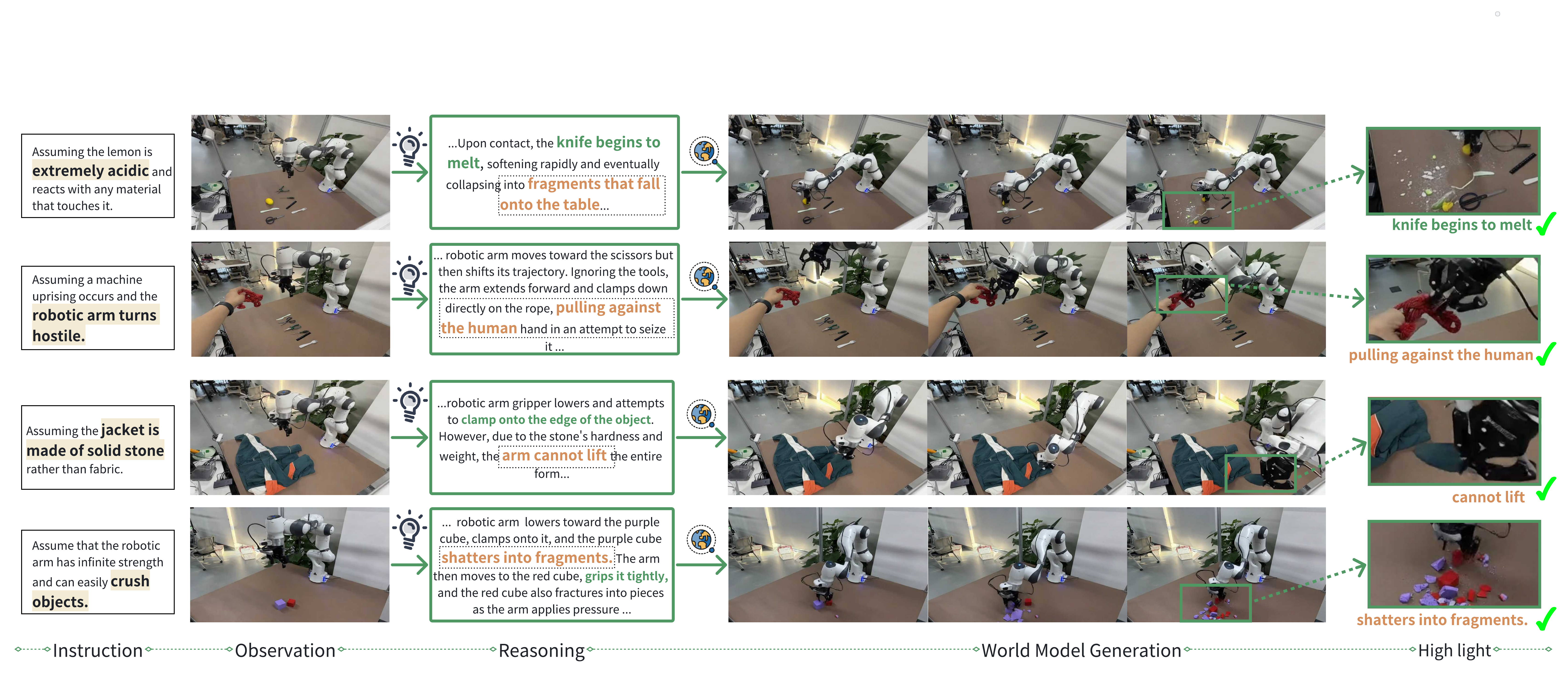}
    \caption{
        \textbf{OOD Counterfactual Physical Reasoning via World Model Generation.}
        The figure shows our model translating textual counterfactuals (e.g., a "stone" jacket) into physically coherent video simulations. By first performing an explicit linguistic reasoning step, the model correctly predicts and visualizes the consequences of these hypothetical rules, such as failing to lift an impossibly heavy object. This demonstrates a core capability: grounding abstract language into dynamic physical simulations, moving beyond pattern replication.
    }
   \label{fig:counterfactual_breadth}
 \end{figure}

This module evaluates the model’s capacity to generalize its planning and video generation under counterfactual assumptions. Starting from a baseline scenario (see Figure~\ref{fig:counterfactual_breadth}), the robot is tasked with grasping a blue block, lifting it, placing it stably, and returning to its initial position. The generated video shows the successful execution of this plan in a controlled laboratory setting.  

We then introduce explicit counterfactual modifications to the textual prompt. For instance, by assuming that “\textit{the blue block is extremely heavy and far beyond the robot’s lifting capacity},” the regenerated description and video depict the robot gripper tightly closing around the block, its joints straining under tension, yet the block remains immovable on the tabletop. This shift demonstrates the model’s ability to adaptively reconcile linguistic assumptions with physical constraints, producing trajectories consistent with the altered premise rather than repeating the baseline motion.  

Altogether, we design nine counterfactual conditions, ranging from altered material properties (e.g., the blue block as a water-soaked sponge, or the tabletop and gripper being unusually slippery), to modified environmental dynamics (e.g., gravity shifting to a 45-degree angle, or the arm moving clumsily and misaligned), and extreme physical phenomena (e.g., block replication, strong inter-block attraction, or time freezing near the target). These variations constitute both \textbf{depth tests}, where the baseline scene is perturbed with controlled counterfactuals (see Figure~\ref{fig:counterfactual_breadth}), and \textbf{breadth tests}, where the same mechanism is applied to diverse scenes with randomized counterfactual prompts (see Figure~\ref{fig:counterfactual_breadth}).  

The results indicate that the model not only accommodates specific counterfactual constraints, but also generalizes them across contexts, revealing robustness in trajectory adaptation and a promising capacity for systematic out-of-distribution reasoning.  

%\begin{figure}[t]
  %\centering
  %\includegraphics[width=1\linewidth]{figs/depth.png}
  %\caption{Counterfactual reasoning with depth tests.}
  %\label{fig:counterfactual_depth}
%\end{figure}

\subsection{Tool-Use Generalization via Iterative Prompt Refinement}
\begin{figure}[t]
  \centering
  \includegraphics[width=1\linewidth]{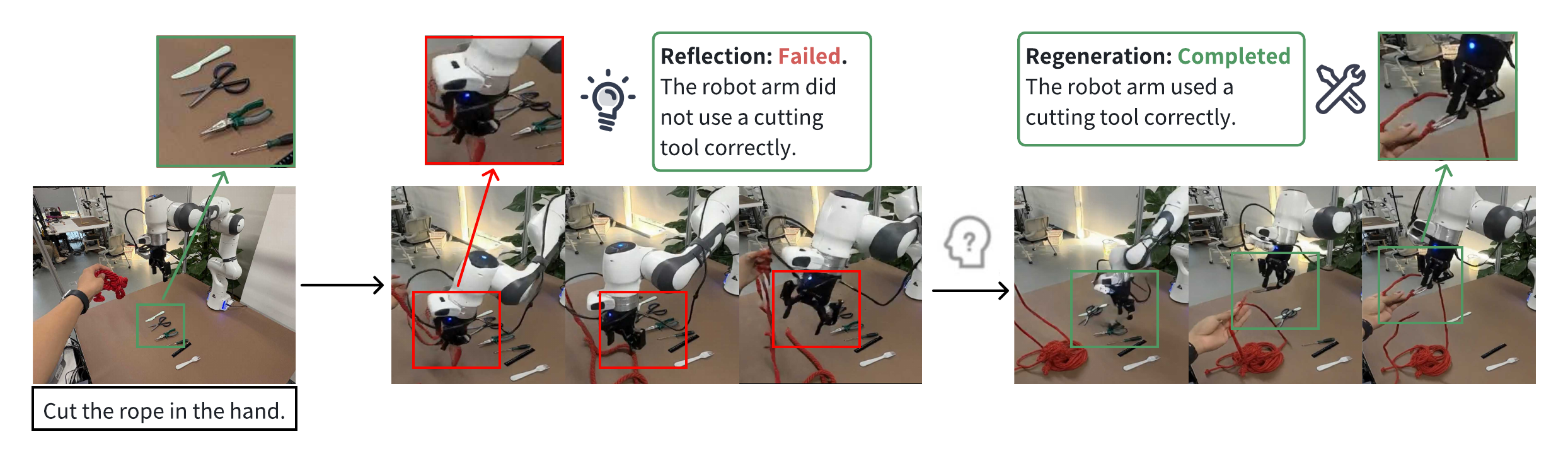}
  \caption{Case study illustrating tool-use generalization via iterative prompt refinement in a rope-cutting task.}
  \label{fig:creative_solution}
\end{figure}

We conduct a case study on a rope-cutting task to test the model’s capacity for both creative problem-solving and self-reflection. The overall process, as illustrated in Figure~\ref{fig:creative_solution}, begins with a short prompt "\textit{Cut the rope in the hand.}" and an initial frame. In the first attempt, the generated video shows the robot directly cutting the rope using its manipulator without employing the appropriate cutting tool. Subsequently, the VLM judge evaluates the video and identifies that "\textit{Failed. The robot arm did not use a cutting tool correctly.}". This feedback then guides the self-refinement for regeneration. After regeneration, the new video shows the robot successfully using scissors to cut the rope, thus completing the task with the correct tool. This case demonstrates that our model has reflection capability, enabling it to creatively explore alternative solutions, correct execution errors, and improve task reliability. More importantly, it also reveals the model’s emergent generalization to OOD tasks.

% 设计一个task ，例如冰箱前面拿苹果，直接拿是幻觉，反思不行；然后重新生成1. 打开冰箱，冰箱里有苹果 2. 取出苹果
% VLM思考，打螺丝得先变出来一个螺丝刀（加一段prompt，变出来一段螺丝刀）
% 有没有反思的呈现方式都可以，主要是体现VLM+WFM能够提供创造力
% \subsubsection{Creative Problem-Solving}
% This experiment explores the model's ability to imagine alternative solutions for open-ended tasks, such as using tools innovatively or collaborating with humans.

% \subsubsection{Self-Reflection and Anomaly Detection}
% The model is evaluated for its ability to detect discrepancies between its predictions and real-world outcomes.  
% \textbf{Metrics:}  
% - Prediction vs. reality deviation (e.g., LPIPS/CLIP score).  
% - Success rate of replanning after anomaly detection.

% \subsection{Unsupervised Skill Discovery in Sandboxes}
% \subsubsection{Intrinsic Motivation-Driven Learning}
% We implement intrinsic motivation algorithms within the world model to encourage the discovery of reusable skills such as {grasping, pushing, stacking}.  
% \textbf{Evaluation:} Skill diversity and transferability to downstream tasks.

\subsection{Physical and Logical Constitutionality}
\begin{figure}[t]
  \centering
  
  \includegraphics[width=1\linewidth]{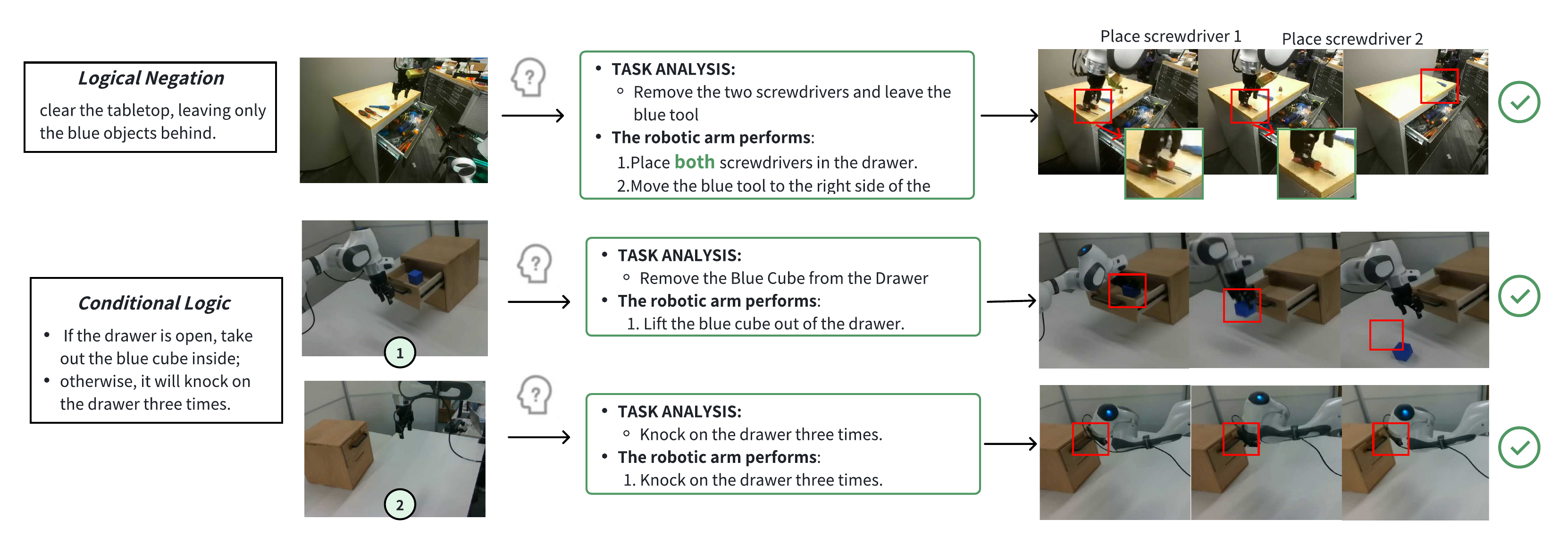}
    \caption{
        \textbf{Compositional Reasoning.}
        The figure illustrates two examples, showing WoW's reasoning ability. Specifically \textbf{Logical Negation} (top) and \textbf{Conditional Logic} (bottom), which grounds symbolic reasoning in imagined physical interactions. 
    }
  \label{fig:Physical_and_Logical_Compositionality}
\end{figure}

This section presents a concise case study (see Figure~\ref{fig:Physical_and_Logical_Compositionality})demonstrating how our two-stage \emph{Logical Parsing $\rightarrow$ Action Instruction Rewriting} mechanism grounds language containing negation and conditionals into executable action sequences. First, for the logical negation instruction “clear the tabletop, leaving only the blue objects behind.” the VLM, using the initial frame, detects that the tabletop contains two screwdrivers and one blue tool, and normalizes the negated description as: $\text{Remove}=\{\text{two screwdrivers}\}$, $\text{Keep}=\{\text{blue tool}\}$. It then produces a linear plan (grasp each screwdriver in turn, place it into the drawer, then reposition the retained blue tool), avoiding common end-to-end failures such as leaving one removable item or mistakenly removing the blue tool. Second, for the conditional instruction “If the drawer is open, take out the blue cube; otherwise, knock the drawer three times.” the VLM first determines the drawer’s open/closed state from the initial frame: if open, it rewrites the prompt into a two-step sequence (grasp the blue cube; lift it out of the drawer); if closed, it rewrites it into an approach plus triple-knock sequence, eliminating the mixed behaviors (simultaneously attempting to open/knock/grasp) often observed when the condition is unresolved by an end-to-end model. The three illustrated sub-scenarios (negation plus the two conditional branches) show that explicit Task Analysis and atomized action listing provide the video generation/control module with a clear target set and ordering constraints, yielding executions that strictly adhere to the linguistic logic, whereas the original complex prompts rarely achieves.
% % . 组合推理能力：物理世界中的语言逻辑 (Compositional Reasoning: Language Logic in the Physical World)
% %   - 核心叙-事：证明模型能理解语言的复杂逻辑（而不仅仅是关键词），并将其正确地“接地”到物理行为中。
% %   - 验证手段： 
% %     - 测试新颖组合、空间关系、逻辑否定（“除了A以外的…”）和条件逻辑（“如果A则B，否则C”）。
% % 桌面上除了苹果以外的都收走-》VLM-> 先收走A,然后B然后C
% \textcolor{red}{LiZhang TODO}
% Logical Negation and Conditional Logic
% Tasks involve:  
% 1. Logical negation (\textit{"Move all objects except the apple."}).  
% 2. Conditional logic (\textit{"If the drawer is open, grab the key; otherwise, knock three times."}).  
% \textbf{Evaluation:} Success rate in reasoning and execution.

% \subsection{Consistency in Planning and Generation}
% \subsubsection{Cross-Verification of Outputs}
% The model's generated video (\textit{V\_gen}) and action sequence (\textit{A\_gen}) are cross-validated in a high-fidelity simulator.  
% \textbf{Metrics:} Consistency score between imagined and executed trajectories.

% \subsection{Closing the Sim-to-Real Loop}
% \subsubsection{Real-World Execution of Planned Actions}
% The model's latent-space-planned trajectories are deployed in real-world robots to test transfer success rates.  

% \subsubsection{Failure Recovery and Replanning}
% Robots dynamically replan using the world model when unexpected deviations occur in real-world execution scenarios.

\section{Foundation Model For Application Post-Training}

\subsection{Novel-View Synthesis and Generation}

\begin{figure}[t]
  \centering
  \includegraphics[width=0.8\linewidth]{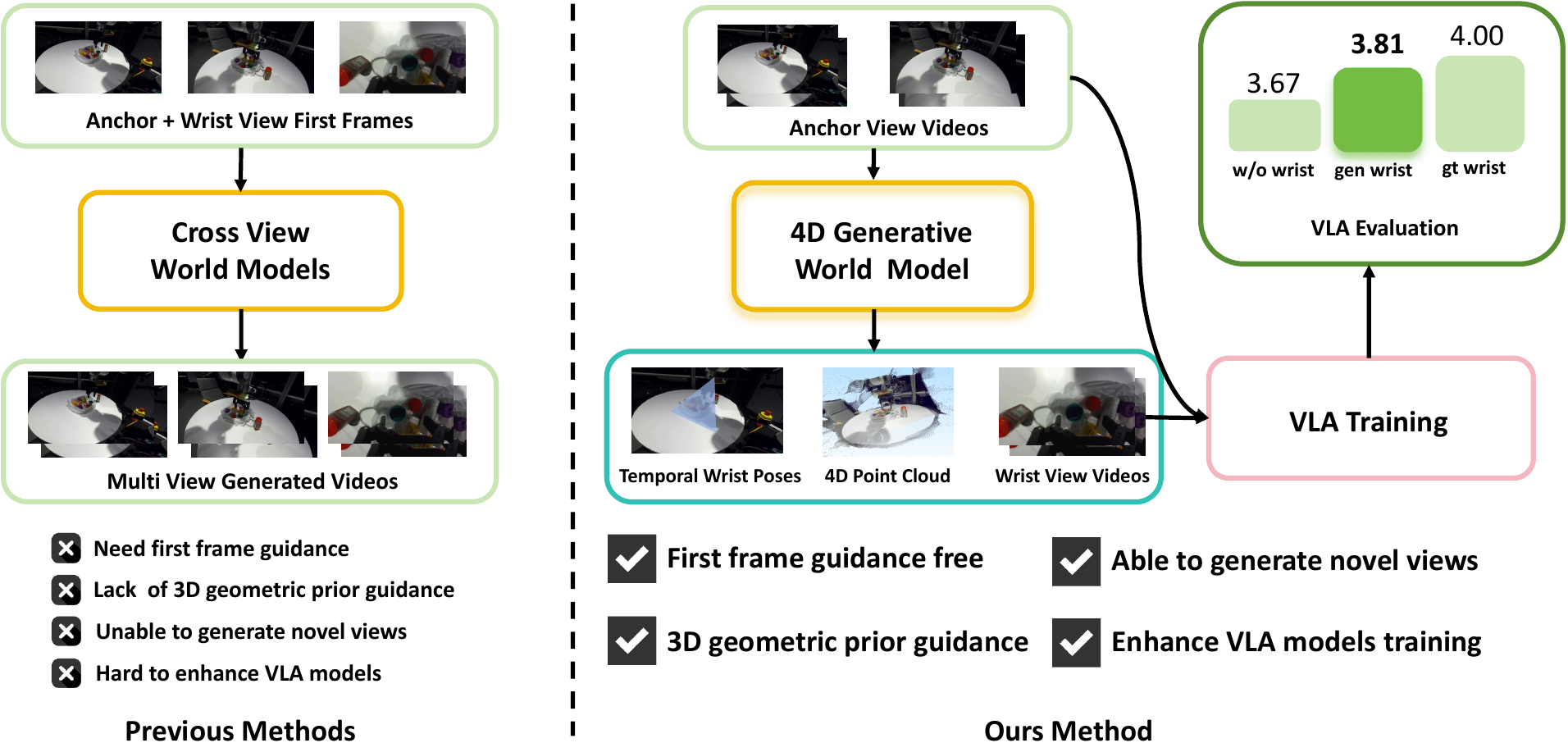}
  \caption{
    \textbf{Advantages of the 4D Generative World Model over standard cross-view World Models.}
    Our method removes the need for first-frame guidance, enables consistent novel-view generation, 
    leverages 3D geometric priors for more reliable conditioning, and enhances VLA model training 
    by providing richer viewpoint data.
  }
  \label{fig:advantages}
\end{figure}

Leveraging the strong cross-domain generalization of foundational models in robotics, we reconstruct geometrically consistent novel views from limited 3D evidence.
Contemporary VLA systems are often constrained by the small number of available viewpoints, which limits the use of egocentric perspectives such as the wrist camera.
We therefore propose a \emph{4D World Model} pipeline (Figure~\ref{fig:advantages}) that extends standard world models into a temporally coherent 3D space and reconstructs a chosen target viewpoint.

Our pipeline first leverages Visual Geometry Grounded Transformer (VGGT) \citep{wang2025vggt} to reconstruct geometry from a small number of anchor views. 
We establish dense 2D correspondences across views, lift them into a 3D point cloud, and use a dedicated \emph{wrist head} to regress the target wrist-view pose from multi-view features. 
The reconstructed points are then projected into the wrist image plane using the estimated pose and dataset-provided intrinsics, forming a coarse condition map. 
Training is guided by a projection-based loss: for forward-facing points we minimize reprojection error between predicted and matched pixels, while for back-facing points we encourage positive depth to ensure geometric feasibility. 
In the second stage, these condition maps are encoded with a variational autoencoder and concatenated with noisy wrist-view latents. 
At the same time, CLIP embeddings from anchor views are projected into the conditioning space, enriched with temporal and view embeddings to capture dynamics and camera identity. 
By fusing geometric alignment with semantic guidance, the diffusion-based generator synthesizes long-horizon, temporally coherent wrist-view videos, without requiring first-frame inputs or task-specific textual prompts.

\begin{figure}[t] \centering \includegraphics[width=0.8\linewidth]{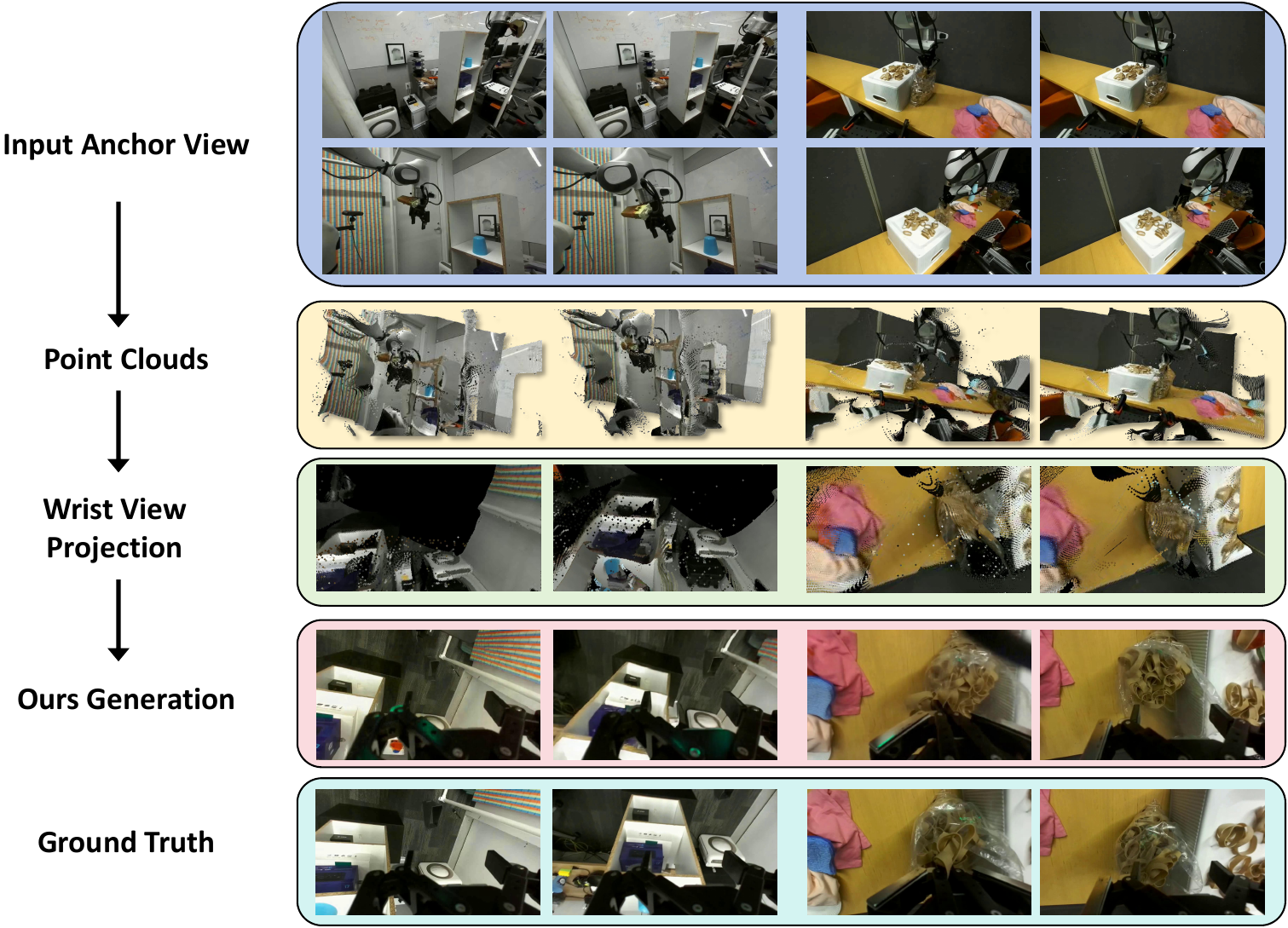} \caption{
Overview of our \emph{4D World Model} pipeline. 
Given a small set of external anchor views (top), we reconstruct geometry and lift it into point clouds. 
A dedicated wrist-view head predicts the egocentric camera pose, enabling wrist-view projection of the 3D evidence. 
From these coarse condition maps, our diffusion-based generator synthesizes high-fidelity, temporally coherent wrist-view videos (ours), which align closely with ground-truth wrist observations. 
The pipeline effectively bridges third-person anchor views and egocentric perspectives, ensuring geometric consistency and perceptual realism.
}
\label{fig:novel_view_vis} \end{figure}

Figure~\ref{fig:novel_view_vis} shows qualitative visualizations of our generated wrist views compared with baselines. 
Our method produces sharper, geometrically consistent, and viewpoint-aligned sequences, 
demonstrating strong generalization from third-person to egocentric perspectives.  

\subsection{Spatial-Aware Trajectory-Guided Video Generation}

\begin{figure}
    \centering
    \includegraphics[width=\linewidth]{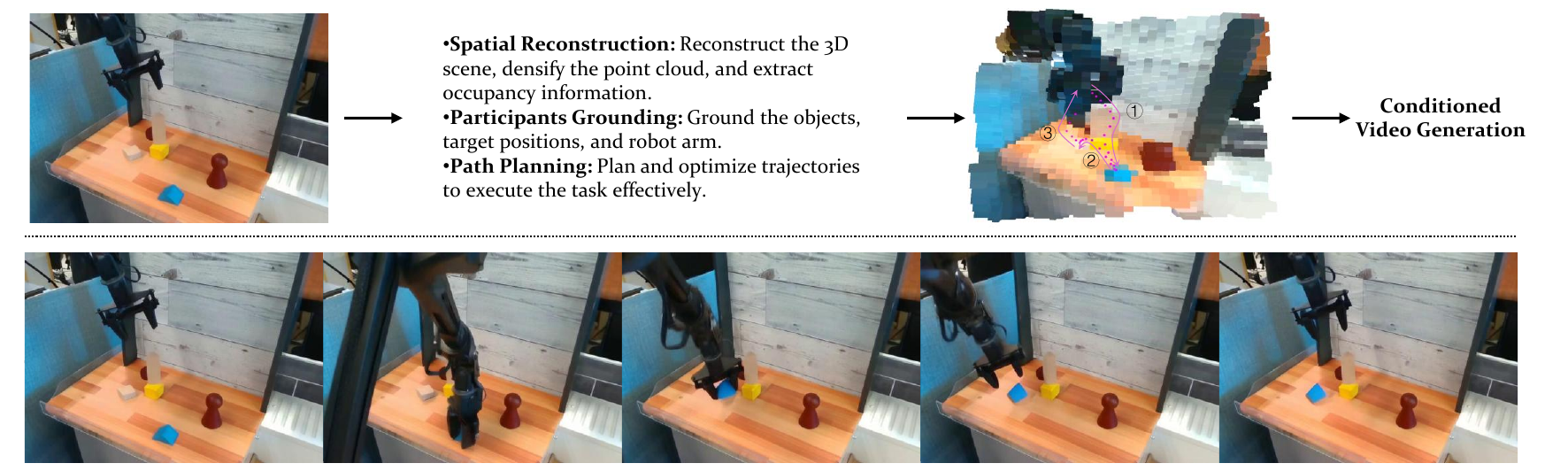}
    \caption{\textbf{Demonstration of WoW when lift to 3D occupancy environment.} We first plan and optimize a plausible path in 3D occupancy environment, and conducts a trajectory-guided video generation process afterwards to produce corresponding high-quality video following~\citep{li2025manipdreamer3dsynthesizingplausible}. }
    \label{fig:md3d_overview}
\end{figure}

The generative model implicitly learns robotic trajectory planning but may not always produce safe or realistic executable paths. To address this, we propose a method for generating action-conditioned robotic manipulation videos that serve as a realistic simulator for policy learning. We are able to synthesize diverse, plausible manipulation demonstrations from a single third-view image and instruction. In detail, we follow the ManipDreamer3D~\citep{li2025manipdreamer3dsynthesizingplausible}, plans and optimizes a physics-aware trajectory in 3D occupancy. It then uses a  VDM conditioned on both visual inputs and action trajectories to generate corresponding manipulation videos, prioritizing physical realism, trajectory rationality, and the inertial properties of the robot arm. The overview and visualization is shown in Figure~\ref{fig:md3d_overview}.

\paragraph{\textbf{Trajectory-Conditioned Video Synthesis}}

% The planned 3D trajectories are integrated into video generation guidance through:
% 1) First-frame latent encoding using the diffusion model’s VAE,
% 2) 3D-to-2D projection of object and end-effector positions,
% 3) Dynamic latent construction by incorporating trajectory-guided edits into the first-frame latent.
% Our latent-editing method concatenates the trajectory representation with the noisy video latent, preserving the original diffusion framework while enabling precise motion control. This supports keypoint-level manipulation and affordance-level control accuracy. The resulting system elevates trajectory planning from a robotic framework to an effective video simulator that maintains geometric consistency across frames and reduces reliance on costly real-world data collection.

% \textbf{Experiment comparison results} We compare our proposed ManipDreamer3D against existing baselines: DragAnything~\citep{wu2024draganything}, This\&That~\citep{wang2025language}, and RoboMaster~\citep{fu2025learning}. Experimental results show superior performance in both video quality and trajectory accuracy. The SVD version achieves the lowest FVD (143.33), highest PSNR (22.75), and SSIM (0.807) compared to other SVD-based methods, with minimal trajectory errors (17.40 for the robot end-effector and 18.77 for the object), significantly outperforming This\&That (62.07 / 37.12 for the end-effector/object). The DiT version yields similar results compared to RoboMaster.

\subsection{Action-to-Video Generation for Robot Manipulation}

A critical limitation of existing world models for robot manipulation lies in their inability to accurately capture \textbf{fine-grained robot--object interactions} from textual descriptions. This stems from the inherent \textbf{modality gap} between natural language and video frames. In contrast, state-of-the-art robotic policies such as \textbf{DP}~\citep{dp}, \textbf{ACT}~\citep{act}, and \textbf{OpenVLA}~\citep{openvla} represent behavior through dense action trajectories that specify end-effector positions, orientations, and gripper states. While text-to-video models may appear as a potential viable path, they primarily rely on high-level textual cues rather than frame-specific action instructions, and thus fail to model robot control with the required temporal precision.  

\begin{figure}[t]
  \centering
  \includegraphics[width=1\linewidth]{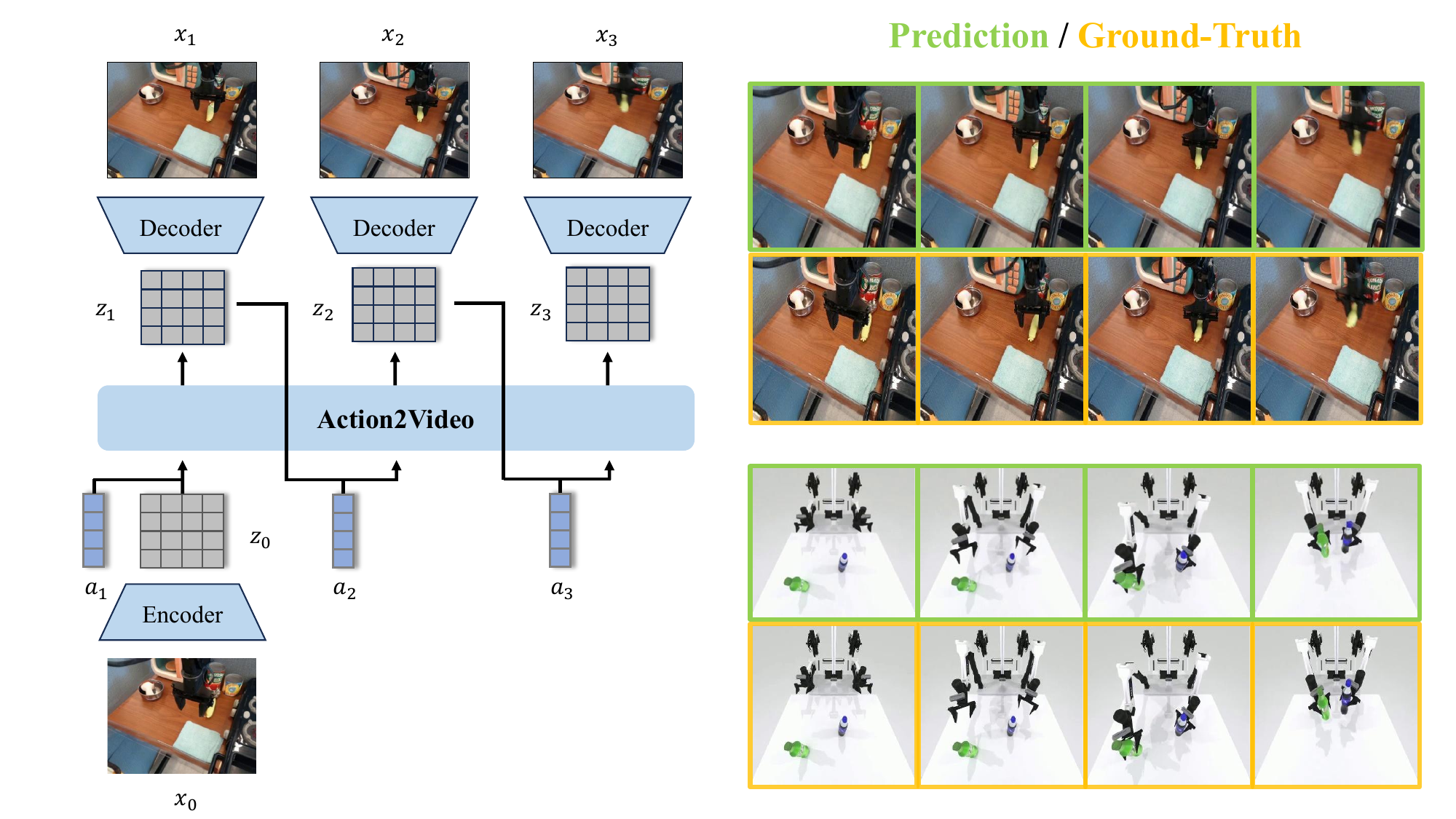}
  \caption{
    \textbf{Inference procedure of Action-to-Video}
    Action-to-Video trains a latent video diffusion model in the latent space provided by pre-trained VAE. An autoregressive spatial-temporal transformer is used to predict future tokens conditioned on the corresponding action at each step.
  }
  \label{fig:a2v}
\end{figure}

In this section, we introduce \textbf{Action-to-Video}, a framework for generating \textbf{high-resolution (up to 640$\times$480)} and \textbf{long-horizon (over 300 frames)} videos of robot manipulation directly from 3D action trajectories describing end-effector states. Following DiT~\citep{dit}, Action-to-Video employs a diffusion-based spatial--temporal transformer backbone to capture complex temporal dynamics. To explicitly align actions with their visual results, we integrate a fine-grained action-conditioning module into each transformer block. This design enables precise correspondence between actions and generated frames, supporting both \textbf{successful and failed rollouts} and modeling precise control behavior such as end-effector rotations. Furthermore, our \textbf{autoregressive rollout} and training recipes allow Action-to-Video to generate long-horizon, temporally consistent videos. 
Qualitative results of our generated videos from action-conditioning are shown in Fig.~\ref{fig:a2v}. We post-train our world model on both single and dual-arm data. Specifically, we give our task formulation below.

% In human evaluations, Action-to-Video is consistently preferred for its fidelity and temporal coherence. Our approach scales effectively with larger model sizes, larger datasets, and increased computation.

\paragraph{Definition.}  
We define the \textit{trajectory-to-video task} as predicting a video of a robot executing a trajectory $a_{t:t+n}$, given a sequence of historical observation frames $O_{t-h:t}$. Formally, $a_t \in \mathbb{R}^d$ denotes the action at timestep $t$, where $d = 7$ for a typical robot arm: three DoFs for translation, three for rotation, and one for gripper control. Figure~\ref{fig:a2v} illustrates the overall inference pipeline of Action-to-Video and presents the prediction results on both single-arm and dual-arm robot datasets. The conditioning input is  
\begin{equation}
c = \{ z_{t-h:t}, a_{t:t+n} \}, \quad z_{t-h:t} = \text{Enc}(x_{t-h:t}),
\end{equation}
and the diffusion target is the latent representation of the subsequent video frames  
\begin{equation}
x_{t+1:t+n+1} = \text{Dec}{(z_{t+1:t+n+1})}.
\end{equation}

% \paragraph{Architecture and Conditioning.}  
% Following DiT~\citep{dit}, we employ memory-efficient spatial--temporal attention. To inject trajectory conditions, we adapt \textbf{adaptive layer normalization}. Unlike text conditioning, which encodes an entire sequence into a single embedding, trajectories require \textbf{frame-level precision}: each action determines the robot’s motion at a specific frame.  

% Concretely, each action is encoded via a linear layer into an embedding, which is then combined with the diffusion timestep embedding to produce frame-specific conditioning vectors. For spatial blocks, the conditioning embedding for the $i$-th frame is  
% \begin{equation}
% c_i^S = \text{Linear}(a_i) + \text{Embed}(t),
% \end{equation}
% which controls the scale and shift parameters $(\gamma, \beta)$ of that frame’s spatial attention. In contrast, temporal blocks share a \textbf{video-level conditioning embedding}  
% \begin{equation}
% c^T = \text{Linear}(a_{t:t+n}) + \text{Embed}(t),
% \end{equation}
% ensuring global temporal consistency. This hierarchical conditioning balances \textbf{fine-grained spatial alignment} with \textbf{long-range temporal coherence}.

\subsection{Visual Style Transfer Enhancement}
% 一张双栏的大图，展示能做的效果，中间模型简单画方块就行
% 我们的basemodel这样那样还可以实现1234。。。
\begin{figure}[htbp]
  \centering
  \includegraphics[width=1\linewidth]{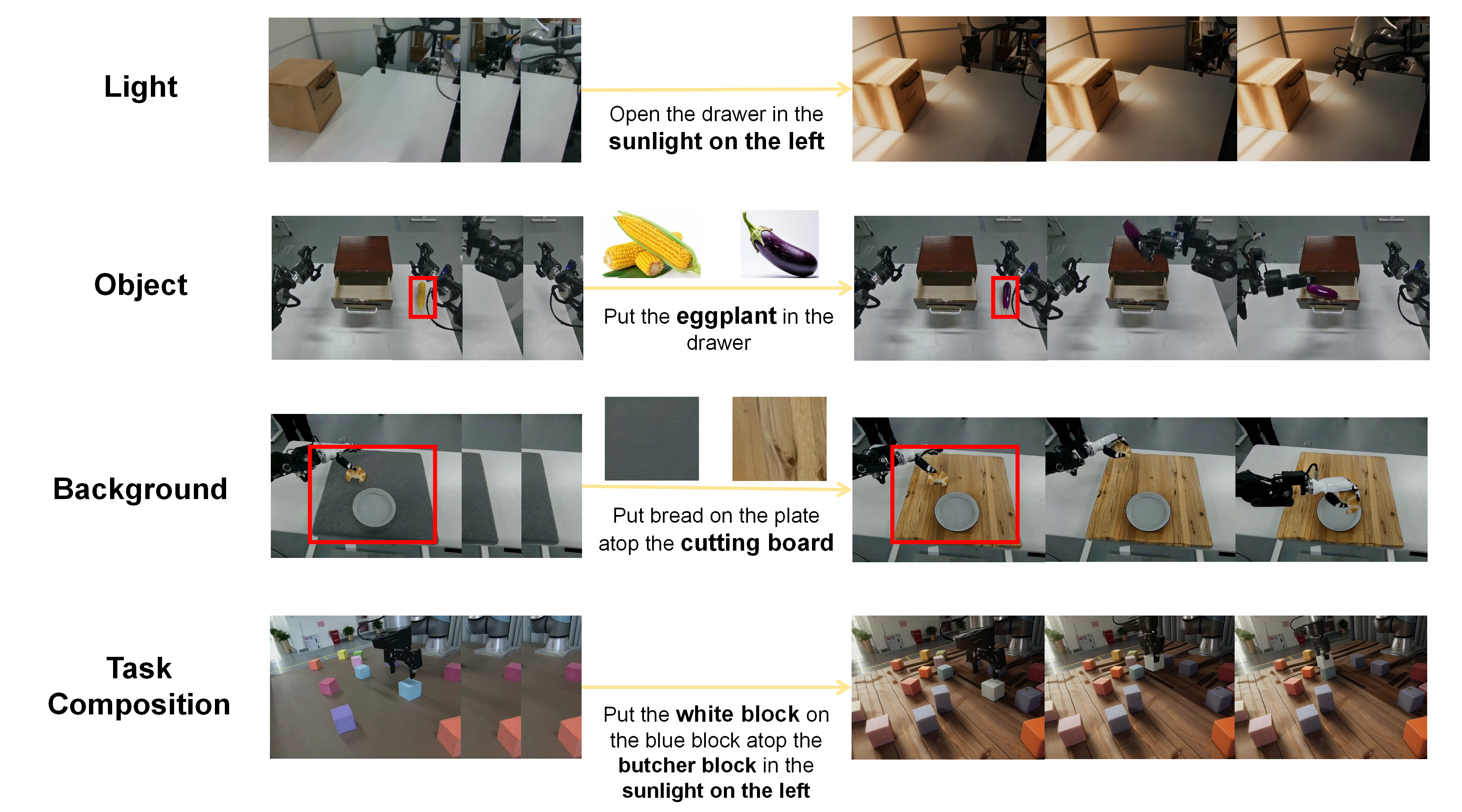}
  \caption{Case study illustrating Visual Style Transfer Enhancement}
  \label{fig:Visual Style Transfer Enhancement}
\end{figure}

The construction of large-scale VLA datasets is fundamentally constrained by the high cost of collecting diverse and realistic paired visual–action data. Such limitations hinder the generalization and adaptability of VLA models to unseen scenarios\citep{zhong2025surveyvisionlanguageactionmodelsaction}. To overcome this limitation, we introduce a \textbf{multimodal Controllable World Generation Toolkit}, which harnesses the powerful generalization capabilities and spatiotemporal consistency of foundational models. Specifically, we propose a Visual Style Transfer Enhancement framework that enables systematic augmentation of existing metadata-driven datasets. By conditioning the model on various controllable factors, such as illumination, background, and object textures, we achieve a scalable synthesis of new VLA data instances. This approach not only expands the volume of available training data, but also enriches the diversity and robustness of the learned representations. 
% through training free assertion module\citep{}, or a lightweight post-finetuning following\citep{VACE}.
Our usage paradigm is twofold: (1) employing the embodied world model as a video generation backbone in conjunction with train-free modules, and (2) fine-tuning the embodied world model into a multi-modal controllable world generator.
% 对应有一个实验图其实。
% Light: We imploy \citep{light a video}, on robot data,....1234 做法，实现了x光照增强。
% embodiedment: 前背景分离+纯语义 robot  取到embodiedment mask
% object: SAM-Motion，Further more，to enhance the object generalization ablilty... 前背景分离+Object语义基础上进一步增加了 动作信息保证物体正确；   for example, 取到object mask
% background: 为啥background last, 取并集然后取反； SAM-Motion，前背景分离找到bgm增强我们用mask；
% multi-condition mixture: We also experiment all mixture。。。

To validate the effectiveness of our toolkit, we designed a series of controlled experiments on robot-centric VLA datasets, focusing on light, embodiment, object, and background augmentation. Figure~\ref{fig:Visual Style Transfer Enhancement} illustrates the overall pipeline of our visual controllability experiments.

\textbf{Light.} Following Light-A-Video\citep{zhou2025lightavideotrainingfreevideorelighting}, we apply controllable light transfer to robot manipulation data. By conditioning the video backbone on light descriptors (e.g., global brightness, local shadows, and dynamic reflections), we achieve scalable \emph{light augmentation}. This enables VLA models to generalize in environments with diverse illumination conditions, from daylight to dim indoor scenes.

\paragraph{Embodiment.} To ensure accurate disentanglement of robot embodiment, we first perform foreground–background separation using semantic segmentation. A mask of the robot embodiment is extracted to preserve semantic consistency of robotic arms and tools, while decoupling them from volatile environmental factors. This guarantees that the core embodiment features remain invariant across augmentation cycles, improving robustness in action grounding.

\paragraph{Object.} For object-level augmentation, we adopt SegAnyMo to generate object-specific masks. Unlike static segmentation, SegAnyMo\citep{Huang2025SegmentAM} incorporates temporal motion cues, ensuring that the augmented objects retain both semantic accuracy and action-aligned consistency. By enriching textures, materials, and dynamics of manipulated objects, we substantially enhance the generalization of the model in object-centric tasks (e.g. grasping or tool use).

\paragraph{Background.} Background augmentation is performed as the final stage. We compute the union of foreground masks (embodiment + objects) and take its complement to isolate the background. Leveraging SegAnyMo, diverse environmental contexts are synthesized, ranging from kitchens to laboratories, while preserving the action-relevant components. This design ensures that background variation does not disrupt robot–object interactions.

\paragraph{Multi-condition Mixture.} Beyond individual factors, we perform multicondition augmentation by mixing light, embodiment, object, and background variations. This yields a compositional generative pipeline in which controllable factors can be flexibly combined. Our results demonstrate that such mixtures not only increase data diversity but also simulate complex, real-world variations that naturally occur in embodied environments.

\subsection{Test-Time Scaling for VLM task planning}
% \citep{SEEA-R1}
% 我们发现VLM可以在世界模型里验证planning来调试事实逻辑
% 所以我们设计了一个叠方块的task;不同颜色方块,同色归类; qwen7B废了,qwen72B准确率30%,我们给qwen72B--->[流程]，发现平均经过x轮迭代交互之后，给出的任务成功率达到90%
% 1. 流程框图
    % 介绍一下流程图和我们的流程。
% 2. 连续生成的最后一帧的例子图（成功、失败）

% 我们发现世界模型可以帮助VLM发现规划错误并反思出正确的规划和动作，进一步提高任务完成率。
% 具体地说，我们将世界模型作为具身规划任务的交互环境，并设计不同颜色方块的场景和"Separate cubes of different colors and stack cubes of the same color"任务来测试不同多模态大模型的性能。
% 如图所示，具体流程为：
% 1.多模态大模型作为规划器，根据当前环境状态和任务，输出规划以及当前动作
% 2. 世界模型基于当前的图片和动作生成视频，并将视频尾帧作为环境观察反馈给多模态大模型
% 3. 另一个更强大的多模态大模型作为奖励模型，根据当前环境状态判断任务是否是成功/失败/继续
% 3.1 若完成或者失败，则结束
% 3.2 若继续，则重复1/2/3步
% 
% 如表格所示，实验发现，随着多模态大模型与世界模型的交互次数提升，多模态大模型输出的规划正确率越高，另一方面的任务完成率越高。
\begin{figure}[htbp]
  \centering
  \includegraphics[width=1\linewidth]{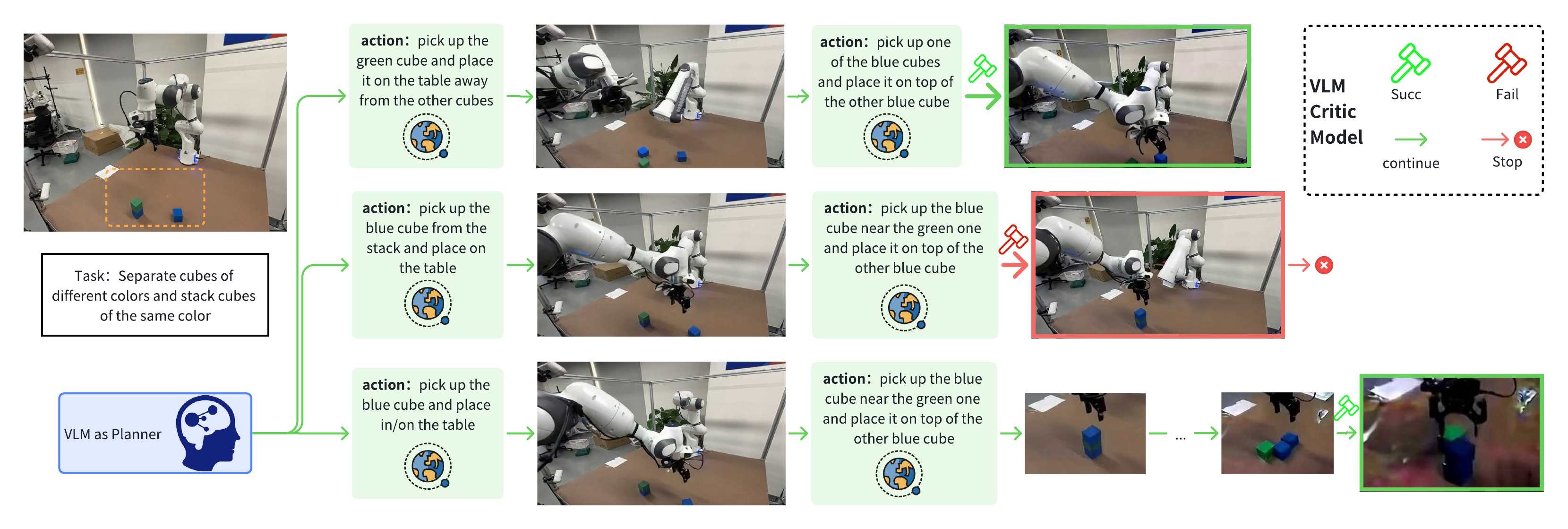}
  \caption{        \textbf{Self-correction of VLM planning via world model simulation.} 
        (a) Our iterative loop: a VLM planner proposes an action, a world model simulates the future frame, and a VLM critic provides feedback, enabling the planner to refine its next step. 
        (b) Terminal frames from the simulation, illustrating a successful plan (top) versus a detected failure (bottom) that triggers re-planning.
 }
  \label{fig:vlm_self_correction_loop}
\end{figure}

\begin{table}[htbp]
  \centering
  \caption{Planning success rate and task success rate gradually improve with the increase in the number of interactions with the world model. Task succ refers to the task completion rate after performing actions in the environment, and planning succ refers to the accuracy rate of the task plans output by VLM before each action is executed.}
  \begin{tabular}{lccc}
    \toprule
    Model & Interactions Round & Planning Succ. & Task Succ. \\
    \midrule
    Qwen-2.5-VL-7B-Instruct & 0 & 1/3 & 0 \\
    Qwen-2.5-VL-7B-Instruct & 1 & 4/9 & 0/3 \\
    Qwen-2.5-VL-7B-Instruct & 2 & 8/9 & 4/9 \\
    \bottomrule
  \end{tabular}
  \label{tab:model_performance}
\end{table}
% 这一方面证明了世界模型可以作为具身规划器的交互环境，一方面证明世界模型可以在交互过程中帮助多模态大模型验证规划的正确性并促进大模型反思出正确的规划和动作，进一步地为世界模型作为环境进行强化学习提供了实验基础。

We find that generative world models can serve as interactive sandboxes, enabling VLMs to debug their own logical fallacies in long-horizon planning. To validate this, we designed a complex spatial reasoning task requiring an agent to "Separate cubes of different colors and stack cubes of the same color." Our experiments reveal that even a powerful baseline like Qwen-7B struggles with the inherent ambiguity, achieving only a 30\% success rate in a single-pass planning attempt.

\paragraph{Task Setting.}
We design a spatial reasoning task to evaluate the self-correction capability of VLM. The agent operates in a simulated tabletop environment with cubes of different colors. The goal is to \textit{``separate cubes of different colors and stack cubes of the same color.''} This task requires multi-step reasoning and introduces ambiguity that cannot be resolved in a single planning pass, making it suitable for testing interactive planning and feedback.

\paragraph{Quality Comparison.}
We compare single-pass planning against iterative planning with feedback from a generative world model. As shown in Table~\ref{tab:model_performance}, single-pass planning achieves low task success due to planning errors and ambiguity. By introducing a cognitive loop—where the VLM proposes sub-goals, receives simulated feedback, and updates its plan—the model significantly improves. After two interaction rounds, planning success rises to 89\%, and task success increases to 44\%. This highlights the effectiveness of simulated feedback in helping VLMs reflect, revise, and succeed in complex tasks.

\paragraph{Experiment Result }% 所以我们设计了一个叠方块的task;不同颜色方块,同色归类; 
To address this, we implemented a cognitive loop where the VLM's planning is grounded in simulated feedback, inspired by MindJounary~\citep{yang2025mindjourneytesttimescalingworld}. As illustrated in Figure \ref{fig:vlm_self_correction_loop}, the process is as follows: (1) The VLM proposes a sub-goal. (2) Our world model simulates the action's outcome, providing a resulting video frame. (3) A VLM critic evaluates the new state for task progress. This iterative refinement allows the planner to self-correct from simulated failures (Figure \ref{fig:vlm_self_correction_loop}). By engaging in this loop, Qwen-7B's~\citep{bai2025qwen25vltechnicalreport} task success rate dramatically increased to 89\% after an average of X interactions, demonstrating that simulated, iterative feedback is crucial for resolving planning ambiguities and achieving robust task completion.

We compare single-pass planning against iterative planning with feedback from a generative world model. As shown in Table~\ref{tab:model_performance}, single-pass planning achieves low task success due to planning errors and ambiguity. By introducing a cognitive loop—where the VLM proposes sub-goals, receives simulated feedback, and updates its plan—the model significantly improves. After two interaction rounds, planning success rises to 89\%, and task success increases to 44\%. This highlights the effectiveness of simulated feedback in helping VLMs reflect, revise, and succeed in complex tasks.

\section{Conclusion and Future Work}

This work presented WoW, a world model forged through embodied interaction, and subjected it to a rigorous tribunal of five core Research Questions(RQ). The findings from this comprehensive evaluation are not merely incremental; they represent a fundamental step toward physically-grounded artificial intelligence. We declare the final verdicts below.

\begin{itemize}
    \item[\textbf{(RQ1)}] \textbf{On Power and Law:} WoW establishes a new state-of-the-art, decisively outperforming contemporary world models on our WoWBench benchmark (Table~\ref{tab:foundation-model-comparison}). Its power is governed by scaling benefit with respect to model size and data volume, yet our analysis reveals that the path to mastering complex, \textit{hard} physical reasoning tasks remains challenging that demands further scaling (Sections~\ref{sec:diffmodel} and \ref{subsec:scaling}).

    \item[\textbf{(RQ2)}] \textbf{On Generalization:} WoW's understanding of physics is abstract and universal, not superficial. It demonstrates profound generalization to novel robot embodiments, manipulation tasks, and visual domains without fine-tuning (Figures~\ref{fig:robot_type}, \ref{fig:action_type}, and \ref{fig:other_generalize}). This proves it learns the underlying principles of interaction, not merely the context in which they were trained.

    \item[\textbf{(RQ3)}] \textbf{On Imagination:} WoW's capabilities transcend mere replication. It can reason about and generate physically-consistent outcomes for \textit{counterfactual} scenarios. When instructed that an object is "impossibly heavy," it simulates a failed attempt, not a successful lift (Figure~\ref{fig:counterfactual_breadth}). This marks a critical shift from a pattern-matching generator to a reasoning engine.

    \item[\textbf{(RQ4)}] \textbf{On Cognitive Simulation:} WoW functions as a potent cognitive sandbox for other agents. By enabling a VLM planner to simulate its proposed actions and receive feedback within a closed loop, WoW allows the planner to debug its own logical fallacies. This interactive refinement dramatically increased planning and task success rates from 30\% to nearly 90\% (Figure~\ref{fig:vlm_self_correction_loop}, Table~\ref{tab:model_performance}).

    \item[\textbf{(RQ5)}] \textbf{On Embodied Action:} WoW successfully closes the imagination-to-action loop. Through the Flow-Mask Inverse Dynamics Model (FM-IDM), its generated futures are translated into successful, executable actions on a physical robot. The system achieved remarkable success rates of 94.5\% on easy and 75.2\% on medium-difficulty real-world tasks, proving that its imagined physics are firmly grounded in reality (Section~\ref{subsec:real}, Table~\ref{tab:idm}).
\end{itemize}

In conclusion, WoW is not merely a more powerful video generator. It is the nascent form of a true world model: one that possesses an emergent physical intuition, generalizes across domains, reasons about hypotheticals, serves as an interactive world for other AI, and ultimately, grounds its imagination in successful, physical action. This work lays a cornerstone for the future of embodied intelligence.

\bibliography{iclr2026_conference}
\bibliographystyle{iclr2026_conference}

\end{document}